\documentclass[letterpaper]{article} % DO NOT CHANGE THIS
\usepackage{aaai2026}  % DO NOT CHANGE THIS
\usepackage{times}  % DO NOT CHANGE THIS
\usepackage{helvet}  % DO NOT CHANGE THIS
\usepackage{courier}  % DO NOT CHANGE THIS
\usepackage[hyphens]{url}  % DO NOT CHANGE THIS
\usepackage{graphicx} % DO NOT CHANGE THIS
\usepackage{natbib}  % DO NOT CHANGE THIS AND DO NOT ADD ANY OPTIONS TO IT
\usepackage{caption} % DO NOT CHANGE THIS AND DO NOT ADD ANY OPTIONS TO IT
\usepackage{algorithm}
\usepackage{algorithmic}
\usepackage{amsmath} 
\usepackage{booktabs} 
\usepackage{adjustbox}
\usepackage{multirow} 
\usepackage{subcaption}
\usepackage[most]{tcolorbox}
\usepackage{amssymb}
\usepackage{mathtools}
\usepackage{amsthm}
\usepackage{enumitem}

\usepackage{newfloat}
\usepackage{listings}
\DeclareCaptionStyle{ruled}{labelfont=normalfont,labelsep=colon,strut=off} % DO NOT CHANGE THIS
\floatstyle{ruled}
\newfloat{listing}{tb}{lst}{}
\floatname{listing}{Listing}
\newtheorem{theorem}{Theorem}[section]
\newtheorem{proposition}[theorem]{Proposition}
\newtheorem{lemma}[theorem]{Lemma}

\newtheorem{definition}[theorem]{Definition}

\newtheorem{remark}[theorem]{Remark}

\newcommand{\bfs}[1]{{\mathbf #1}}
\newcommand{\R}{\mathcal{R}}
\newcommand{\E}{\mathbb{E}}
\newcommand{\F}{\mathcal{F}}
\newcommand{\I}{\mathcal{I}}
\newcommand{\Q}{\mathcal{Q}}
\newcommand{\D}{\mathcal{D}}
\newcommand{\Loss}{\mathcal{L}}
\title{EXAGREE: Mitigating Explanation Disagreement with \\ Stakeholder-Aligned Models}
\author{
    Sichao Li\textsuperscript{\rm 1},
    Tommy Liu\textsuperscript{\rm 1},
    Quanling Deng\textsuperscript{\rm 2},
    Amanda Barnard\textsuperscript{\rm 1}
}
\affiliations{
    \textsuperscript{\rm 1}The Australian National University, 
    \textsuperscript{\rm 2}Tsinghua University
}

\usepackage{bibentry}
\begin{document}

\maketitle

\begin{abstract}
Conflicting explanations, arising from different attribution methods or model internals, limit the adoption of machine-learning models in safety-critical domains.  We turn this disagreement into an advantage and introduce EXplanation AGREEment (EXAGREE), a two‑stage framework that selects a Stakeholder‑Aligned Explanation Model (SAEM) from a set of similar-performing models. The selection maximizes Stakeholder‑Machine Agreement (SMA), a single metric that unifies \emph{faithfulness} and \emph{plausibility}. 
EXAGREE couples a differentiable mask‑based attribution network (DMAN) with monotone differentiable sorting, enabling gradient‑based search inside the constrained model space. Experiments on six real‑world datasets demonstrate simultaneous gains of faithfulness, plausibility, and fairness over baselines, while preserving task accuracy. Extensive ablation studies, significance tests, and case studies confirm the robustness and feasibility of the method in practice.
\end{abstract}


\section{Introduction}

% Machine Learning (ML) models are increasingly deployed in critical fields such as healthcare, science, and finance, where the demand for explainability has grown in high-stakes decision-making processes \citep{kailkhura2019reliable, wiens2018machine, carvalho2022artificial, agarwal2022openxai, ghassemi2021false}. However, an open problem persists, where different ML models or explanation methods produce conflicting explanations \citep{krishna2022disagreement, rudin2019stop, li2024diverse}, and poses a barrier to trust and usability in these domains. 

Machine Learning (ML) models are increasingly deployed in critical fields such as healthcare, science, and finance, where explainability is increasingly important for high-stakes decision-making processes \citep{kailkhura2019reliable, wiens2018machine, carvalho2022artificial, agarwal2022openxai, ghassemi2021false}. 
Unfortunately, state‑of‑the‑art eXplainable Artificial Intelligence (XAI) techniques frequently disagree. Different attribution algorithms, or the same algorithm applied across equally accurate models for a given task may produce different feature attributions\citep{krishna2022disagreement, rudin2019stop, li2024diverse}. Such \emph{explanation disagreement} erodes trust, impedes auditing, and complicates regulatory compliance.

\paragraph{Why does disagreement occur?} (i) Attribution algorithms encode different axioms (sensitivity, additivity, counterfactual consistency) meaning that they emphasize different features \citep{sundararajan2020many, krishna2022disagreement, li2024diverse}. (ii) Most tasks admit a \emph{Rashomon set} of near‑optimal models with various internal mechanisms \citep{fisher2019all, rudin2019stop, dong2020exploring, ghorbani2019interpretation, adebayo2018sanity, weber2024xai}; choosing one model over another changes the explanation even when accuracy is unchanged. Disagreement among explanations is thus an inevitable symptom of model-class multiplicity.

\paragraph{Stakeholders' needs first.}  Satisfying diverse human demands is more urgent in practice \citep{chromik2020taxonomy, kong2024toward}. However,
diverse stakeholders such as developers, professionals, and end-users, have distinct objectives, expertise, and priorities. This diversity makes it impractical to identify a single explanation method, or ML model that satisfies all perspectives simultaneously \citep{miller2023explainable, krishna2022disagreement, imrie2023multiple, binns2018fairness, hong2020human}. 
To bridge the gap across the diverse needs, we distill the problem into a single question:

\begin{quote}
\emph{Given a set of near-optimal models, can we pick the one with explanations that match a specific goal of the stakeholder?}
\end{quote}

\noindent \textbf{Our answer.} We treat disagreement not as a defect to eliminate but as a resource to \textit{mitigate} conflicting stakeholder needs. Our key contributions can be summarized as:
\begin{itemize}[itemsep=0pt,topsep=0pt]
    \item We formalize the \textit{Stakeholder–Machine Agreement} (SMA), a rank correlation metrics that quantifies the alignment between model-grounded and stakeholder-grounded explanations. SMA characterizes the Pareto frontier between \emph{faithfulness} (agreement with the model’s internal logic) and \emph{plausibility} (agreement with stakeholder priors).
    % \ale{QD:model/stakeholder-oriented instead of grounded? Or maybe use stakeholder-oriented in the title instead of stakeholder-aligned? }
    \item We introduce the \textit{EXplanation
    AGREEment} (EXAGREE) framework within a set of near-optimal models, as known as Rashomon set \cite{fisher2019all}, aiming to identifying a \textit{Stakeholder-Aligned Explanation Model} (SAEM) by maxmizing the SMA. 
    \item We develop a two-stage optimization strategy based on: (i) Rashomon set exploration via constrained sampling; (ii) a differentiable sorting–ranking objective optimizing SMA; and (iii) efficient model search using a multi-head, mask-based attribution network. 
    \item EXAGREE improves up to {\small +0.43} on faithfulness and {\small +0.51} on plausibility by maximizing SMA, while reducing subgroup fairness gaps by {\small -0.28}, all without sacrificing predictive accuracy across six public datasets.
\end{itemize}

\noindent \textbf{In short.} EXAGREE turns the ``explanation paradox'' into an actionable model-selection problem, offering a principled foundation for stakeholder-centered XAI.

\section{Preliminaries}\label{sec:preliminaries}
\noindent \textbf{Notation and Setting.}
Let $\mathbf{X}\!\in\!\mathbb{R}^{n\times p}$ be a data matrix of $n$ instances and
$p$ features, and $\mathbf{y}\!\in\!\mathbb{R}^{n}$ the corresponding targets.
A model $M$ defines a predictor $f_{M}\!:\mathbb{R}^{p}\!\to\!\mathbb{R}$ optimized
for loss $\mathcal{L}(f_{M}(\mathbf{X}),\mathbf{y})$.
We denote $\mathcal{M}$ as the \emph{Rashomon set}
\begin{equation} \label{eq:grs}
\mathcal{M}_{\epsilon} = \{ M : \Loss(M(\bfs{X}), \bfs{y}) \leq (1 + \epsilon) \Loss(M^*(\bfs{X}), \bfs{y}) \}.
\end{equation}
where each model in the set meets a predefined performance threshold $\epsilon$ and $M^*$ is a reference well-trained model for a given task \citep{fisher2019all, dong2020exploring, Li2023VTF}. Generally, bold lower‐case symbols are vectors, bold upper‐case symbols are matrices; A \textit{summary} of notations is provided in Appendix Table 4. 

\noindent \textbf{Feature Attributions and Explanations Rankings.}
We discuss the explanation in the context of feature attribution for ML models, where the attribution assigned to a feature is a measure of that feature's contribution to the model's prediction \citep{krishna2022disagreement, sundararajan2020many}. For a model $M \in \mathcal{M}$ and an explanation method $\varphi \in \Phi$, we calculate feature attributions as: 
$\bfs{a}^{M}_{\varphi} = (a_{\varphi, 1}^{M}, a_{\varphi, 2}^{M}, \ldots, a_{\varphi, p}^{M}).$ These attributions yield a ranking:
$\bfs{r}^{M}_{\varphi} = (r_{\varphi, 1}^{M}, r_{\varphi, 2}^{M}, \ldots, r_{\varphi, p}^{M}),$ where $r_{\varphi, i}^{M}$ represents the rank of feature $i$. 
This ranking is derived from the ordering of attributions:
$a_{(1)} \succ a_{(2)} \succ \ldots \succ a_{(p)}$,
where $a_{(i)}$ denotes the $i$-th largest attribution. 
For interpretable models $M_\I \in \mathcal{M}_\I \subset \mathcal{M}$, such as decision trees and linear regressions, we can obtain attributions $\bfs{a}^{M_\I}_{\text{true}}$ and rankings $\bfs{r}^{M_\I}_{\text{true}}$ that reflect the importance that model parameters assign to inputs. 

\begin{definition}[Delivered Explanation]
    We define the term \textit{delivered explanation} ($\bfs{r}^{M}_{\varphi}$) any explanation provided to a given stakeholder. This explanation can be intrinsic or post-hoc, as a general category of explanations provided to stakeholders in practical settings, irrespective of whether it is intrinsic or post-hoc.
\end{definition}

% \subsection{Notations}
% Bold lowercase letters such as $\bfs{v}$ to represent a vector and ${v}_i$ denotes its $i$-th element.
% Let the bold uppercase letters such as $\bfs{A}$ denote a matrix with ${a}_{[i, j]}$ being its $i$-th row and $j$-th column entry. 
% The vectors $\bfs{a}_{[i, \cdot]}$ and $\bfs{a}_{[\cdot, j]}$ are its $i$-th row and $j$-th column, respectively.
% Usually, the row is an instance and the column is a variable or feature.
% Let $(\bfs{X}, \bfs{y})\in \mathbb{R}^{n\times (p+1)}$ denote a dataset 
% where $\bfs{X} = [\bfs{x}_{[\cdot, 1]},  \bfs{x}_{[\cdot, 2]}, ..., \bfs{x}_{[\cdot, p]}]$ is a $n \times p$ covariate input matrix and $ \bfs{y} $ is a $n$-length output vector. $\bfs{X}_{\setminus \bfs{i}}$ is the input matrix when the feature of interest (denoted as $\bfs{i}$ here) is replaced by an independent variable. Let $f :  \mathbb{R}^{n\times p} \to  \mathbb{R}^n$ be a predictive model and $\Loss :  (f(\bfs{X}), \bfs{y}) \to \mathbb{R}$ be the loss function. Attributions are measured by $s_{i} =  \E[\Loss(f(\bfs{X}_{\setminus i}), \bfs{y})] - \E[\Loss(f(\bfs{X}), \bfs{y})]$, where $\E[\Loss(f(\bfs{X}), \bfs{y})]$ is the expected loss and baseline effect that provides interpretability \citep{li2023exploring}. We denote $\bfs{s} = \{s_{1}, s_{2}, ..., s_{p}\}$ as all attributions for all variables and a potential corresponding rank is $\bfs{r}_{\bfs{s}} = \{s_{1} \succ s_{2} \succ ... \succ s_{p}\}$. 

% \textbf{Problem Statement}

\begin{figure}[t!]
    \centering
    \includegraphics[width=\linewidth]{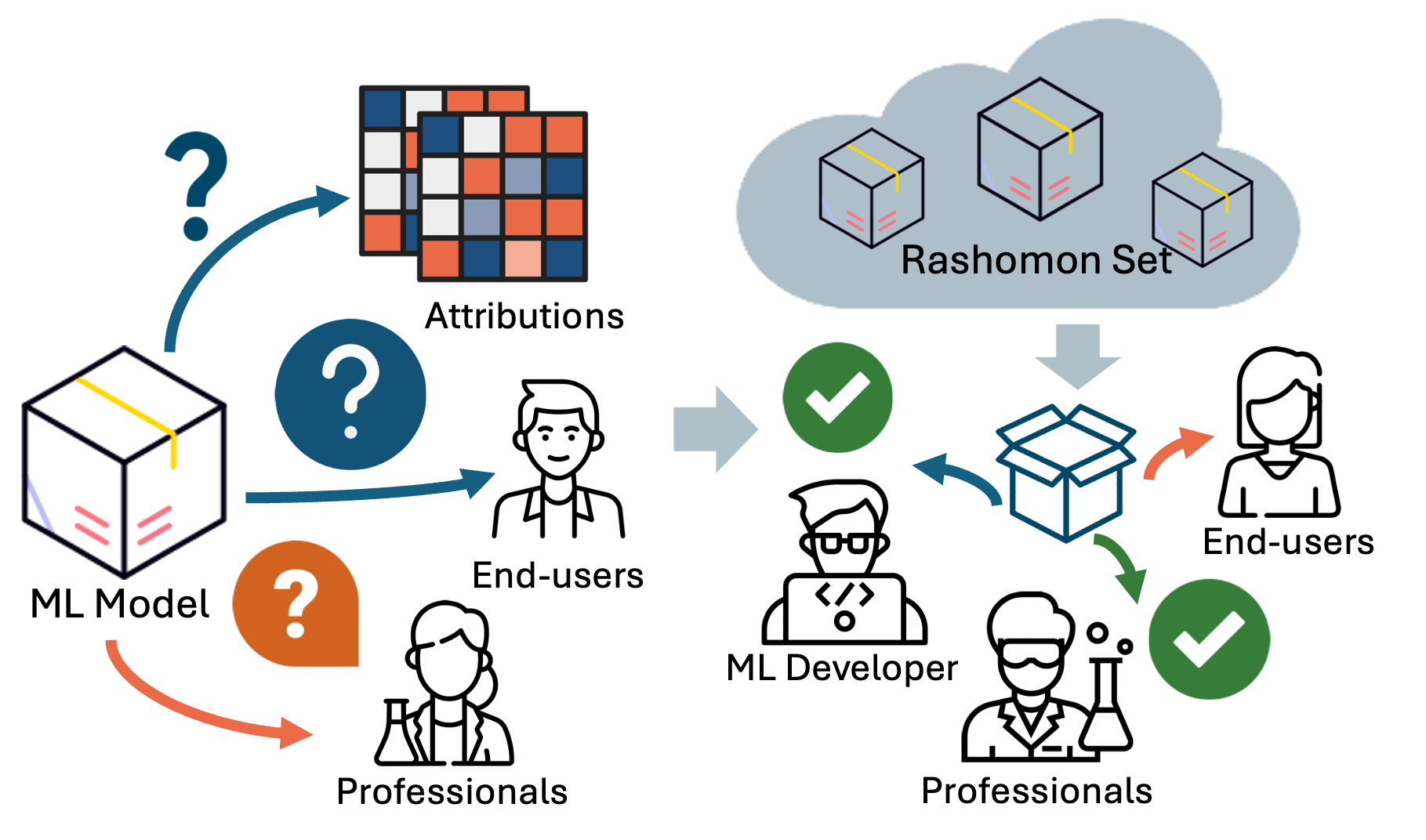}
    \caption{
    An illustration showing that an explanation relying on a single ML model cannot satisfy all stakeholders.
    % A practical problem in XAI: explanations relying on a single ML model can not satisfy all stakeholders
    }\label{fig:practical-problem}
\end{figure}

\noindent \textbf{Evaluating Explanations.}
Ideally, a good explanation should be \textit{understandable}, relatively \textit{faithful} to how the model works and \textit{useful} for the stakeholder’s end-goals \citep{liao2023ai}. Rankings can be used as a unified means for analyzing model behavior and provide a general foundation for understanding explanations across different scales. Let $\mathcal{O} (\cdot,\cdot)$ be \textit{Spearman’s rank correlation} and $\mathbf{r}^{k}$ the ranking preferred by stakeholder $k$.
We follow the general definitions of \textit{faithfulness} and \textit{plausibility} as \cite{agarwal2024faithfulness, jin2023xai, Sithakoul2024}:
\begin{itemize}
    \item \textbf{Faithfulness (\(\mathcal{A}_{\text{faith}}\)):} An explanation is faithful if it accurately reflects the reasoning process of the model.
    \[
    \mathcal{A}_{\text{faith}} = \mathcal{O}(\bfs{r}^{M}_{\varphi}, \bfs{r}^{M^*}_{\text{true}}),
    \]
    \item \textbf{Plausibility (\(\mathcal{A}_{\text{plaus}}\)):} An explanation is plausible if it aligns with stakeholder-grounded explanations (\(\bfs{r}^k\)).
    \[
\mathcal{A}_{\text{plaus}} = \mathcal{O}(\bfs{r}^{M}_{\varphi},\bfs{r}^k),
    \]
\end{itemize}
\begin{definition}[Stakeholder-Machine Agreement]
    We define Stakeholder-Machine Agreement (SMA) as the degree of alignment between the stakeholder need and the internal machine structure, denoted as: 
    \[\mathcal{A}_{\text{SMA}} = \mathcal{O}(\bfs{r}^k, \bfs{r}^{M^{*}}_{\text{true}}).\]
\end{definition}

\begin{lemma}[Faithfulness and Plausibility Trade‐Off]
If $\mathcal{A}_{\text{SMA}}\!<\!1$, no delivered explanation can simultaneously maximize $\mathcal{A}_{\text{faith}}$ and $\mathcal{A}_{\text{plaus}}$ (proof see Appendix Lemma 3.1).  
\end{lemma}

This divergence necessarily induces a tension between faithfulness and plausibility.

\begin{figure}[b!]
    \centering
    \includegraphics[width=.7\linewidth]{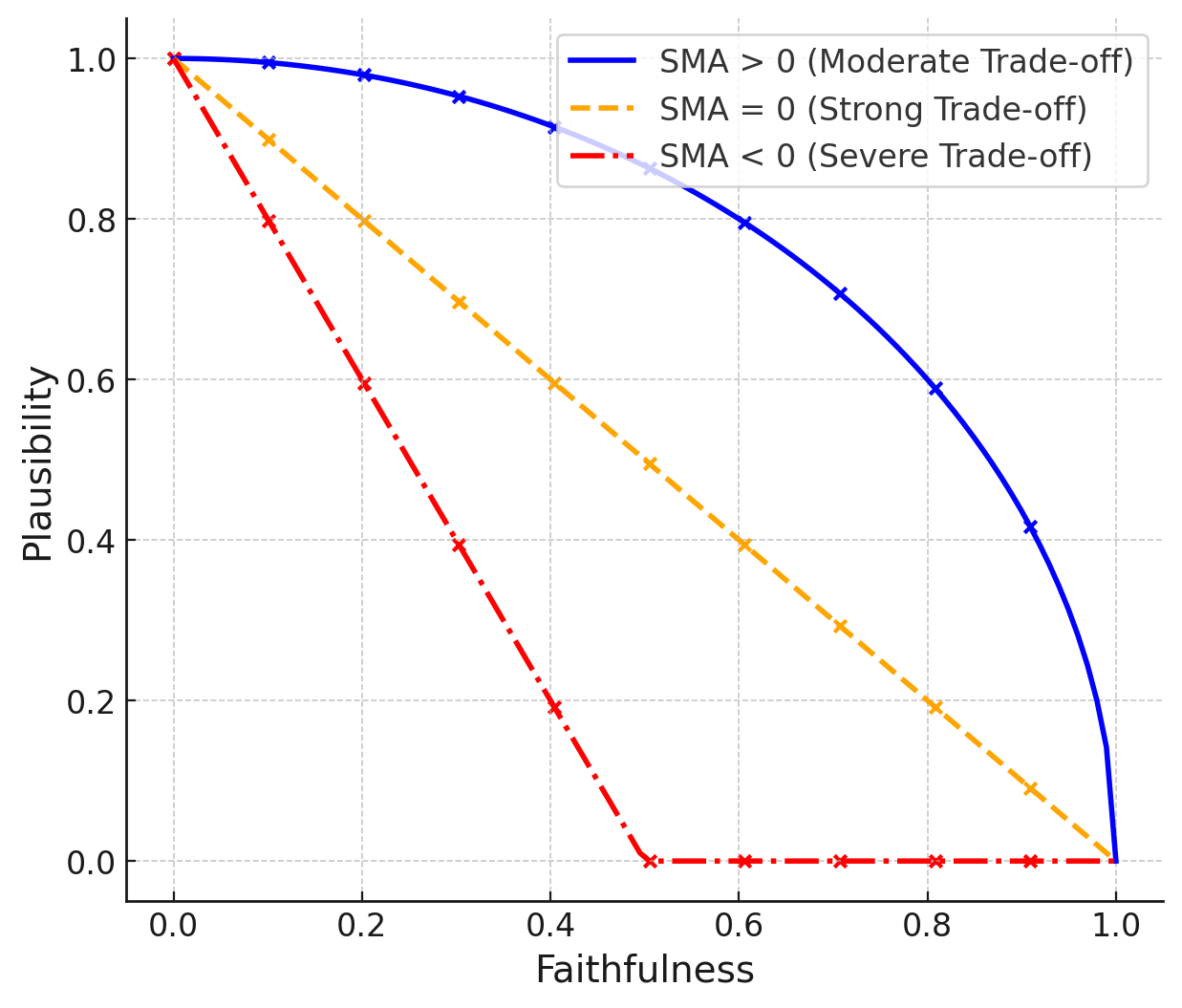}
    \caption{Pareto frontier of faithfulness and plausibility in practice with $\mathcal{A}_{\text{SMA}} > 0$, $\mathcal{A}_{\text{SMA}} = 0$, and $\mathcal{A}_{\text{SMA}} < 0$, indicating the strength of trade-off.}
    \label{fig:trade-off-f-and-p}
\end{figure}

\begin{figure*}[htp]
    \centering
    \includegraphics[width=.75\linewidth]{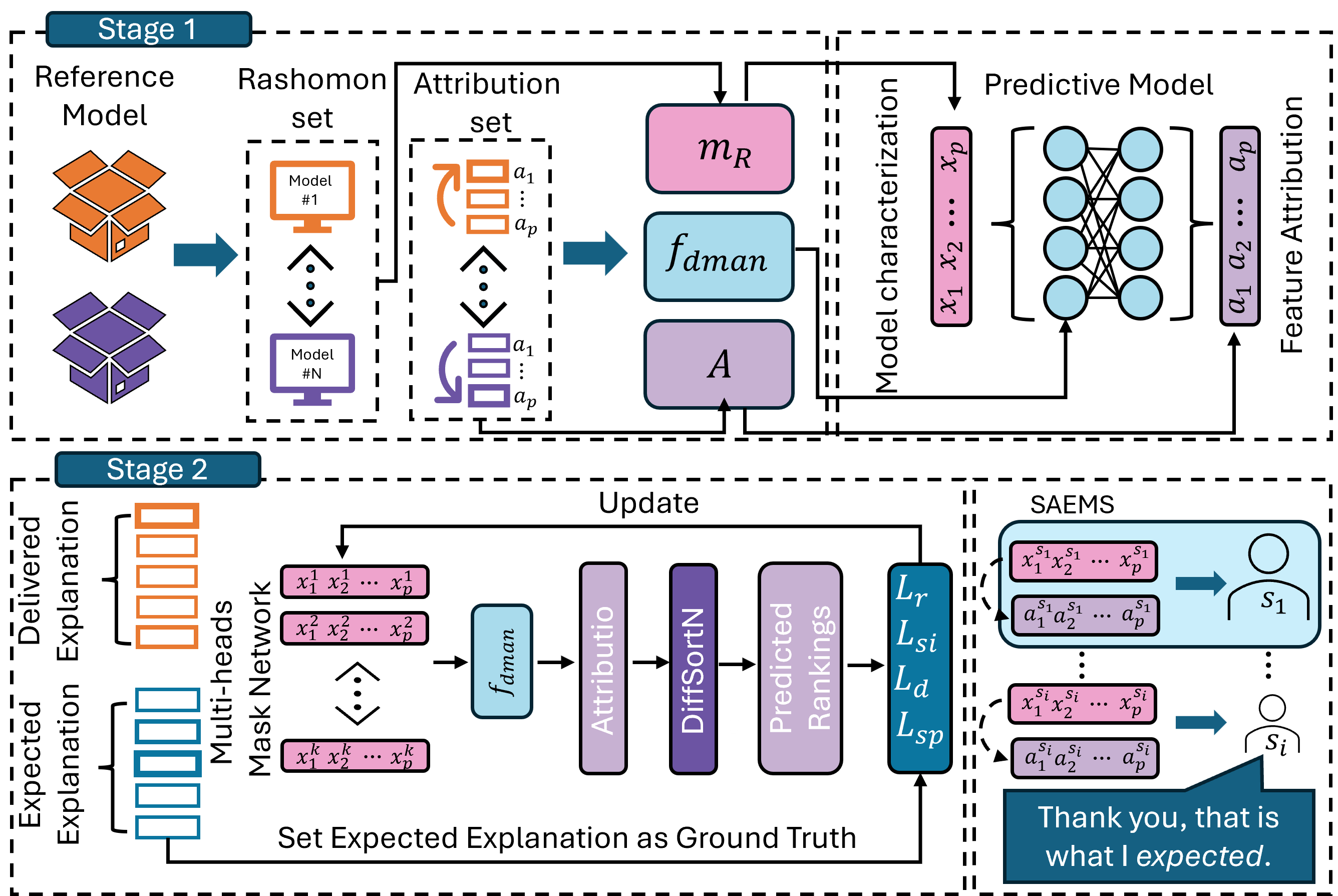}
        \caption{Overview of EXAGREE framework, illustrating the two-stage processes from top-left to bottom-right. Stage 1: Exploring Rashomon Set and Attribution Mapping; Stage 2: Identification of SAEMs under ranking supervision.}
    \label{fig:eef}
\end{figure*}

\begin{lemma}[Trade-Off Tension]\label{prop:A_sma}
    A higher $\mathcal{A}_{\text{SMA}}$ is desirable and necessary for faithful and plausible explanations.
\end{lemma}
\noindent
\textit{Proof.} Pareto frontier of faithfulness and plausibility \citep{agarwal2024faithfulness} in general practice is shown in Fig. \ref{fig:trade-off-f-and-p} (see proof in Appendix Lemma 3.2).

\begin{remark} 
    If and only if $\mathcal{A}_{\mathrm{SMA}} = 1$, there is an ideal situation in which the delivered explanation is both perfectly faithful (to the model) and perfectly plausible (to the user).
\end{remark}
\noindent
\textit{Proof.}  
    Spearman’s rank correlation attains its maximum value $1$ exactly when two rankings are identical  $\mathbf{r}^{k} \!=\! \mathbf{r}^{M}_{\text{true}}$, under which condition $\mathcal{A}_{\text{faith}}$ and $\mathcal{A}_{\text{plaus}}$ can be simultaneously optimized.

\noindent \textbf{Sources of Explanation Disagreement.} We denote the disagreement that arises from different choices of model and explanation methods as \textit{technical disagreement}, formulated as:
$$\exists M, M' \in \mathcal{M}, \varphi, \varphi' \in \Phi,  : \bfs{r}^{M}_{\varphi} \neq \bfs{r}^{M}_{\varphi'} \neq \bfs{r}^{M'}_{\varphi} \neq \bfs{r}^{M'}_{\varphi'}).$$
\textit{Stakeholder disagreement} arises when human stakeholders ($\mathcal{S}$) may preference different rankings based on their needs or knowledge and is given as: $$\exists k, l \in \mathcal{S}, k \neq l : \bfs{r}^k \neq \bfs{r}^l.$$

\paragraph{Stakeholder-centered Objective.}
In this work, we utilize technical disagreements as a strategy to identify the best model for a given stakeholder that balances faithfulness and plausibility. In the ideal case, there exists a model that satisfies the condition \(\exists M \in \mathcal{M}, \varphi \in \Phi: \bfs{r}^{M}_{\varphi} \simeq \bfs{r}^k\).
We formulate the problem as the following optimization task: 
% \begin{equation}
% \small
% \label{eq:optimization}
% \max_{M}  \quad \sum_{k \in \mathcal{S}} \mathcal{O}_{k}(\bfs{r}^k, \bfs{r}^{M}_{\varphi}), \quad \text{s.t.} \quad \Loss(M(\bfs{X}), \bfs{y}) \leq \tau,  
% \end{equation}
\begin{equation}
\label{eq:optimization}
\small
M_k^{*} \in 
\arg\max_{M} \;
\mathcal{O}_{k}\bigl(\mathbf{r}^k, \mathbf{r}^{M}_{\varphi}\bigr)
\quad \text{s.t.} \quad 
\Loss\!\left(M(\mathbf{X}), \mathbf{y}\right) \le \tau, 
\end{equation}
where $\tau$ is the performance threshold, as a foundation of stakeholder's trust and we refer to $M_k^{*}$ as a \emph{Stakeholder-Aligned Explanation Model (SAEM)} for stakeholder $k$
\cite{ortigossa2024explainable, senoner2024explainable, rogha2023explain}. 
The overall objective is to identify a set that contains each stakeholder and satisfies their needs.
\[
\mathcal{M}^{*}_{\mathcal{S}}
\;=\; \bigl\{ M_k^{*} \;:\; k \in \mathcal{S} \bigr\}.
\]

% \ale{QD: The maximization problem (2) is not well-defined if you maximizing for each $k$ independently. It would make more sense with a summation $\max_{M} \sum_k  \mathcal{O}_{k}(\bfs{r}^k, \bfs{r}^{M}_{\varphi})$. In the your current definition, you may have no solution, while with summation, you may have multiple solutions. }

\section{EXAGREE Framework}\label{sec:framework}

To achieve the objective in Eq. \eqref{eq:optimization}, we propose a framework that dynamically identifies models through an interactive process with the user. A Large Language Model (LLM)-based \textit{interface} enables users to express their preferences in natural language, removing the need for technical expertise. The framework \textit{efficiently} identifies the best alternative models available according to the trade-off between faithfulness and plausibility based on the user-provided feedback.
%, ensuring that explanations remain both meaningful and contextually relevant.
% In this work, we focus on exploring a set of well-performing models as the foundation of the framework, providing a technical search space. 
By setting the pre-defined performance threshold $\epsilon$ in Eq. \eqref{eq:grs}, the objective in Eq. \eqref{eq:optimization} is constrained. 

\noindent \textbf{Summary of EXAGREE.} 
EXAGREE is designed as an interactive model selection framework that aims to meet stakeholders' diverse needs in practice, as illustrated in Fig. \ref{fig:eef} and involves two main stages:
\begin{enumerate}[label=(\roman*),leftmargin=*,itemsep=1pt,topsep=1pt]
    \item Stage 1 (Sec. \ref{sec:stage1}) explores the Rashomon set and fits a \textit{Differentiable Mask-based Model to Attribution Network} (DMAN) that maps feature attributions from model characterizations for use in the next stage; 
    \item Stage 2 (Sec. \ref{sec:stage2}) identifies models that align with stakeholder needs within the Rashomon set via differentiable sorting and ranking optimization. 
\end{enumerate}

\subsection{Exploring Rashomon Set on ANY Model}\label{sec:stage1}
There are several algorithms to construct the Rashomon set for a given reference model \cite{dong2020exploring,hsu2022rashomon,zhong2022explainable, li2023exploring}. Since the model structure in practice is not fixed, we adopt the model-agnostic General Rashomon Subset Sampling algorithm (GRS) from \citeauthor{li2024practical}. We then use a permutation-based explanation method, Feature Importance Score (FIS), for attribution \cite{fisher2019all, dong2020exploring}.
% This approach  approximates the Rashomon set for any reference model $M$ and guarantees a fair and consistent comparison of explanations by using a model-agnostic approach.
FIS is a model-agnostic method inspired by model reliance and similar methods \cite{fisher2019all}, which measures the change in the loss by replacing the variable of interest with a new random independent variable. 
This is denoted by:
\begin{equation}\label{eq:fis}
    \varphi_{i}(M) = \E[\Loss(M(\bfs{X}_{\setminus i}), \bfs{y})] - \E[\Loss(M(\bfs{X}), \bfs{y})],
\end{equation}
where $\bfs{X}_{\setminus \bfs{i}}$ is the input matrix that is replaced by an independent variable. In practice, we usually permute a feature of interest multiple times to achieve a similar measurement \citep{datta2016algorithmic}.

The core idea of the GRS sampling method is that every model in the Rashomon set can be replaced by a mask concatenating into the reference model, defined as:
\begin{align}\label{eq:mask_based_rset}
& \forall M \in \mathcal{M} \ \text{with} \ \E[\Loss(M(\bfs{X}), \bfs{y})], \quad \nonumber
\\ & \exists m  \, \text{s.t.} \ \E[\Loss(M \circ m(\bfs{X}), \bfs{y})] \leq \E[\Loss(M^{*}(\bfs{X}), \bfs{y})] + \epsilon,
\end{align}
where $m \in \mathbb{R}$ is the concatenating layer and the method enables us to sample \textit{masks} as characterizations of each model in the Rashomon set. 

Taking them together, for each model in the Rashomon set, characterized by $m$, there will be a corresponding feature attribution list, denoted as:
\begin{equation}\label{eq:attribution-list}
    \forall M \in \mathcal{M}_{\epsilon}, \bfs{a}^{M}_{\varphi} =  
\{\varphi_i(M) \mid i \in p \}.
\end{equation}

Given the sampled Rashomon set $\mathcal{M}_{\epsilon}$, the optimization in
Eq.~\eqref{eq:optimization} is instantiated for each stakeholder $k \in \mathcal{S}$ with $\mathbf{r}^{M}_{\varphi}$ computed 
via Eq.~\eqref{eq:attribution-list}:
\begin{equation}
\label{eq:optimization_simplified}
M_k^{*} \;\in\;
\arg\max_{M \in \mathcal{M}_{\epsilon}}
\; \mathcal{O}_k\!\left(\mathbf{r}^{M}_{\varphi},\, \mathbf{r}^{k}\right).
\end{equation}

\paragraph{Constrained Sorting and Ranking of Attributions in the Rashomon set}

\begin{figure}[]
    \centering
    \includegraphics[width=\linewidth]{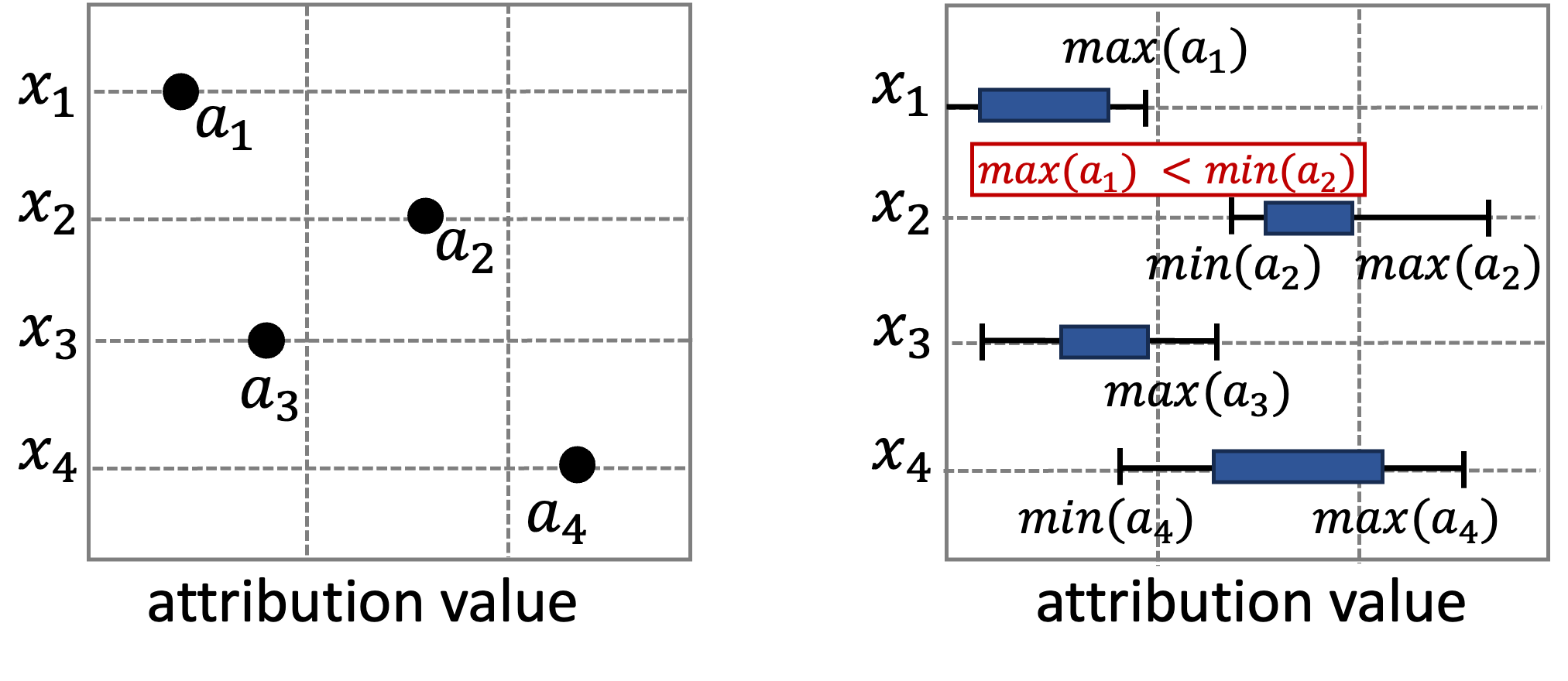}
    \caption{An illustration of feature attribution distributions from a single model (\textit{left}) and a model space (\textit{right}), where the ordering
    $a_{(1)} \succ a_{(2)}$ can never occur.}
    \label{fig:rankings}
\end{figure}

However, this optimization is constrained within the Rashomon space, as not all attribution swaps are feasible. This constraint distinguishes our problem from conventional ranking problems, as feature attributions from a single model and a model space are shown in Fig. \ref{fig:rankings}.
\begin{lemma}\label{lemma:impossible-swap_I}
    In a Rashomon set $\mathcal{M}_\epsilon$, not all pairwise attribution swaps are possible.
\end{lemma}
\begin{proof}
    Recent Rashomon-related studies have shown that feature attribution range or model class reliance is not unlimited \citep{fisher2019all, li2023exploring, hsu2022rashomon, NEURIPS2022_5afaa8b4, li2024practical}.
    Consider a scenario where the feature attribution matrix from the Rashomon set is: $$[min(\bfs{a}_i), max(\bfs{a}_i)]^{p}_{i=1},$$ 
    where we specify $\max(a_1) < \min(a_2)$, as illustrated in the middle panel of Fig. \ref{fig:rankings}. The swap between $a_{(1)}$ and $a_{(2)}$ is not possible within this Rashomon set in this case.
\end{proof}

% Our primary objective of minimizing disagreement between model explanations and stakeholder expectations naturally leads to an end-to-end ranking optimization problem. The Rashomon set concept allows us to transform single attribution values into ranges, as illustrated in Fig. \ref{fig:rankings}, enabling us to find a potential model with stakeholder-expected rankings from $\mathcal{M}_\epsilon$. 
% \begin{proof}
%     Assume, for the sake of contradiction, that any pairwise swap of attributions is possible within the Rashomon set. However, this contradicts the fact that the feature attribution range or model class reliance is not unlimited, as demonstrated in prior Rashomon-related theoretical and empirical studies \citep{fisher2019all, li2023exploring, hsu2022rashomon, NEURIPS2022_5afaa8b4, li2024practical}.
%     Consider a scenario where the feature attribution matrix from the Rashomon set is: $$[min(\bfs{a}_i), max(\bfs{a}_i)]^{p}_{i=1},$$ 
%     where we specify $\max(a_1) < \min(a_2)$, as illustrated in the middle panel of Fig. \ref{fig:rankings}. In this case, the swap $a_{(1)}$ between $a_{(2)}$ is impossible within this Rashomon set.
% \end{proof}
\begin{proposition}%\label{prop:not-always-satisfied}
    Based on the previous lemma \ref{lemma:impossible-swap_I}, we can conclude that there does not always exist a model within the Rashomon set $\mathcal{M}_\epsilon$ that satisfies a stakeholder's expectation (see proof in Appendix Lemma 3.4).
\end{proposition}
% \begin{proof}
%     As shown in the lemma, consider an expected ranking is $\mathbf{r}^* = (a_{(1)} \succ a_{(2)} \succ \cdots \succ a_{(p)})$, and the feature attribution matrix from the Rashomon set exhibits the property $\max(a_1) < \min(a_2)$, there does not exist a model $M \in \mathcal{M}_\epsilon$ that can satisfy the condition $a_{(1)} \succ a_{(2)}$, formulated as:
%     \begin{equation*}
%         \nexists M \in \mathcal{M}_\epsilon : a_{(1)} \succ a_{(2)}.
%     \end{equation*}
%     This means that the stakeholder's expected ranking $\mathbf{r}^*$ is inaccessible within this Rashomon set $\mathcal{M}_\epsilon$, as the necessary pairwise attribution swaps to achieve the desired ranking order is not feasible.
% \end{proof}
Given this limitation, our approach shifts towards finding models within the Rashomon set that most align with stakeholder needs, even if it doesn't perfectly match them, motivating the development of our multi-head framework.

\begin{table*}[t!]
\centering
\scriptsize
\renewcommand{\arraystretch}{1.5}
\setlength{\tabcolsep}{3pt} % Adjust column padding

\begin{adjustbox}{width=\textwidth}
\begin{tabular}{p{4.2cm}p{1.5cm}p{1.5cm}p{1.5cm}p{3.5cm}p{4.5cm}}
% {p{4.2cm}p{2cm}p{2.2cm}p{2cm}p{3.5cm}p{4.5cm}}
\toprule
\begin{tabular}[t]{@{}l@{}}\textbf{General Context in Practice}\\ (1) Delivered Model $M$ \\ (2) Arbitrary Needs $\bfs{r}^k_\alpha$  \\ (3) Delivered Explanation $\bfs{r}^{M}_{\varphi}$\end{tabular}  &
\textbf{Stakeholder Grounded Explanation ($\bfs{r}^k_\alpha$)} & 
\textbf{Machine Grounded Explanation ($\bfs{r}^{M_{\mathcal{I}}}_{\text{true}}$)} &
\textbf{Delivered Explanation ($\bfs{r}^{M}_{\varphi}$)}

& \begin{tabular}[t]{@{}l@{}}\textbf{Stakeholder-Centered Agreement} \\ $\mathcal{A}_{\text{plaus}} = \mathcal{O}(\bfs{r}^k_\alpha, \bfs{r}^{M}_{\varphi})$ \\ $\mathcal{A}_{\text{SMA}} = \mathcal{O}(\bfs{r}^k_\alpha, \bfs{r}^{M_{\mathcal{I}}}_{\text{true}})$ \\ $\mathcal{A}_{\text{faith}} = \mathcal{O}(\bfs{r}^{M}_{\varphi}, \bfs{r}^{M_{\mathcal{I}}}_{\text{true}})$\end{tabular}
&\begin{tabular}[t]{@{}l@{}}\textbf{General Objective}\\(1) Plausibility \\ (2) Faithfulness\end{tabular} \\ 
\midrule

Case 1-1: Black-box model (ANN), \textit{constant need} and diverse explanation method 
& $\bfs{r}^{\text{LR}}_{\text{true}}$ 
& - 
& $\bfs{r}^{\text{ANN}}_{\text{post}}$ 
& \raggedright $\mathcal{A}_{\text{plaus}} = \mathcal{O}(\bfs{r}^{\text{LR}}_{\text{true}}, \bfs{r}^{\text{ANN}}_{\text{post}})$ \newline $\mathcal{A}_{\text{faith}}$ and $\mathcal{A}_{\text{SMA}}$ unavailable 
& Identify models that provide higher plausibility on the same explanation method from $\mathcal{M}_{\epsilon}$\\
\hline

Case 1-2: Black-box model (ANN), \textit{constant need} and interpretable proxy (DT)
& $\bfs{r}^{\text{LR}}_{\text{true}}$ 
& $\bfs{r}^{\text{DT}}_{\text{true}}$
& $\bfs{r}^{\text{DT}}_{\text{true}}$ 
& \raggedright $\mathcal{A}_{\text{plaus}} = \mathcal{O}(\bfs{r}^{\text{LR}}_{\text{true}}, \bfs{r}^{\text{DT}}_{\text{true}})$  
% $\mathcal{A}^{\text{proxy}}_{\text{faithfulness}}$ and $\mathcal{A}^{\text{proxy}}_{\text{SMA}}$ available 
& Proxy illustrates plausibility optimization independent of ANN internals\\
\hline
Case 2-1: White-box model (LR), 
\textit{constant need}, diverse explanation method
& $\bfs{r}^{\text{LR}}_{\text{true}}$ 
& $\bfs{r}^{\text{LR}}_{\text{true}}$ 
& $\bfs{r}^{\text{LR}}_{\text{post}}$
& \raggedright $\mathcal{A}_{\text{SMA}} = 1$, $\mathcal{A}_{\text{faith}} = \mathcal{A}_{\text{plaus}} = \mathcal{O}(\bfs{r}^{\text{LR}}_{\text{post}}, \bfs{r}^{\text{LR}}_{\text{true}})$ 
& Demonstrate the constant $\mathcal{A}_{\text{SMA}}=1$ and improve both faithfulness and plausibility \\
\hline
Case 2-2: White-box model (LR), arbitrary needs, diverse explanation method
& $\bfs{r}^k_\alpha$ 
& $\bfs{r}^{\text{LR}}_{\text{true}}$ 
& $\bfs{r}^{\text{M}}_{\text{post}}$ 
& \raggedright $\mathcal{A}_{\text{plaus}} = \mathcal{O}(\bfs{r}^k_\alpha, \bfs{r}^{\text{M}}_{\text{post}})$ $\mathcal{A}_{\text{faith}} = \mathcal{O}(\bfs{r}^{\text{M}}_{\text{post}}, \bfs{r}^{\text{LR}}_{\text{true}})$
& Demonstrate $\mathcal{A}_{\text{SMA}} \leq 1$ when needs vary; study the trade-off between $\mathcal{A}_{\text{faith}}$ and $\mathcal{A}_{\text{plaus}}$ \\
\bottomrule
\end{tabular}
\end{adjustbox}
\caption{\label{tab:exp-setup} Controlled Experimental Setup for Evaluating Stakeholder-Centered Explanation Agreement. To measure the relationship among $\mathcal{A}_{\text{faith}}$, $\mathcal{A}_{\text{plaus}}$, and $\mathcal{A}_{\text{SMA}}$, we ensure that at least one variable remains \textit{fixed} in each case. }
\end{table*}

\paragraph{Differentiable Model Attribution Network (DMAN)}
To enable efficient optimization and allow \textit{gradient-based} ranking optimization, a differentiable mapping from masks to feature attributions is required.
We propose the DMAN ($f_{\text{dman}}$) as a surrogate model that bridges the gap between models in the Rashomon set and their feature attributions. DMAN is a neural network trained to approximate the relationship between masks (representing models in the Rashomon set) and their corresponding attributions. The training process uses a dataset $\D_{att}=\{\bfs{m}_{\R}, \bfs{A}\}$, where $\bfs{m}_{\R}$ are masks and $\bfs{A}$ are corresponding attributions. The parameter optimization of the network is expressed as:

\begin{equation}
    f^*_{\text{dman}, \theta} = \underset{\theta \in \Theta}{\operatorname{arg\,min}} \, \mathcal{L}_{\text{MSE}}(f_{\text{dman}, \theta}, \mathcal{D}_{\text{att}})
\end{equation}

While DMAN provides an approximation as a surrogate model, its accuracy is crucial for the overall framework. To ensure reliability, we calculate actual attributions when evaluating the final results. 

This stage allows us to efficiently utilize the Rashomon set for attribution prediction while maintaining an end-to-end differentiable pipeline for further optimization.

\subsection{SAEM Identification}\label{sec:stage2}
We search the Rashomon set to identify SAEMs, predictive models whose expected explanations optimally balance plausibility and faithfulness. The final stage of differentiable optimization involves mapping feature attributions to ranking targets, e.g., $\bfs{r}^{M}_{\varphi}$, as defined in Eq.~\eqref{eq:attribution-list}.

\noindent \textbf{Ranking Supervision and Correlation Metric.} We employ monotonic differentiable sorting networks in our framework from the work of \cite{petersen2022monotonic}. This network, denoted as $f_{\text{diffsort}}$, enables ranking supervision where the ground truth order of features is known while their absolute values remain unsupervised. Additionally, Spearman's rank correlation provides differentiability in Eq. \eqref{eq:optimization_simplified} \citep{dodge2008concise,petersen2022monotonic, huang2022relational}. The correlation for a specific ranking from stakeholder $k$ can be calculated as: 
\begin{equation}
    \mathcal{O}^{k}(\bfs{r}^{M}_{\varphi}, \bfs{r}^k)=\frac{\operatorname{Cov}(f_{\text{diffsort}}(|\bfs{a}^{M}_{\varphi}|), \bfs{r}^k)}{\operatorname{Std}(\bfs{r}^{M}_{\varphi})\operatorname{Std}(\bfs{r}^k)}.
\end{equation}

It's important to note that feature attributions have directions that do not necessarily represent their strength. 
To address this, we use the absolute value of attributions in the correlation calculation, ensuring both positive and negative importances are appropriately accounted for in the ranking. 
In practice, stakeholders may or may not require information about the direction of feature importance. To accommodate this variability, we incorporate a sign loss in our optimization process when applicable. Loss function details are discussed in the following section.

\noindent \textbf{Multi-heads Architecture.} 
From Proposition 3.2, the uncertainty of finding a model perfectly matching stakeholder-expected rankings can be mitigated by a multi-head architecture.
For a single stakeholder, we are motivated by the following lemma:
\begin{lemma}
    For a given target ranking, there may exist multiple distinct rankings that achieve the same Spearman’s rank correlation. These alternatives can reflect different trade-offs between faithfulness and plausibility, depending on stakeholders' needs and alignment priorities. (See proof in Appendix Lemma 3.5).
\end{lemma}

% When considering multiple stakeholders in practice, the multi-head architecture becomes critical. This is motivated by an important observation:
% \begin{proposition}\label{prop:probability_main}
%     The increase in disagreement among stakeholders leads to greater opportunity to find a more faithful model (proof see Appendix \ref{prop:probability}), shown as:
%     \begin{equation}
%         \mathbb{P}(\exists M^* \in \mathcal{M}, \mathcal{O}(\bfs{r}^{M}_{\varphi}, \bfs{r}^j) < \mathcal{O}(\bfs{r}^{M^*, \varphi}, \bfs{r}^j)) \propto  (1-\mathcal{O}(\bfs{r}^{i}, \bfs{r}^{j})),
%     \end{equation}
%     where $1-\mathcal{O}(\bfs{r}^{i}, \bfs{r}^{j})$ is the disagreement between two stakeholders $i$ and $j$ and $i \neq j$.
% \end{proposition}
Consequently, we integrate multiple heads into the architecture, each corresponding to a potential SAEM. By integrating the above components $f_{\text{dman}}$ and $f_{\text{diffsort}}$ into the architecture, our objective function is reformulated to minimize negative ranking correlation across all heads in a \textit{Multi-heads Mask Network} (MHMN):
\begin{equation*}
    \min_{\Theta} \Loss_{\text{rank}} = \min_{\Theta} \sum_{j=1}^{h} \min_{M_j \in \mathcal{M}} -\mathcal{O}(\bfs{r}^{M_j, \varphi}, \bfs{r}^*),
\end{equation*}
where $\Theta$ represents the set of parameters for all $h$ heads and $\bfs{r}^*$ is the target ranking. 

To ensure that our multi-head architecture produces meaningful and diverse solutions while respecting stakeholder input, we introduce several key constraints: \textit{attribution direction} ($\Loss_{\text{sign}}$), \textit{sparse constraint} ($\Loss_{\text{sparsity}}$) and \textit{diverse constraint} ($\Loss_{\text{diversity}}$) (see details in Appendix Sec. 4). The overall objective function is reformulated as:
\begin{equation}
    \min_{\Theta} (\Loss_{\text{rank}} + \mathcal{L}_{\text{sign}} + \lambda_1 \mathcal{L}_{\text{sparsity}} + \lambda_2 \mathcal{L}_{\text{diversity}}),
\end{equation}
where $\lambda_1$ and $\lambda_2$ are hyperparameters that control the weight of the sparsity and diversity losses. This multi-head network is trained by propagating the error backward through the surrogate network and the sorting network. The algorithm is provided as pseudocode in the Appendix Algorithm 1.

\section{Experiment \& Discussion}\label{sec:experiments}

\paragraph{Experimental Objective} We design experiments to explore how faithfulness, plausibility, and SMA interact under various practical conditions, without compromising predictive accuracy, aiming to:
\begin{enumerate}[label=(\roman*),itemsep=1pt,topsep=1pt]
    \item Evaluate faithfulness and plausibility improvements for delivered black-box models (cases 1-1 and 1-2);
    \item Improve faithfulness and plausibility for white-box models in the special case of $\mathcal{A}_{\text{SMA}}=1$ (case 2-1);
    \item Navigate the trade-off between faithfulness and plausibility in general settings $\mathcal{A}_{\text{SMA}}\leq1$ (Case 2-2). 
\end{enumerate}
Table \ref{tab:exp-setup} presents a detailed overview of the controlled experimental setup along with the corresponding objectives. For each case, we further evaluate whether the generated explanations adversely affect predictive performance or subgroup fairness.

% Does improved faithfulness and plusibility statistically significant?

\begin{figure*}[htp]
    \centering
    \includegraphics[width=\linewidth]{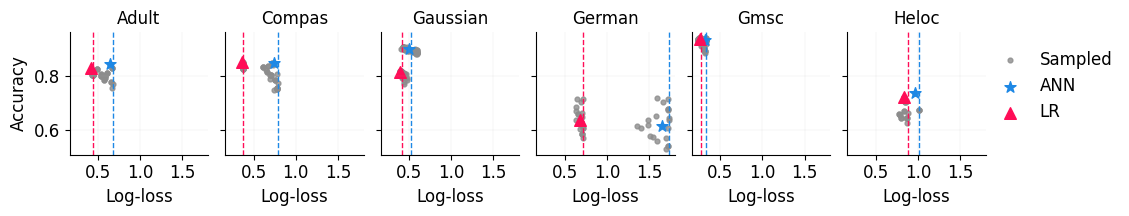}
    \caption{Performance of sampled models (gray) across all datasets from two canonical backbones: ANN (blue) and LR (red). The vertical dashed line denotes the log-loss threshold $\epsilon$ = 0.05 used to define the Rashomon set.}
    \label{fig:sampling-model}
\end{figure*}

\noindent \textbf{Benchmark \& Tasks.}
We benchmark our framework on the \textit{OpenXAI} suite \cite{agarwal2022openxai}, which offers six public datasets spanning from economic, finance, criminal-justice, healthcare and synthetic domains (details in Appendix Table 3). Each dataset comes with two pre-trained models: (i) a logistic regressor (LR) whose weight vector provides the stakeholder-grounded reference ranking $\mathbf{r}^{\text{LR}}_{\text{true}}$, and  
(ii) a higher-capacity artificial neural network (ANN) with post-hoc explanations $\mathbf{r}^{\text{ANN}}_{\text{post}}$.

\noindent \textbf{Baselines.} We compare \textsc{EXAGREE} with four widely used, model-agnostic \emph{global} explanation methods:  
(1) Random (uniform feature ranking) as a baseline;  
(2) SHAP values \cite{lundberg2017unified};  
(3) FIS, a permutation-importance variant \cite{fisher2019all,li2024practical}; and  
(4) a surrogate decision tree (DT) fitted to the dataset for approximating black-box models $\mathbf{r}^{\text{DT}}_{\text{true}}$ \cite{molnar2022}.  
All baselines operate on the same underlying LR/ANN outputs to ensure comparability. 

\noindent \textbf{Evaluation Metrics \& Protocol.} 
We report three \textit{overall} agreement metrics $\mathcal{A}_{\text{plaus}}$, $\mathcal{A}_{\text{faith}}$, and $\mathcal{A}_{\text{SMA}}$ for each scenario, which agreement corresponds to the ranking correlation (RC) score in OpenXAI. We also follow the OpenXAI protocol by reporting seven complementary \textit{top-$k$} feature agreement metrics: FA, RA, SA, SRA, PRA, PGI, PGU, and one fairness score (see Appendix Table 1 for more details). All metrics are averaged over 5 fixed seeds and $\mathbf{\pm}$\emph{95\%} confidence interval is reported.

% All metrics are averaged over 5 fixed random seeds (0–5). 
% Significance versus the strongest post-hoc baseline (SHAP) is tested with a paired two-sample $t$‐test; $p$ values are Holm–Bonferroni-corrected per metric ($\alpha=0.05$).

To identify the stakeholder-aligned model, the Rashomon set sampling is controlled with $\epsilon = 0.05$ on the reference model ($f_{\text{ANN}}$ or $f_{\text{LR}}$), ensuring models in the set have at most a 5\% performance drop, with further \textit{ablation studies} detailed in Appendix Sec. 5. We then optimize $f_{\text{dman}}$ using the sampled Rashomon set for stage 2. The final step involves an MHMN with $f_{\text{diffsort}}$ training to optimize a combined loss over faithfulness and plausibility, as formalized in Sec.~\ref{sec:framework}.

% Across eight benchmarks, EXAGREE improves faithfulness (FA) by Δρ = +0.18 ± 0.04 and plausibility (PRA) by Δρ = +0.22 ± 0.03 on average, compared to SHAP (paired t, Holm–Bonferroni corrected, p < 0.05 on 7/8 datasets). See Table 2 for per-dataset results and Figure 3 for the FA–PRA Pareto front with 95 % error bars.

%--------------------------------------------------
\begin{table}[t!]
\centering
\small       
\setlength{\tabcolsep}{2pt}
\renewcommand{\arraystretch}{1.15}
\begin{tabular}{p{1.9cm}p{0.7cm}p{0.7cm}p{0.7cm}p{0.7cm}p{1.2cm}p{1cm}}
\toprule
\small
\multirow{2}{*}{Dataset}
& \multicolumn{5}{c}{Delivered Explanation (Black-box)} & \multirow{2}{*}{Avg.$\Delta$} \\
\cmidrule(lr){2-6} & 
$\bfs{r}^{\text{ANN}}_{\text{Random}}$ & $\bfs{r}^{\text{ANN}}_{\text{SHAP}}$ & $\bfs{r}^{\text{ANN}}_{\text{FIS}}$ & $\bfs{r}^{\text{DT}}_{\text{true}}$ & $\bfs{r}^{\text{SAEM}}_{\text{FIS}}$ \\
\midrule
Adult            & 0.18 & 0.75 & 0.84 & 0.83 & \textbf{0.85}$^{\pm0.02}$ & +0.20 $\uparrow$ \\
Gaussian     &-0.09 & 0.65 & 0.79 & 0.60 & \textbf{0.80}$^{\pm0.02}$ & +0.33 $\uparrow$\\
COMPAS           & 0.79 &-0.04 & 0.86 & \textbf{0.95} & 0.94$^{\pm0.01}$ & +0.31 $\uparrow$\\
German Credit    &-0.15 &-0.02 & 0.06 & 0.21 & \textbf{0.25}$^{\pm0.06}$ & +0.23 $\uparrow$\\
HELOC            &-0.30 &-0.10 & 0.57 & 0.56 & \textbf{0.60}$^{\pm0.04}$ & +0.42 $\uparrow$\\
GMSC             &-0.24 & 0.36 & 0.62 & 0.75 & \textbf{0.80}$^{\pm0.02}$ & +0.43 $\uparrow$\\
\bottomrule
\end{tabular}
\caption{\label{tab:results-black-box}Plausibility score $\mathcal{A}_{\text{plaus}}\;(\uparrow)$ in the black-box setting (cases 1-1 and 1-2) analysis across all datasets. Each entry is the Spearman~$\rho$ between the delivered explanation (${\bf r}^{\mathrm{ANN}}_{\text{post}}$ for post-hoc methods; $\bfs{r}^{\text{DT}}_{\text{true}}$ for the surrogate DT)
and the stakeholder-grounded reference (${\bf r}^{\mathrm{LR}}_{\text{true}}$).
Higher is better; \textbf{bold} indicates the best per row; $\Delta$ (Gain) shows SAEM's average improvements over the baselines.}
\end{table}
\begin{table}[t]
\centering
\small
\setlength{\tabcolsep}{2pt}
\renewcommand{\arraystretch}{1.15}
\begin{tabular}{p{1.9cm}p{0.7cm}p{0.7cm}p{0.7cm}p{0.7cm}p{1.2cm}p{1cm}}
\toprule
\multirow{2}{*}{\textbf{Dataset}}
& \multirow{2}{*}{$\mathcal{A}_{\text{SMA}}$} 
& \multicolumn{4}{c}{Delivered Explanation} 
& \multirow{2}{*}{Avg.\,$\Delta$} \\
\cmidrule(lr){3-6}
& & $\bfs{r}^{\text{LR}}_{\text{Random}}$ & $\bfs{r}^{\text{LR}}_{\text{SHAP}}$ & $\bfs{r}^{\text{LR}}_{\text{FIS}}$ & $\bfs{r}^{\text{SAEM}}_{\text{FIS}}$ & \\ 
\midrule
Adult           & 1.00 &  0.18 & 0.48 & 0.82 & \textbf{0.90}$^{\pm0.03}$ & +0.44 $\uparrow$ \\
Gaussian       & 1.00 & -0.09 & 0.96 & 0.97 & \textbf{0.99}$^{\pm0.01}$ & +0.39 $\uparrow$ \\
COMPAS          & 1.00 &  0.79 & 0.11 & 0.86 & \textbf{0.98}$^{\pm0.02}$ & +0.41 $\uparrow$ \\
German Credit   & 1.00 & -0.15 & 0.49 & \textbf{0.70} & \textbf{0.70}$^{\pm0.05}$ & +0.38 $\uparrow$ \\
HELOC           & 1.00 & -0.30 & 0.57 & 0.68 & \textbf{0.69}$^{\pm0.05}$ & +0.42 $\uparrow$ \\
GMSC            & 1.00 & -0.24 & 0.31 & 0.67 & \textbf{0.70}$^{\pm0.06}$ & +0.51 $\uparrow$ \\
\bottomrule
\end{tabular}
\caption{\label{tab:results-white-box-2-1}Faithfulness score $\mathcal{A}_{\text{faith}} = \mathcal{A}_{\text{plaus}}$ in the white-box setting (case 2-1), where $\mathcal{A}_{\text{SMA}}{=}1$. Each entry is Spearman~$\rho$ between the delivered explanation and the stakeholder-grounded reference $\mathbf{r}^{\text{LR}}_{\text{true}}$. Higher is better; \textbf{bold} = best in row. $\Delta$ (Avg. Gain) shows SAEM's average improvement over the three baselines.}
\end{table}
%--------------------------------------------------

\begin{figure*}[htp]
    \centering
    \includegraphics[width=\linewidth]{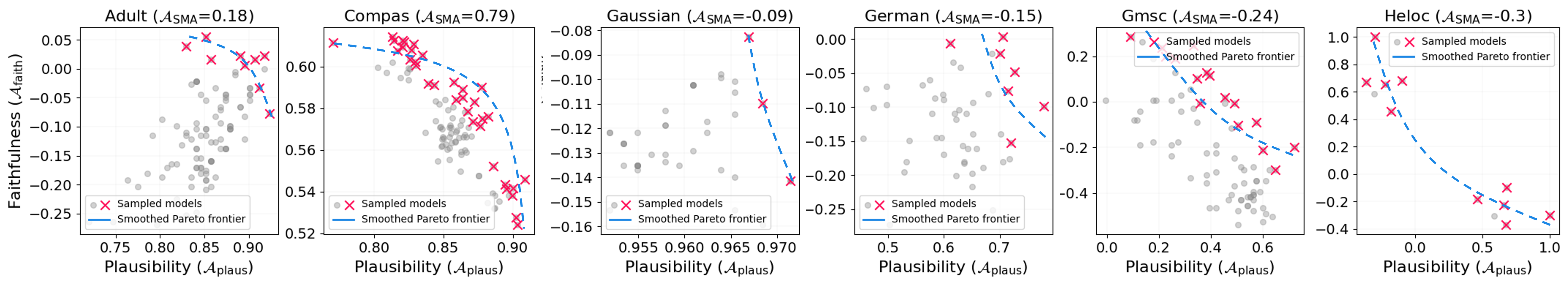}
    \caption{Faithfulness–plausibility trade-offs in the general setting (Case 2‑2), where $\mathcal{A}_{\text{SMA}} < 1$. Each subplot shows sampled models (grey dots and red marks) and a smoothed Pareto frontier (blue dashed) on a given dataset. Titles include the corresponding dataset and $\mathcal{A}_{\text{SMA}}$ value for the random needs $[-0.3, 0.79]$ from stakeholders.}
    \label{fig:pareto-frontier}
\end{figure*}

\paragraph{Improving Plausibility for Black-box Settings.}

We begin by considering a common, practical black-box scenario, where a stakeholder may express an arbitrary need, while the internal workings of the model remain inaccessible. 
Since it is infeasible to exhaustively represent all stakeholder preferences, we adopt a reproducible proxy, where the machine-grounded feature attribution from the pre-trained LR ($\bfs{r}^{\text{LR}}_{\text{true}}$) serves as a \textit{constant stakeholder-grounded explanation} and the delivered black-box model is the pre-trained ANN with unknown logic. 
We then compute $\mathcal{A}_{\text{plaus}}$ for baseline post-hoc methods applied in ANN (${\bf r}^{\mathrm{ANN}}_{\text{post}}$) by correlating their attributions with the fixed stakeholder need ($\bfs{r}^{\text{LR}}_{\text{true}}$), and report the gains achieved by SAEM in Table~\ref{tab:results-black-box}.

\noindent \textbf{Findings.} Plausibility varies widely across methods and datasets (e.g., SHAP vs.\ Surrogate‑DT on COMPAS and HELOC), underscoring the risk of random explainer selection in practice \cite{krishna2022disagreement}. Because every method is applied to the same ANN, simply swapping explainers \emph{cannot guarantee} higher $\mathcal{A}_{\text{plaus}}$ and may even mislead stakeholders (e.g., SHAP on German Credit).
Instead, by optimizing within the Rashomon set towards specific needs, we identify SAEMs that improve \textit{$\mathcal{A}_{\text{plaus}}$ by up to +0.43 $\rho$}, dominating common post-hoc baselines. Performance as a function of the top‑$k$ features (percentage) is shown for Adult in Fig.~\ref{fig:adult_income_comparison_k}; curves for the remaining datasets appear in the Appendix Fig.~3 and Fig.~4.
\begin{figure}[b!]
\centering
    \begin{subfigure}[t]{0.23\textwidth}
        \centering
        \includegraphics[width=\textwidth]{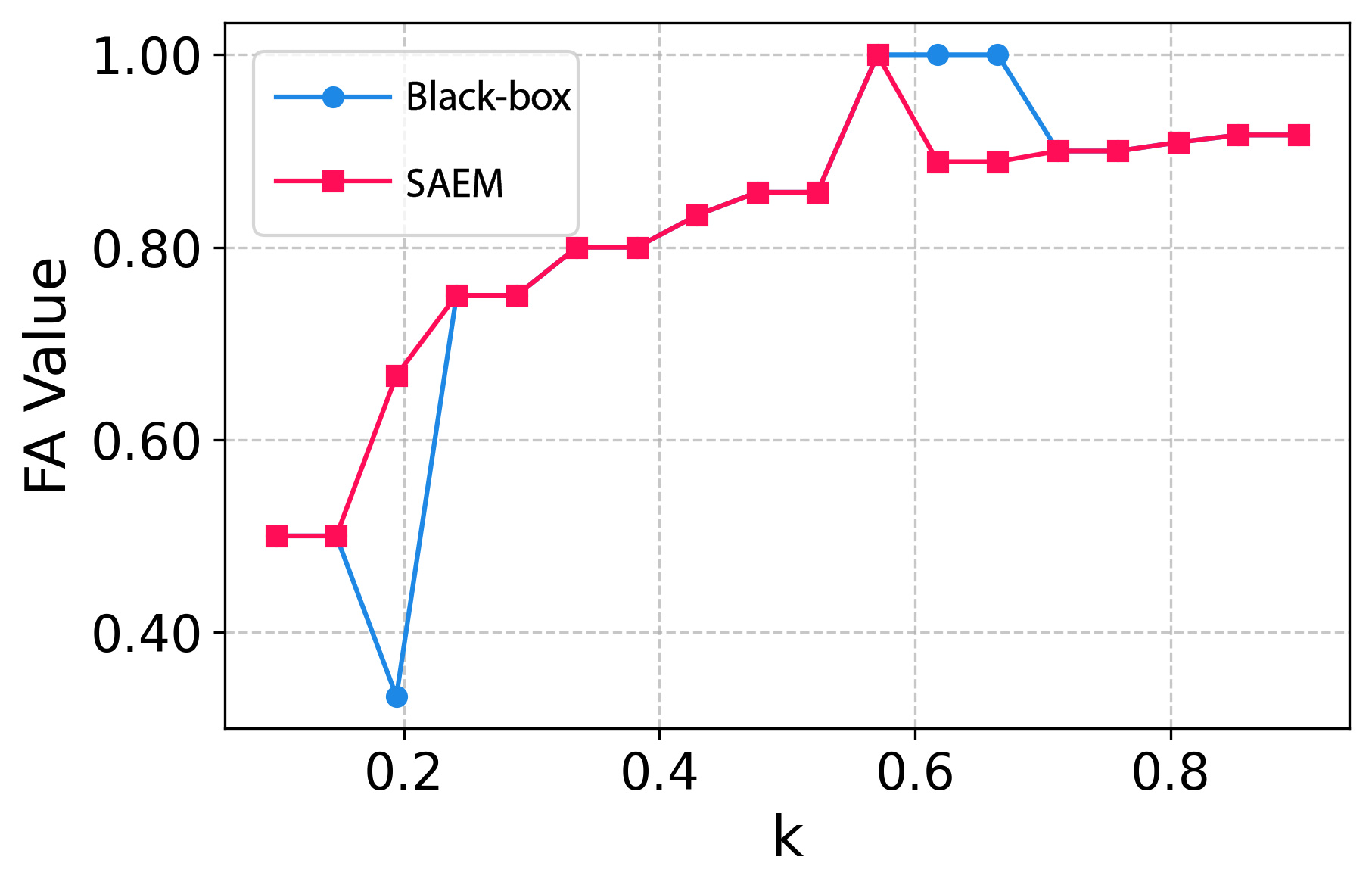}
        % \caption{FA improvement on Ground Truth}
    \end{subfigure}%
    \hfill
    \begin{subfigure}[t]{0.23\textwidth}
        \centering
        \includegraphics[width=\textwidth]{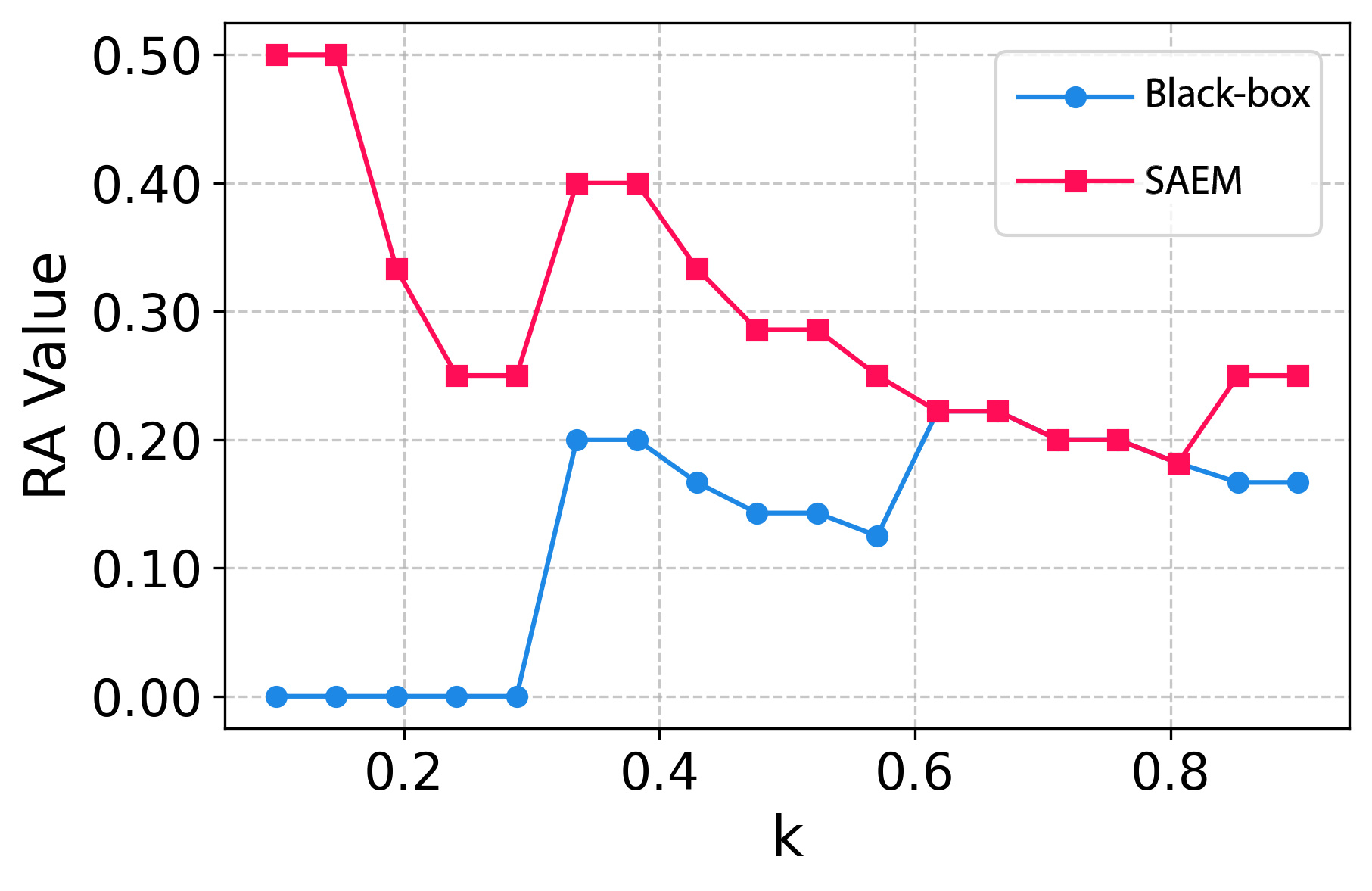}
        % \caption{RA improvement on Ground Truth}
    \end{subfigure}
    \vfill
    \begin{subfigure}[t]{0.23\textwidth}
        \centering
        \includegraphics[width=\textwidth]{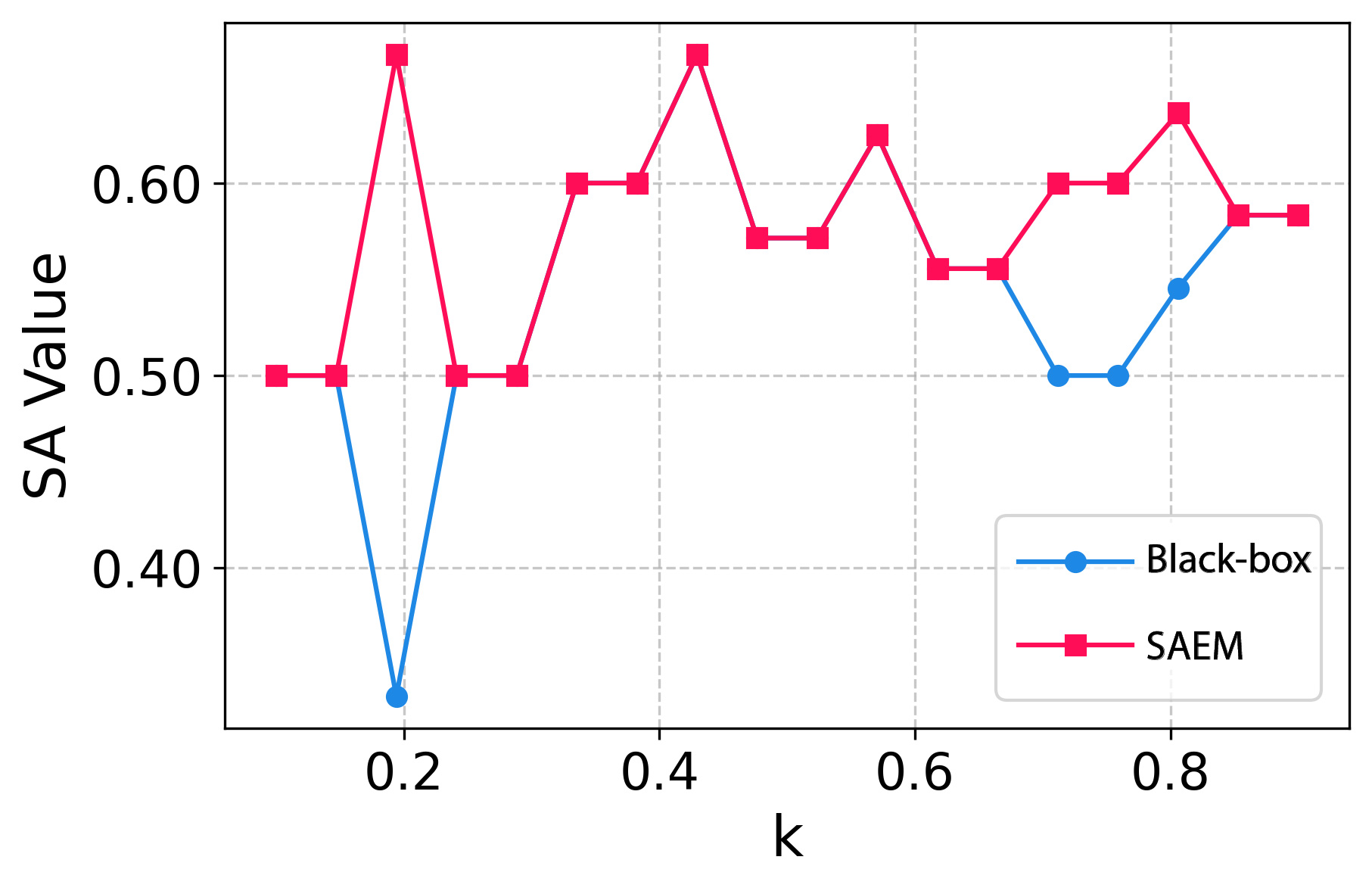}
        % \caption{SA improvement on Ground Truth}
    \end{subfigure}
    \hfill
    \begin{subfigure}[t]{0.23\textwidth}
        \centering
        \includegraphics[width=\textwidth]{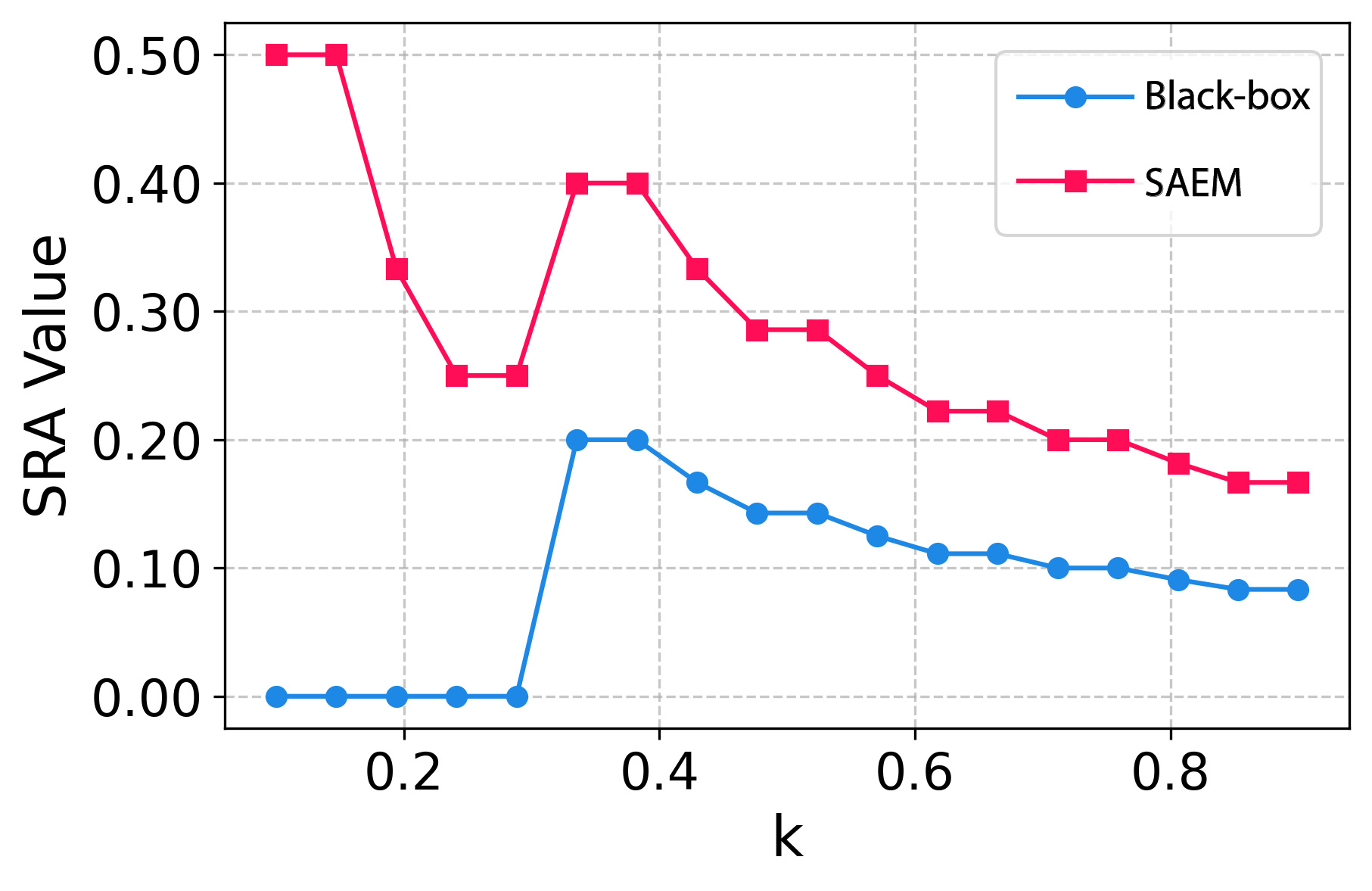}
        % \caption{SRA improvement on Ground Truth}
    \end{subfigure}
\caption{\label{fig:adult_income_comparison_k} Comparison of metrics (FA, RA, SA, SRA) between delivered ANN and the SAEMs across varying $k$s on the Adult Income dataset.} 
\end{figure} 

\paragraph{Dual Improvements for White-box Settings ($\mathcal{A}_{\text{SMA}}{=}1$).}
We study an interpretable delivered model (LR) under the ideal condition where stakeholder and model rankings coincide, $\mathbf{r}^{k} \equiv \mathbf{r}^{\text{LR}}_{\text{true}}$; hence $\mathcal{A}_{\text{SMA}}{=}1$. 
From Lemma 2.3, this implies $\mathcal{A}_{\text{faith}}{=}\mathcal{A}_{\text{plaus}}{=}1$ in the limit, i.e., faithfulness and plausibility are simultaneously maximized.

A practical instance of this setting occurs when the underlying model is interpretable but stakeholders have only limited system access and receive a \emph{post-hoc} rendering of LR’s explanation, $\mathbf{r}^{\text{LR}}_{\text{post}}$ (e.g., via an API or compliance report). 
Table~\ref{tab:results-white-box-2-1} confirms this behavior empirically: across all datasets the delivered LR explanations achieve near-perfect agreement, and SAEM either matches or slightly improves the scores while preserving predictive accuracy.

\noindent \textbf{Findings.} when $\mathcal{A}_{\text{SMA}}{=}1$, faithfulness and plausibility align; EXAGREE recovers an SAEM that achieves dual improvements under access-limited, post-hoc delivery.

\paragraph{Balancing Trade-off for White-box Settings ($\mathcal{A}_{\text{SMA}}{<}1$).}

We simulate heterogeneous stakeholders by sampling preferences at random from the pre-trained LR explainer, i.e., $\mathbf{r}^{k} \coloneqq \mathbf{r}^{\text{LR}}_{\text{Random}}$.  
For each dataset we sample from the Rashomon set and, for the fixed stakeholder $k$, plot their $(\mathcal{A}_{\text{faith}},\,\mathcal{A}_{\text{plaus}})$ pairs.  
As shown in Fig.~\ref{fig:pareto-frontier}, $\mathcal{A}_{\text{SMA}}$ varies substantially across such stakeholders in the range \mbox{$[-0.30,\,0.79]$}, indicating the variance of needs and some needs cannot be met by a single fixed model.

Each subplot displays sampled near–optimal models (grey dots), the SAEM selected by EXAGREE (red marker), and a smoothed empirical Pareto frontier (blue dashed).    
We can observe a clear faithfulness–plausibility frontier but do not attempt to enumerate it explicitly. Instead, EXAGREE selects a SAEM that moves toward the stakeholder’s preference along this frontier while respecting the accuracy tolerance $\varepsilon$, yielding a principled, user-controllable operating point without changing the explainer.  
Fig.~\ref{fig:adult_income_comparison_k} details how FA/RA/SA/SRA vary with top-$k$; fixed $k{=}25\%$ summaries appear in Appendix Sec.~7 (Tables~5–10).

\noindent \textbf{Findings.}  When stakeholder and model rankings diverge ($\mathcal{A}_{\text{SMA}}{<}1$), no single model maximizes both objectives.  
EXAGREE provides principled control of the trade-off by selecting a stakeholder-aligned model from the Rashomon set.

% \clearpage

% \paragraph{Balancing the Trade-off Between Faithfulness and Plausibility.} In the previous setting, where faithfulness is $1.00$ and $\mathcal{A}_{\text{SMA}} = \mathcal{A}_{\text{plaus}} < 1$, plausibility cannot be improved unless an alternative model is selected for the given stakeholder e.g., $\bfs{r}^{k}_{\text{FIS}}$. To address this, we identified an alternative model from the Rashomon set, SAEM, which enhances plausibility at the expense of faithfulness, as shown in the last column of Fig. \ref{fig:pareto-frontier}. We presented the increment of plausibility in terms of other metrics (FA, RA, SA, SRA) for varying top-$k$ values, as shown in Fig. \ref{fig:adult_income_comparison_k}. A fixed $k=25\%$ baseline is reported in Appendix Sec. 7 (Tables 5-10).

% In a particularly ideal case of $\mathcal{A}_{\text{SMA}} = 1$, the machine-ground explanation aligns perfectly with the stakeholder-grounded explanation. However, the delivered explanation can be $\bfs{r}^{\text{LR}}_{\text{post}}$. As we have discussed in Sec. \ref{sec:preliminaries}, the faithfulness and plausibility can be improved at the same time in this case, leading to the optimal state $\mathcal{A}_{\text{SMA}} = \mathcal{A}_{\text{faith}}= \mathcal{A}_{\text{plaus}}=1$, as shown in Table \ref{tab:results-white-box-2-1}.

\paragraph{LLM-assisted Interface.}
We provide a lightweight, LLM-assisted interface (Gemini API~\citep{GoogleGemini2023}) that lets stakeholders express preferences and feedback in natural language. Prompts are translated into a target ranking $\mathbf{r}^k$ (or soft attribution weights) that EXAGREE uses as the stakeholder reference. This enables domain experts to express their requirements in natural language and inject knowledge without ML expertise. An illustrative use case is showcased in Appendix Sec. 2, Fig. 1.

\paragraph{Fairness Evaluation in Subgroups.} 
Following work of \citet{dai2022fairness}, we evaluate whether explanation quality is consistent across sensitive subgroups, defined as fairness. Although fairness is not an explicit optimization target, we observe that the selected SAEMs surprisingly reduce subgroup gaps (difference in metric between majority and minority groups) across benchmarks in Fig.~\ref{fig:fairness_comparison} for the adult income dataset. Detailed metrics and more dataset examples are reported in Appendix Sec.~8. 
\begin{figure}[t!]
\centering
    \begin{subfigure}[t]{0.48\textwidth}
        \centering
        \includegraphics[width=\textwidth]{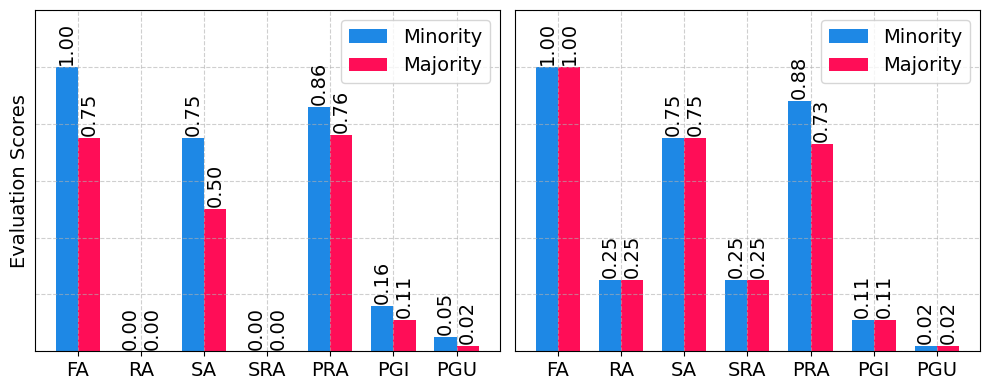}
    \end{subfigure}%
\caption{\label{fig:fairness_comparison} Subgroup fairness on Adult Income Dataset at $k=0.25$: comparison of LR (\textit{left}) and SAEM (\textit{right}). Larger red–blue gaps
indicate greater disparity;}
\end{figure} 

\section{Conclusion}
In this work, we introduced \textit{EXAGREE}, a framework that re-frames \textit{explanation disagreement} as an opportunity to align machine-learning explanations with diverse stakeholder needs. By searching within a Rashomon set, EXAGREE identifies SAEMs that flexibly prioritize faithfulness or plausibility depending on stakeholder objectives. Empirical evaluations on both synthetic and real-world datasets from the \textit{OpenXAI} benchmark demonstrate that EXAGREE consistently balances the faithfulness–plausibility trade-off across practical scenarios and, improves explanation fairness across demographic subgroups.

We also recognize several limitations that point toward impactful future work.
(i) Alternative strategies for Rashomon set sampling and more advanced differentiable sorting or ranking algorithms may further enhance optimization quality.
(ii) Applying EXAGREE in real scientific settings with human-subject validation would strengthen its practical relevance.
(iii) A more capable LLM-based interface, or even multi-agent systems, could improve usability and accessibility for non-expert stakeholders.
These limitations do not lower the value of this pilot study; rather, they highlight natural directions for extending EXAGREE into a broader and more powerful stakeholder-centered XAI framework.

\bibliography{aaai2026}

\appendix

\section*{Impact Statement}
EXAGREE contributes to the field of XAI by offering a new paradigm to the disagreement problem, empowering both AI researchers, professionals, and end-users with tools to navigate conflicting explanations systematically. By aligning explanations with stakeholder expectations, this work has practical implications for high-stakes domains such as healthcare, finance, and policy-making, where interpretability is critical for decision-making. Furthermore, the identified model potentially improves explanation fairness, enhancing trust in AI-driven systems.

% Our work highlights that no single model or explanation method can universally satisfy all stakeholders' needs. However, EXAGREE demonstrates that it's possible to balance faithfulness and plausibility or increase both in certain scenarios.

% The impact of this work could be substantial. By enhancing explanation agreement across different stakeholder groups, EXAGREE has the potential to increase trust in AI systems. 
% This is particularly crucial in high-stakes domains such as healthcare, finance, and criminal justice, where the consequences of decisions are far-reaching and the need for trustworthy explanations is paramount.
% Furthermore, our findings highlight the need for deeper exploration of the interplay between Human-Computer Interaction (HCI) and Explainable AI (XAI). This intersection represents a rich area for future research, potentially leading to more user-centered and effective explainable AI systems.

\nocite{langley00}

\section{Related Work}
\label{sec:related_work}
Our work contributes to the expansive field of explainable artificial intelligence, with a specific focus on explainable ML \cite{krishna2022disagreement}. This section discusses closely related works and their connections to our framework, clarifying key terminologies and highlighting the ongoing challenges in the field. Given the interdisciplinary nature of our research, which spans areas including sorting and ranking, non-differential optimization, Rashomon sets, and human-centered interaction, we acknowledge that an exhaustive review of all related works is beyond the scope of this section. Instead, we concentrate on literature directly pertinent to our main objective: addressing the explanation disagreement problem. Other related areas are discussed briefly, as they serve as tools or methodologies to achieve this primary purpose.

\subsection{Terminology and Core Concepts}
In the literature, the terms explanation/interpretation are often used interchangeably \citep{doshi2017towards, lipton2018mythos}. Similarly, concepts such as explanation disagreement, inconsistency, and diversity all refer to scenarios where explanations differ, whether between models, methods, or human understanding \citep{krishna2022disagreement, roscher2020explainable}. The problem of explanation disagreement remains a significant open
challenge, hindering the impact of ML models \citep{krishna2022disagreement, adebayo2018sanity, rudin2019stop, ghassemi2021false, roscher2020explainable, ribeiro2016should}. This issue exists in various forms: when a single model generates different explanations, when similar-performing models produce distinct explanations, or when model explanations diverge from human expectations.

One of the most influential works from \cite{rudin2019stop} is using interpretable models instead of black box models to avoid the problem. The idea is naturally true.
However, constructing interpretable models practically poses challenges for many groups \citep{adadi2018peeking}. For instance, it is unrealistic for end users to construct an interpretable model before they ask for reasons behind their predicted results or find a model that meets all stakeholders' needs within a given context. More importantly, even though interpretable models, such as decision trees, generate rationale behind the prediction, such explanations are not always what stakeholders expected, formulated as ground truth disagreement.
Both post-hoc for black-box models and ante-hoc for interpretable models - achieving consistent and reliable explanation agreement remains a challenge \citep{varshney2017safety, jimenez2020drug, huang2023explainable, zhong2022explainable, barnard2023importance, barnard2022explainable, reichstein2019deep, roscher2020explainable}. This leads to a common sense that ensuring human oversight of both predictions and their explanations is crucial for maintaining confidence in ML-assisted decision-making processes.

\subsection{Optimization and Ranking in Explanations}

End-to-end optimization in ML often involves sorting and ranking operations, which present unique challenges due to their non-differentiable nature. Sorting is a piecewise linear function with numerous non-differentiable points, while ranking is a piecewise constant function with null or undefined derivatives. These properties make it difficult to incorporate sorting and ranking directly into gradient-based optimization frameworks.

Sorting networks, a concept dating back to the 19th century \citep{knuth1997art}, offer a potential solution. These highly parallel, data-oblivious sorting algorithms use conditional pairwise swap operators to map inputs to ordered outputs. Recent advancements have led to the development of differentiable sorting networks, also known as soft rank methods. These techniques approximate the discrete sorting operation with a continuous, differentiable function. One popular approach is the use of the neural sort operator \citep{grover2019stochastic}, which employs a differentiable relaxation of the sorting operation. Another method involves using the optimal transport formulation to create a differentiable proxy for sorting \citep{cuturi2019differentiable}. In the EXAGREE framework, we adopt the recent DiffSortNet algorithm proposed by  \cite{petersen2022monotonic}. This approach offers several advantages, including the simplicity of using a logistic sigmoid function and the guarantee of monotonicity in the sorting operation. DiffSortNet provides a differentiable approximation of sorting that maintains the essential properties of traditional sorting while enabling gradient-based optimization. Sorting and ranking attributions in a Rashomon set also poses difficulties, we discussed the constraints in Sec. \ref{sec:stage1}

\subsection{Rashomon sets Related Works}
Recent research has increasingly advocated for exploring sets of equally good models, rather than focusing on a single model \cite{rudin2019stop, rudin2024amazing, hsu2022rashomon, li2023exploring}. This approach, known as the Rashomon set concept, provides a more comprehensive understanding of model behavior and feature importance.
\cite{fisher2019all} first introduced the concept of Model Class Reliance (MCR). Building on this work, \cite{dong2020exploring} explored the cloud of variable importance (VIC) for the set of all good models, providing concrete examples in linear and logistic regression. Further expanding on these ideas, \cite{hsu2022rashomon} investigated instance-level explanations within a set of models, while \cite{li2023exploring} introduced the concept of Feature Interaction Score (FIS) in the Rashomon set. While these existing works have significantly advanced our understanding of model behavior through the lens of Rashomon sets, the practical benefits and applications of this approach have remained largely unexplored.
Our work represents a significant step forward in this domain, as it presents the first practical application of the Rashomon set concept to address the challenge of explanation disagreement.

\section{Use Case of EXAGREE with LLM}\label{sec:user-demo}
\begin{figure*}[]
\centering
    \begin{subfigure}[t]{.75\textwidth}
        \centering
        \includegraphics[width=\textwidth]{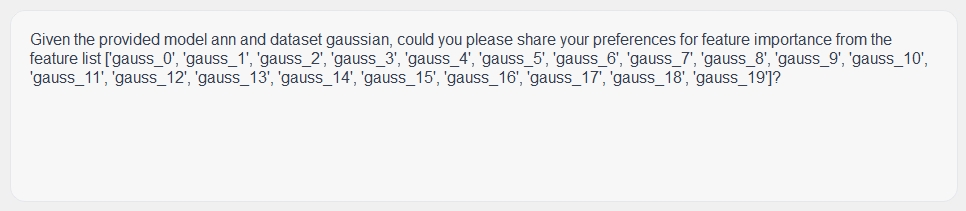}
        \caption{The welcome message from our demo GUI}
    \end{subfigure}%
    \hfill
    \begin{subfigure}[t]{.75\textwidth}
        \centering
        \includegraphics[width=\textwidth]{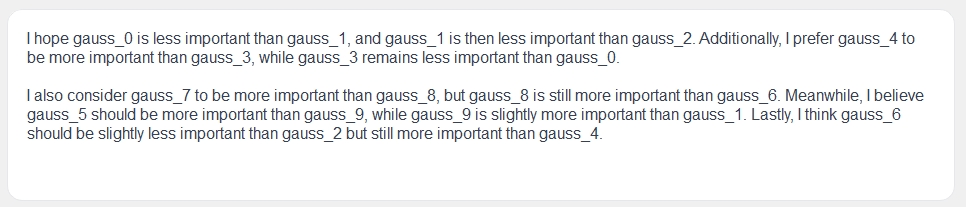}
        \caption{The preference from user as stakeholder-grounded explanations}
    \end{subfigure}%
    \hfill
    \begin{subfigure}[t]{.75\textwidth}
        \centering
        \includegraphics[width=\textwidth]{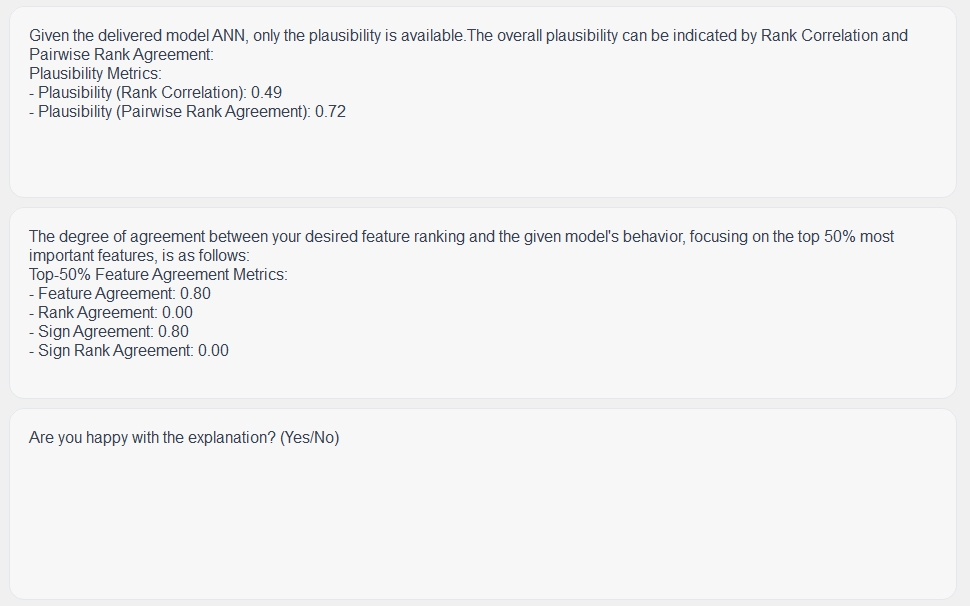}
        \caption{Initial response from EXAGREE, showing the initial plausibility is 0.49}
    \end{subfigure}
    \hfill
    \begin{subfigure}[t]{.75\textwidth}
        \centering
        \includegraphics[width=\textwidth]{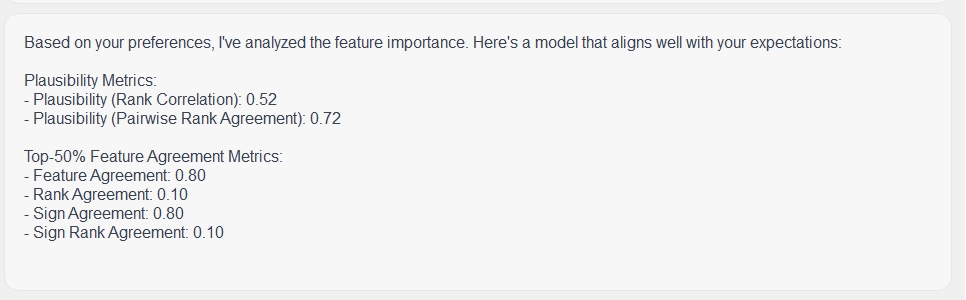}
        \caption{Second response from identified SAEM with higher plausibility 0.52 (6\% improvment)}
    \end{subfigure}
\caption{\label{fig:exagree_demo} Demonstration of EXAGREE's user-friendly interface for improving plausibility by identifying SAEM according to stakeholder preferences. (a) The system prompts the user to provide feature importance preferences for a given ann (black-box model) and synthetic dataset. (b) The user expresses preferences using natural language. (c) EXAGREE evaluates the plausibility between the user's desired feature ranking and the current model's behavior. (d) The user is not happy with the outcome. After optimization, EXAGREE presents an improved model that better aligns with the user's expectations, showing 6\% improvement in plausibility.}
\end{figure*}

The EXAGREE framework incorporates user-friendly functionality that bridges complex technical implementations and stakeholder needs. It accommodates various input forms, potentially enhancing its practical utility. Generally, stakeholders can directly provide expected feature attribution rankings, ensuring usability for diverse groups. Additionally, experts can specify detailed attributions or directional preferences for features. For instance, a stakeholder might express, ``I believe this feature should have a negative impact on the outcome'', or provide specific attribution values. 
Leveraging advancements in LLMs, EXAGREE utilizes natural language processing capabilities through the Gemini API \citep{GoogleGemini2023}. This allows stakeholders to express preferences and feedback without extensive ML expertise, making explainable AI more accessible in real-world applications. 
Here we provided a demo to illustrate the usage of these functionalities on a stakeholder with arbitrary needs working with the provided black-box model ANN and the Synthetic dataset, as shown in Fig. \ref{fig:exagree_demo}.

\section{Additional Proof}

\begin{lemma}[Faithfulness and Plausibility Trade‐Off]
If $\mathcal{A}_{\text{SMA}}\!<\!1$, no delivered explanation can simultaneously maximize $\mathcal{A}_{\text{faith}}$ and $\mathcal{A}_{\text{plaus}}$.  
\end{lemma}
\begin{proof}
    Assume strict (tie-free) rankings, then
    $\mathcal{A}_{\text{SMA}} < 1
    \;\;\Longleftrightarrow\;\;
    \mathbf r^{k} \;\neq\; \mathbf r^{M^{\!*}}_{\text{true}}.$
    Given \(\bfs{r}^k \neq \bfs{r}^{M^*}_{\text{true}}\), it is not possible for a single explanation \(\bfs{r}^{M}_{\varphi}\) to fully align with both \(\bfs{r}^k\) and \(\bfs{r}^{M^*}_{\text{true}}\). 
    This results in a discrepancy between \(\mathcal{A}_{\text{faith}}\) and \(\mathcal{A}_{\text{plaus}}\). Thus, the trade-off between faithfulness and plausibility follows directly.
\end{proof}

\begin{lemma}\label{lemma:trade-off}
    Given a delivered explanation $\bfs{r}^{M}_{\varphi}$ and a misalignment between stakeholder needs and the machine-grounded explanation (\(\bfs{r}^k \neq \bfs{r}^{M^*}_{\text{true}}\)), there is a trade-off between faithfulness and plausibility.
\end{lemma}
\begin{proof}
    Given \(\bfs{r}^k \neq \bfs{r}^{M^*}_{\text{true}}\), it is not possible for a single explanation \(\bfs{r}^{M}_{\varphi}\) to fully align with both \(\bfs{r}^k\) and \(\bfs{r}^{M^*}_{\text{true}}\). 
    This results in a discrepancy between \(\mathcal{A}_{\text{faith}}\) and \(\mathcal{A}_{\text{plaus}}\), making $
    \mathcal{A}_{\text{faith}} - \mathcal{A}_{\text{plaus}} \neq 0.
    $ Thus, the trade-off between faithfulness and plausibility follows directly.
\end{proof}

\begin{proposition}\label{prop:A_sma}
    A higher $\mathcal{A}_{\text{SMA}}$ is desirable and necessary for faithful and plausible explanations.
\end{proposition}
\begin{proof}\label{proof:agreement-relations}
Faithfulness, plausibility, and SMA can be expressed in the following format based on Spearman’s rank correlation:

\small{\begin{align*}
\mathcal{A}_{\text{faith}} &= \mathcal{O}(\mathbf{r}^{M^*, \varphi}, \mathbf{r}^{M^*}_{\text{true}}) 
= 1 - \frac{6}{n(n^2 - 1)} \sum_{i=1}^n \left( r^{M^*}_{\text{true},i} - r^{M^*}_{\varphi,i} \right)^2, \\
\mathcal{A}_{\text{plaus}} &= \mathcal{O}(\mathbf{r}^{M^*, \varphi}, \mathbf{r}^k) 
= 1 - \frac{6}{n(n^2 - 1)} \sum_{i=1}^n \left( r^k_i - r^{M^*}_{\varphi,i} \right)^2, \\
\mathcal{A}_{\text{SMA}} &= \mathcal{O}(\mathbf{r}^k, \mathbf{r}^{M^*}_{\text{true}}) 
= 1 - \frac{6}{n(n^2 - 1)} \sum_{i=1}^n \left( r^k_i - r^{M^*}_{\text{true},i} \right)^2.
\end{align*}}

Next, we apply the binomial expansion to the SMA term:
\begin{align*}
\left( r^k_i - r^{M^*}_{\text{true},i} \right)^2 
&= \left( r^k_i - r^{M^*}_{\varphi,i} + r^{M^*}_{\varphi,i} - r^{M^*}_{\text{true},i} \right)^2 \\
&= \left( r^k_i - r^{M^*}_{\varphi,i} \right)^2 + \left( r^{M^*}_{\text{true},i} - r^{M^*}_{\varphi,i} \right)^2 \\
&\quad - 2 \left( r^k_i - r^{M^*}_{\varphi,i} \right)\left( r^{M^*}_{\text{true},i} - r^{M^*}_{\varphi,i} \right).
\end{align*}

Summing over all \(i\), we obtain:
\small{\begin{align*}
\sum_{i=1}^n \left( r^k_i - r^{M^*}_{\text{true},i} \right)^2 
&= \sum_{i=1}^n \left( r^k_i - r^{M^*}_{\varphi,i} \right)^2 
+ \sum_{i=1}^n \left( r^{M^*}_{\text{true},i} - r^{M^*}_{\varphi,i} \right)^2 \\
&\quad - 2 \sum_{i=1}^n \left( r^k_i - r^{M^*}_{\varphi,i} \right)\left( r^{M^*}_{\text{true},i} - r^{M^*}_{\varphi,i} \right).
\end{align*}}

Substituting into the SMA definition, we derive the core decomposition:
\begin{equation}
\mathcal{A}_{\text{SMA}} = \mathcal{A}_{\text{plaus}} + \mathcal{A}_{\text{faith}} - \mathcal{A}_{\text{interaction}},
\label{eq:relation-all}
\end{equation}

where the interaction term is defined as:
\[
\mathcal{A}_{\text{interaction}} = \frac{6}{n(n^2 - 1)} \sum_{i=1}^n \left( r^k_i - r^{M^*}_{\varphi,i} \right)\left( r^{M^*}_{\text{true},i} - r^{M^*}_{\varphi,i} \right).
\]

\(\mathcal{A}_{\text{SMA}}\) indicates the divergence between the dual objectives of plausibility and faithfulness. When \(\mathcal{A}_{\text{SMA}} = 1\), faithfulness and plausibility are perfectly aligned, and the delivered explanation (\(\bfs{r}^{M}_{\varphi}\)) satisfies both objectives simultaneously. 
We summarize the relationship between SMA, faithfulness, and plausibility as follows:
\begin{itemize}
    \item \textit{Case of SMA = \(1\):} Faithfulness and plausibility are equal. The interaction term is zero, indicating no trade-off between faithfulness and plausibility. A faithful post-hoc explanation is also plausible, achieving high scores in both metrics.
    
    \item \textit{Case of SMA = \(0\):} Faithfulness and plausibility diverge. A faithful explanation may not be plausible (\(\mathcal{O}_{\text{faithfulness}} \gg \mathcal{O}_{\text{plausibility}}\)), and a plausible explanation may not be faithful (\(\mathcal{O}_{\text{plausibility}} \gg \mathcal{O}_{\text{faithfulness}}\)). The interaction term reflects a significant trade-off between the two metrics.

    \item \textit{Case of SMA = \(-1\):} Faithfulness and plausibility are mutually exclusive (\(\mathcal{O}_{\text{faithfulness}} + \mathcal{O}_{\text{plausibility}} \approx 0\)). A faithful explanation minimizes plausibility, and a plausible explanation minimizes faithfulness. The interaction term is maximized, representing an extreme trade-off.
\end{itemize}
\end{proof}

This identity shows that SMA captures the joint alignment of faithfulness and plausibility, penalized by the interaction term. If $\mathcal{A}_{\text{SMA}} = 1$, then $\mathcal{A}_{\text{faith}} = \mathcal{A}_{\text{plaus}} = 1$ and $\mathcal{A}_{\text{interaction}} = 1$, indicating perfect agreement. In general, SMA quantifies how much these objectives diverge—and motivates choosing a model for which both can be jointly maximized.

\begin{lemma}\label{lemma:stakeholder}
    Given the diverse and conflicting expectations of stakeholders in practical settings, there always exists at least one stakeholder who is not fully satisfied with the delivered explanation.
\end{lemma}
\begin{proof}
    Consider an interpretable model $M^{*}_{\mathcal{I}}$ delivering an explanation $\bfs{r}^{M^{*}}_{\text{true}}$ to multiple stakeholders $\mathcal{S}$, we suppose two stakeholders have different needs $\exists k, j \in \mathcal{S}, \bfs{r}^{k} \neq \bfs{r}^{j}$. That will lead to different \( \mathcal{A}_{\text{SMA}}\) values, where at least one \( \mathcal{A}_{\text{SMA}} \neq 1 \). An inherent trade-off between faithfulness and plausibility needs to be considered. A single delivered explanation cannot optimize both for all stakeholders simultaneously. Consequently, at least one stakeholder will require an alternative model that either increases faithfulness or improves plausibility, depending on their individual preference. 

    Consider a black box model $M^{*}$, if one stakeholder is perfectly satisfied with the delivered explanation (\( \mathcal{A}_{\text{plaus}} = 1 \)), another stakeholder with different needs will necessarily find the explanation misaligned with their expectations. Formally, there exists at least one stakeholder \( k \) for whom \( \mathcal{O}(\bfs{r}^k, \bfs{r}^{M^{*} }_{\varphi}) \neq 1 \), indicating dissatisfaction. This dissatisfaction necessitates an alternative model that offers a more plausible explanation for that stakeholder.  
\end{proof}

\begin{lemma}
\label{lemma:Spearman}
For a given target ranking \( \bfs{r}^{*} \), there may exist multiple distinct rankings that have the same Spearman's rank correlation coefficient with the target ranking.
\end{lemma}
\begin{proof}

The Spearman’s rank correlation coefficient \( \mathcal{O} \) between any ranking \( \bfs{r} \) and the target ranking \( \bfs{r}^{*} \) can be expressed as:
\[
\mathcal{O}(\bfs{r}, \bfs{r}^{*}) = 1 - \frac{6 \sum_{i=1}^n d_i^2}{n(n^2 - 1)},
\]
where \( d_i = r^{*}_i - r_i \) is the difference between the ranks of the \( i \)-th element in the target ranking \( \bfs{r}^{*} \) and the compared ranking \( \bfs{r} \), and \( n \) is the total number of elements in the ranking. This formula shows that \( \mathcal{O} \) depends solely on the sum of squared rank differences \( D = \sum_{i=1}^n d_i^2 \).

Let us define two distinct rankings: $\bfs{r} = [r_1, r_2, \ldots, r_n]$ and 
$\bfs{r}' = [r_1', r_2', \ldots, r_n']$. 
The rank difference for the \( i \)-th element in relation to the target ranking is given by: $d_i = r^{*}_i - r_i,$ and $d_i' = r^{*}_i - r_i'$. Thus, 
The sum of squared rank differences for each ranking in $\bfs{r}$ is expressed as:
\[
D_r = \sum_{i=1}^n d_i^2 = \sum_{i=1}^n (r^{*}_i - r_i)^2,
\]
and for $\bfs{r}'$:
\[
D_{r'} = \sum_{i=1}^n d_i'^2 = \sum_{i=1}^n (r^{*}_i - r_i')^2.
\]

To establish that $D_r = D_{r'}$ , we expand the expressions:
$$D_r = \sum_{i=1}^{n}(r^{*}_i - r_i)^{2} = \sum_{i=1}^n (r^{*2}_i - 2r^{*}_{i} r_i + r^2_i).$$
$$\quad D_{r'} = \sum_{i=1}^n (r^{*}_i - r_i')^2 = \sum_{i=1}^n (r^{*2}_i - 2r^{*}_i r_i' + r_i'^2)$$

Setting \( D_r = D_{r'} \) leads to:
\[
\sum_{i=1}^n (r^{*2}_i - 2r^{*}_i r_i + r_i^2) = \sum_{i=1}^n (r^{*2}_i- 2r^{*}_i r_i' + r_i'^2).
\]

Cancelling \( \sum_{i=1}^n r^{*2}_i \) from both sides gives:
\[
\sum_{i=1}^n (-2r^{*}_i r_i + r_i^2) = \sum_{i=1}^n (-2r^{*}_i r_i' + r_i'^2).
\]

Rearranging, we find that for \( D_r = D_{r'} \), the following condition must hold:
\[
\sum_{i=1}^n (r_i^2 - r_i'^2) = 2 \sum_{i=1}^n r^{*}_i (r_i' - r_i).
\]

This implies that either:
1. The differences between $r_i'$ and $r_i$ must balance out when weighted by the corresponding \( r^{*}_i \).
2. The squared values of the ranks in \( r_i \) and \( r_i' \) must differ in a way that maintains the overall relationship with \( r^{*}_i \). These conditions are not mutually exclusive and can be satisfied simultaneously. They allow for the existence of distinct rankings $\bfs{r}$ and $\bfs{r}'$ that maintain the same Spearman's rank correlation with $\bfs{r}^{*}$.

In the case $\mathcal{O} = 1$ or $\mathcal{O} = -1$, there are no distinct rankings, such as \( \bfs{r} \) and \( \bfs{r}' \) that can yield the same sum of squared rank differences \( D_r = D_{r'} \), as any deviation would introduce non-zero differences, violating the condition \( D_r = D_{r'} \).

In the case \( \mathcal{O} \in (-1, 1)\), the constraints are relaxed, allowing for the possibility of multiple distinct rankings \( r \) and \( r' \) yielding the same \( D \). The possibility of such cases increases with a greater number of elements \( n \), as the number of distinct permutations that can maintain the same squared rank differences increases. Thus, the condition for two distinct rankings \( \bfs{r} \) and \( \bfs{r}' \) to have the same sum of squared rank differences \( D \) with respect to a target ranking \( \bfs{r}^{*} \) is established as described above.

Therefore, multiple distinct rankings can achieve the same Spearman's rank correlation coefficient with the target ranking \( \bfs{r}^{*} \), which completes the proof.
\end{proof}

% \begin{lemma}\label{lemma:impossible-swap}
%     In a Rashomon set $\mathcal{M}_\epsilon$, not all pairwise attribution swaps are possible.
% \end{lemma}
% \begin{proof}
%     Assume, for the sake of contradiction, that any pairwise swap of attributions is possible within the Rashomon set. However, this contradicts the fact that the feature attribution range or model class reliance is not unlimited, as demonstrated in prior Rashomon-related theoretical and empirical studies \citep{fisher2019all, li2023exploring, hsu2022rashomon, NEURIPS2022_5afaa8b4, li2024practical}.
%     Consider a scenario where the feature attribution matrix from the Rashomon set is: $$[min(\bfs{a}_i), max(\bfs{a}_i)]^{p}_{i=1},$$ 
%     where we specify $\max(a_1) < \min(a_2)$, as illustrated in the middle panel of Fig. \ref{fig:rankings}. In this case, the swap $a_{(1)}$ between $a_{(2)}$ is impossible within this Rashomon set.
% \end{proof}

\begin{proposition}\label{prop:not-always-satisfied}
    Based on the previous lemma \ref{lemma:impossible-swap}, we can conclude that there does not always exist a model within the Rashomon set $\mathcal{M}_\epsilon$ that satisfies a stakeholder's expectation.
\end{proposition}
\begin{proof}
    As shown in the lemma, consider an expected ranking is $\mathbf{r}^* = (a_{(1)} \succ a_{(2)} \succ \cdots \succ a_{(p)})$, and the feature attribution matrix from the Rashomon set exhibits the property $\max(a_1) < \min(a_2)$, there does not exist a model $M \in \mathcal{M}_\epsilon$ that can satisfy the condition $a_{(1)} \succ a_{(2)}$, formulated as:
    \begin{equation*}
        \nexists M \in \mathcal{M}_\epsilon : a_{(1)} \succ a_{(2)}.
    \end{equation*}
    This means that the stakeholder's expected ranking $\mathbf{r}^*$ is inaccessible within this Rashomon set $\mathcal{M}_\epsilon$, as the necessary pairwise attribution swaps to achieve the desired ranking order is not feasible.
\end{proof}

\begin{remark}
    One example is: let $\bfs{r}^* = [1, 2, 3, 4, 5]$ be our target ranking. Consider the following two distinct rankings:
    $\bfs{r} = [1, 3, 2, 5, 4]$ and 
    $\bfs{r}' = [2, 1, 3, 5, 4]$,
    it is easy to calculate their correlations    $\mathcal{O}(\bfs{r}^*, \bfs{r}) = \mathcal{O}(\bfs{r}^*, \bfs{r}') = 0.8,$ and show the equivalence. 
\end{remark}

\section{Loss Functions on Attribution Direction, Sparse and Diverse Constraints}\label{appendix:loss_fns}
To ensure that our multi-head architecture produces meaningful and diverse solutions while respecting stakeholder input, we introduce several key constraints:

\textbf{Attribution Direction} ($\Loss_{\text{sign}}$): We recognize the importance of maintaining the direction of feature attributions as specified by stakeholders. To achieve this, we incorporate a sign loss:
\begin{equation*}
    \mathcal{L}_{\text{sign}} = \mathcal{L}_{\text{MSE}}(\text{sign}(\mathbf{a}^{M}_{\varphi}), \text{sign}(\mathbf{a}^*_{\text{true}}))
\end{equation*}
This ensures that the sign of the attributions in our identified models aligns with the stakeholder-specified directions, when $\mathbf{a}^*_{\text{true}}$ provided as ground truth attributions. The corresponding target ranking is derived as $\mathbf{r}^*_{\text{true}} = f_{\text{diffsort}}(|\mathbf{a}^*_{\text{true}}|)$.

\textbf{Sparsity Constraint} ($\Loss_{\text{sparsity}}$) and \textbf{Diversity Constraint} ($\Loss_{\text{diversity}}$): To encourage both variation across masks and within each mask, we implement sparsity and diversity losses.  
$\Loss_{\text{sparsity}}$ controls the distribution of values across different masks and $\Loss_{\text{diversity}}$
maximizes the variance within each mask, ensuring they focus on distinct feature subsets. We will refine their definitions mathematically.
\begin{equation*} \mathcal{L}_{\text{sparsity}} = - \frac{1}{|\mathcal{M}|} \sum_{M \in \mathcal{M}} ||\mathbf{m}^{M}||_2, \mathcal{L}_{\text{diversity}} = - \frac{1}{p} \sum_{i=1}^{p} ||\mathbf{m}_{i}||_2,\end{equation*}
where $\mathbf{m}_{i}$ is the vector of mask values for feature $i$ across all models, $\mathbf{m}^{M}$ is the vector of mask values for model $M$ across all features. $||\bfs{m}||_2$ computes the L2 norm across features or models.
The overall objective for a stakeholder group, incorporating these constraints, is formulated as:
\begin{equation}
    \min_{\Theta} (\Loss_{\text{rank}} + \mathcal{L}_{\text{sign}} + \lambda_1 \mathcal{L}_{\text{sparsity}} + \lambda_2 \mathcal{L}_{\text{diversity}}),
\end{equation}
where $\lambda_1$ and $\lambda_2$ are hyperparameters that control the weight of the sparsity and diversity losses. By propagating the error backward through the surrogate network and the sorting network, we can train the multi-head network. The algorithm is provided as pseudocode in Algorithm \ref{alg:a1}.

\section{Ablation Study}\label{sec:ablation}

This study investigates how the size of the Rashomon set, controlled by the parameter $\epsilon$, affects EXAGREE's ability to identify models that align with stakeholder expectations. We conducted experiments on the Synthetic dataset using both pre-trained LR and ANN models provided by OpenXAI, varying $\epsilon$ values (0.05, 0.1, and 0.2). Intuitively, a larger Rashomon set should provide more opportunities for finding stakeholder-aligned models due to a larger search space.

\begin{figure*}[]
\centering
    \begin{subfigure}{\textwidth}
    \centering
        \begin{subfigure}[t]{0.24\textwidth}
        \centering
        \includegraphics[width=\textwidth]{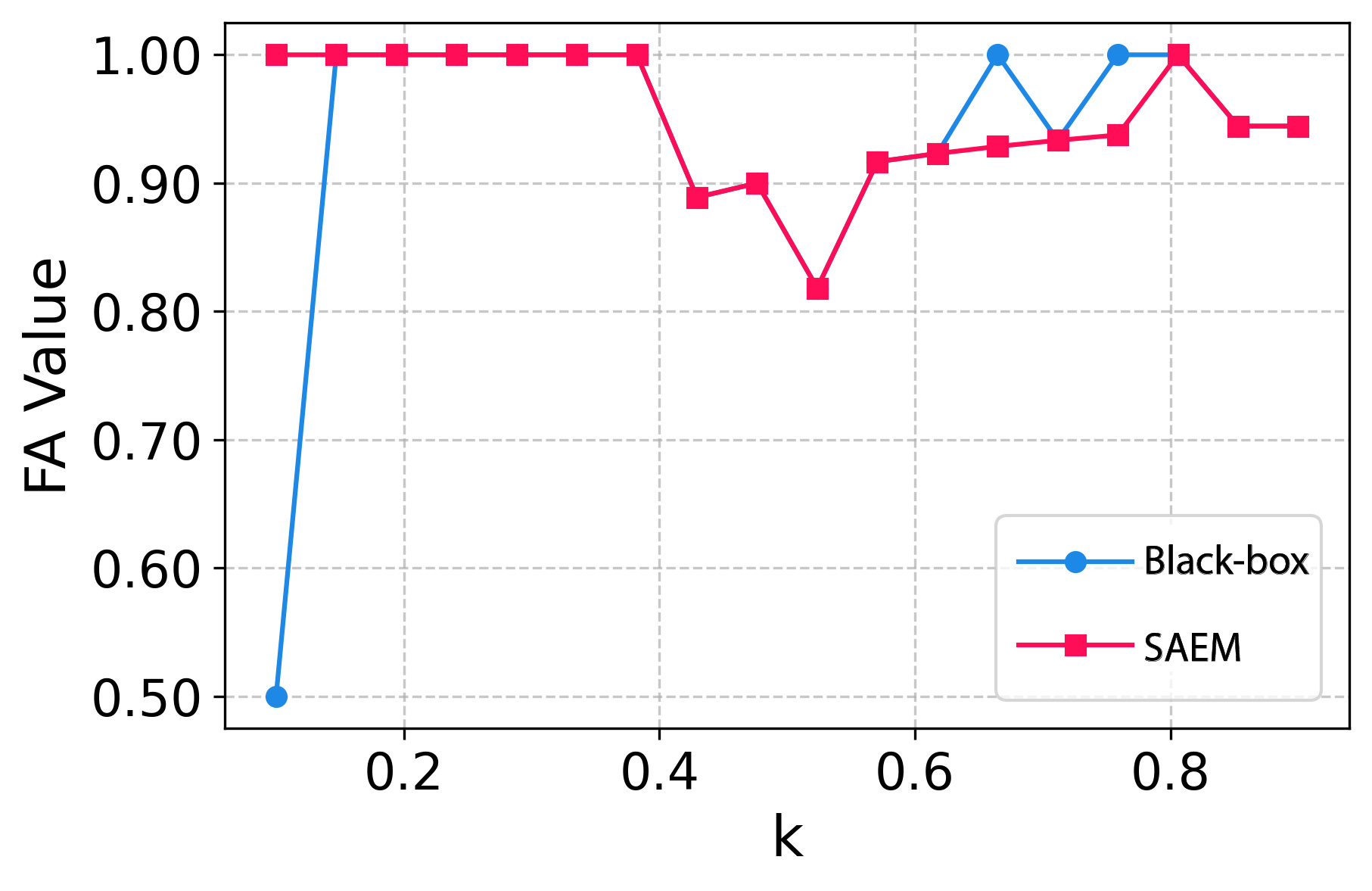}
        % \caption{FA improvement on Ground Truth}
        \end{subfigure}%
        \hfill
        \begin{subfigure}[t]{0.24\textwidth}
        \centering
        \includegraphics[width=\textwidth]{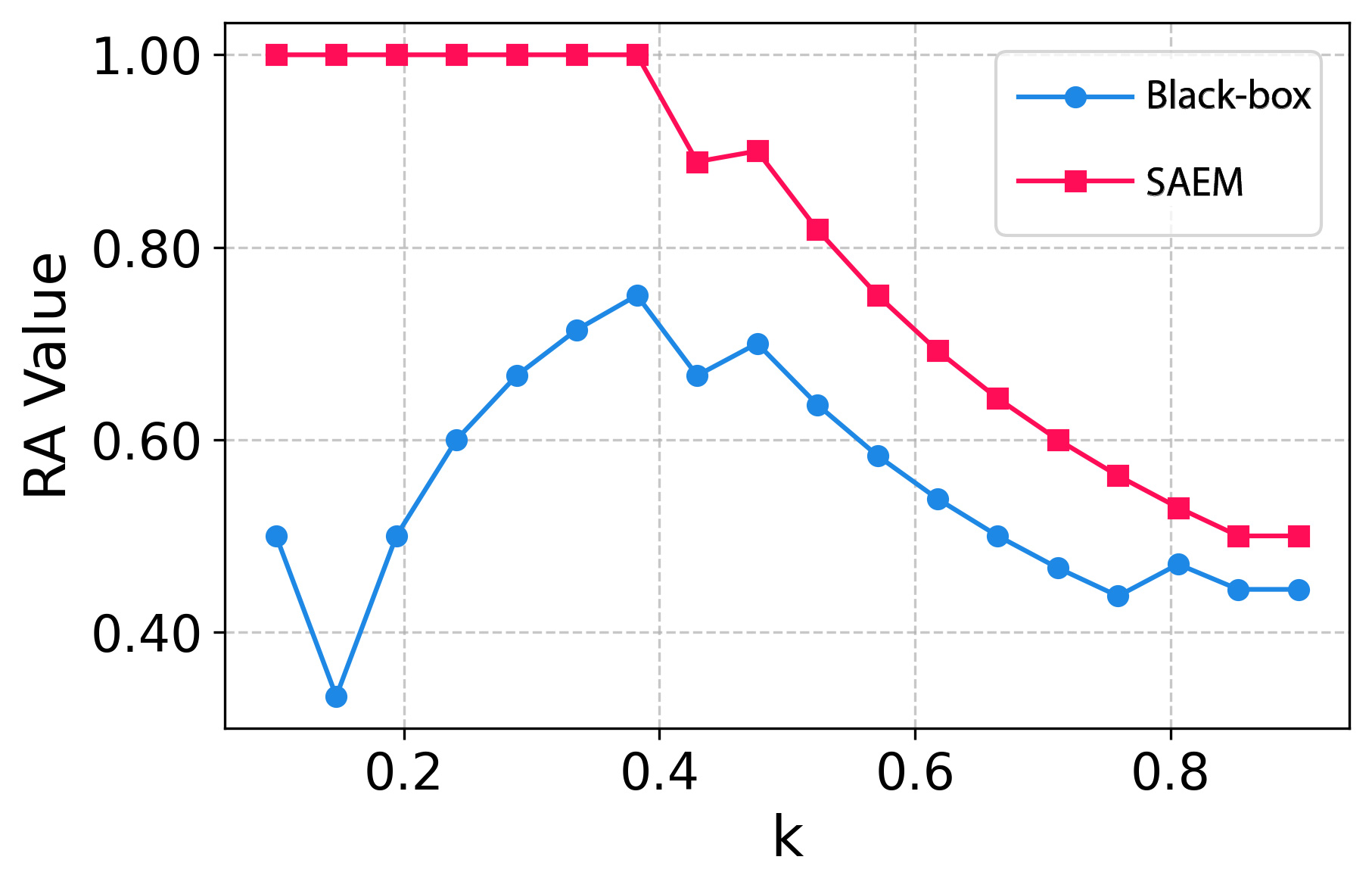}
        % \caption{RA improvement on Ground Truth}
        \end{subfigure}
        \hfill
        \begin{subfigure}[t]{0.24\textwidth}
        \centering
        \includegraphics[width=\textwidth]{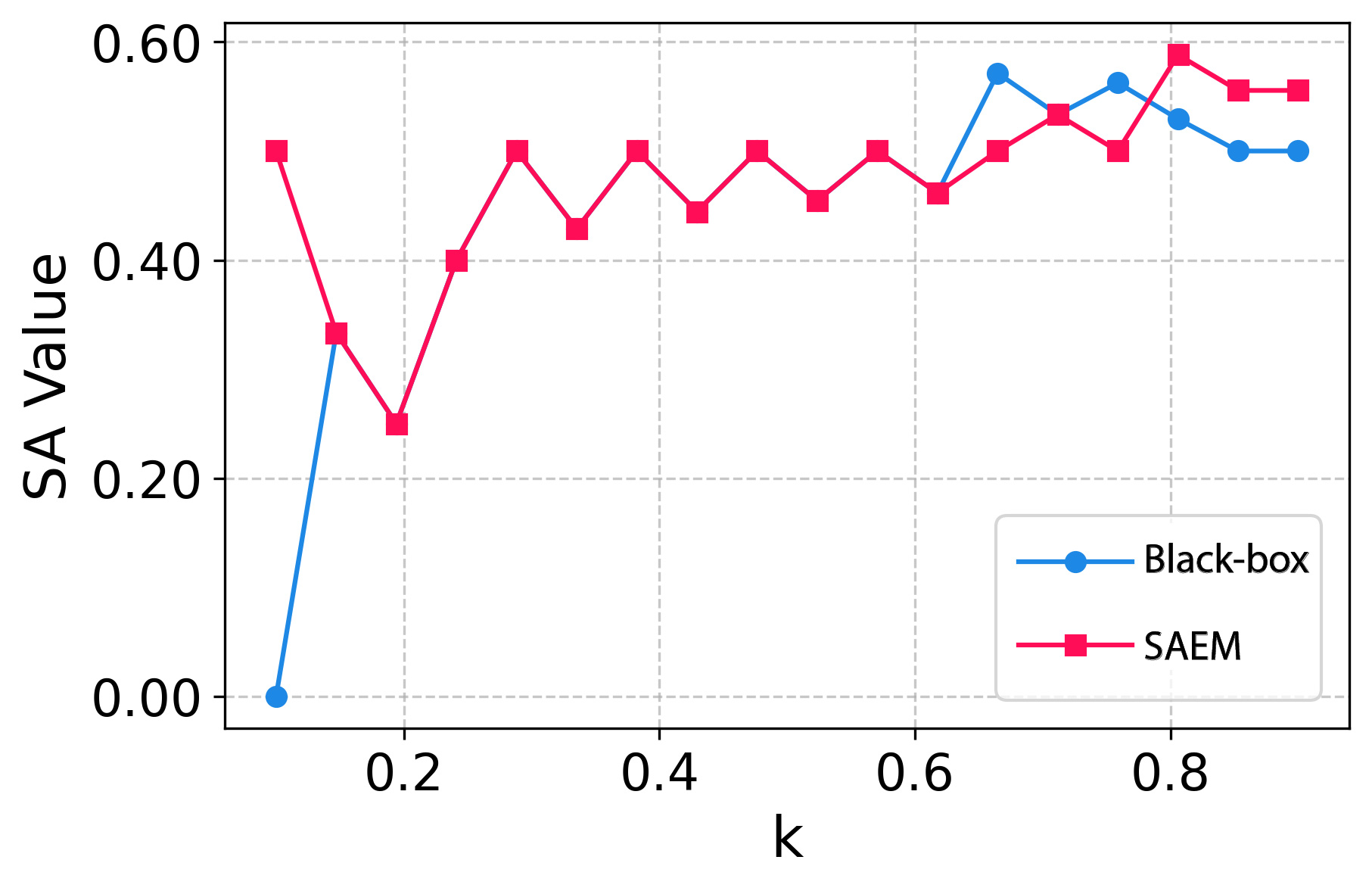}
        % \caption{SA improvement on Ground Truth}
        \end{subfigure}
        \hfill
        \begin{subfigure}[t]{0.24\textwidth}
        \centering
        \includegraphics[width=\textwidth]{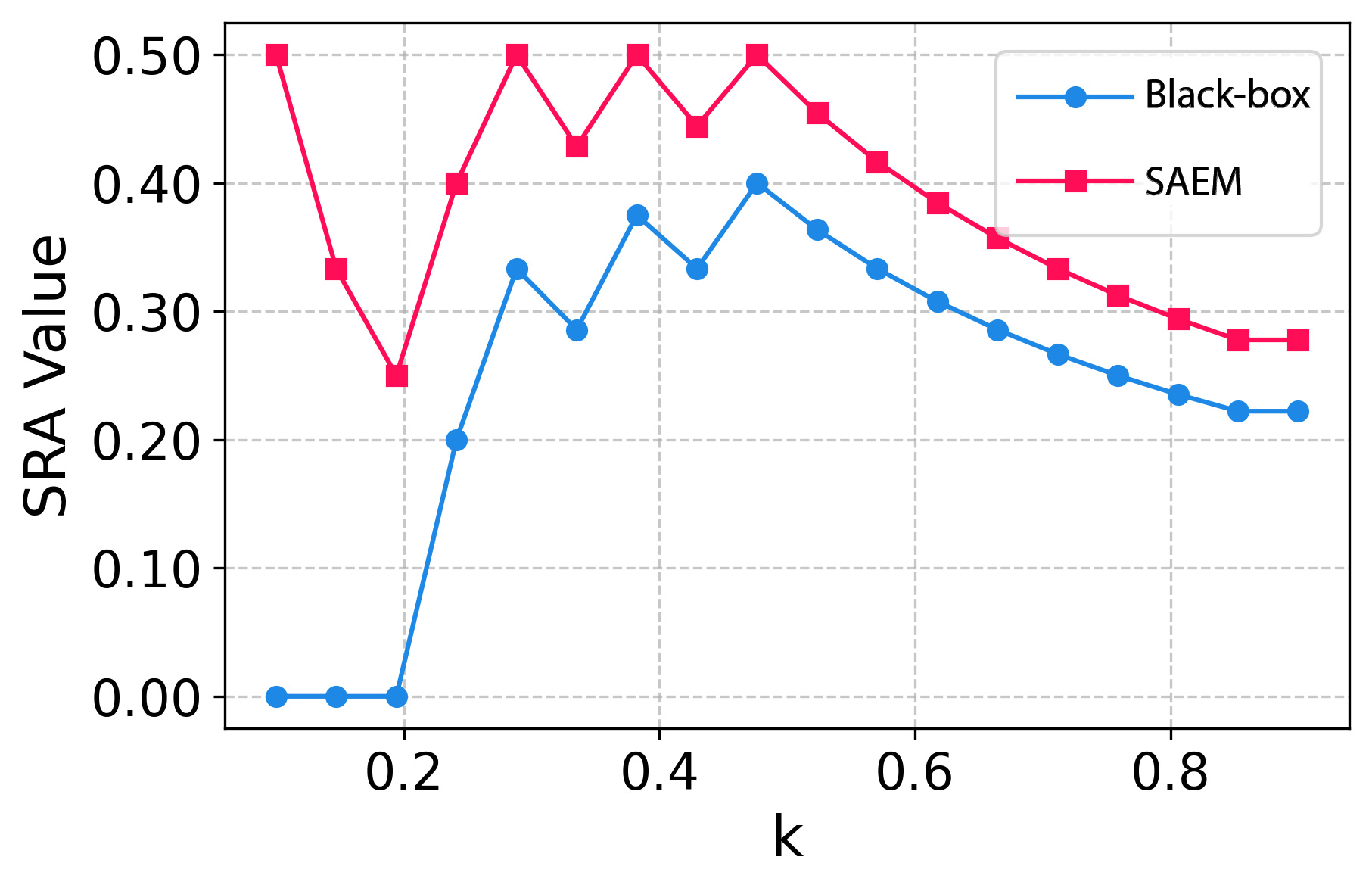}
        % \caption{SRA improvement on Ground Truth}
        \end{subfigure}
        \vfill
        \begin{subfigure}[t]{0.24\textwidth}
        \centering
        \includegraphics[width=\textwidth]{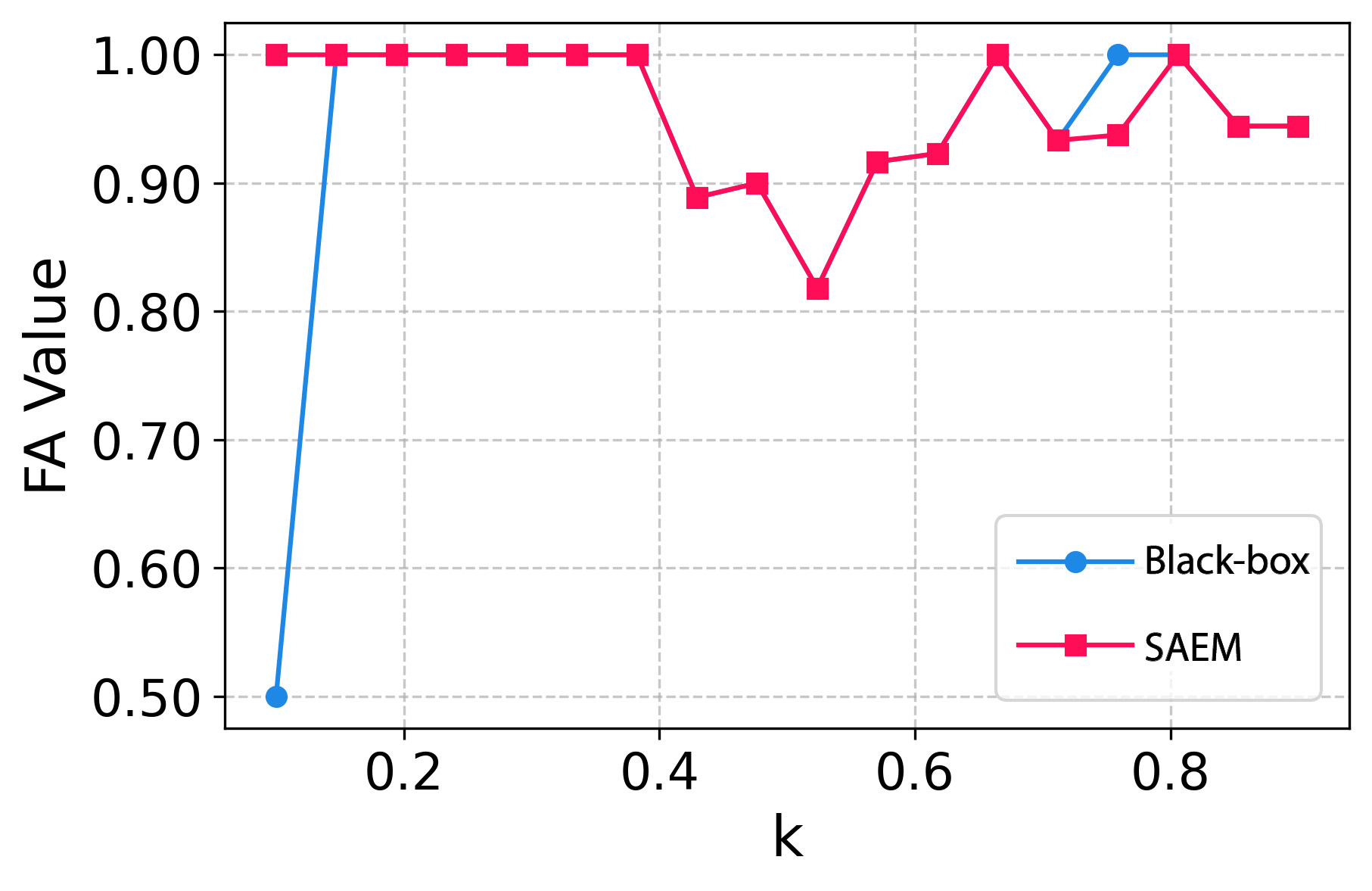}
        % \caption{Before}
        \end{subfigure}%
        \hfill
        \begin{subfigure}[t]{0.24\textwidth}
        \centering
        \includegraphics[width=\textwidth]{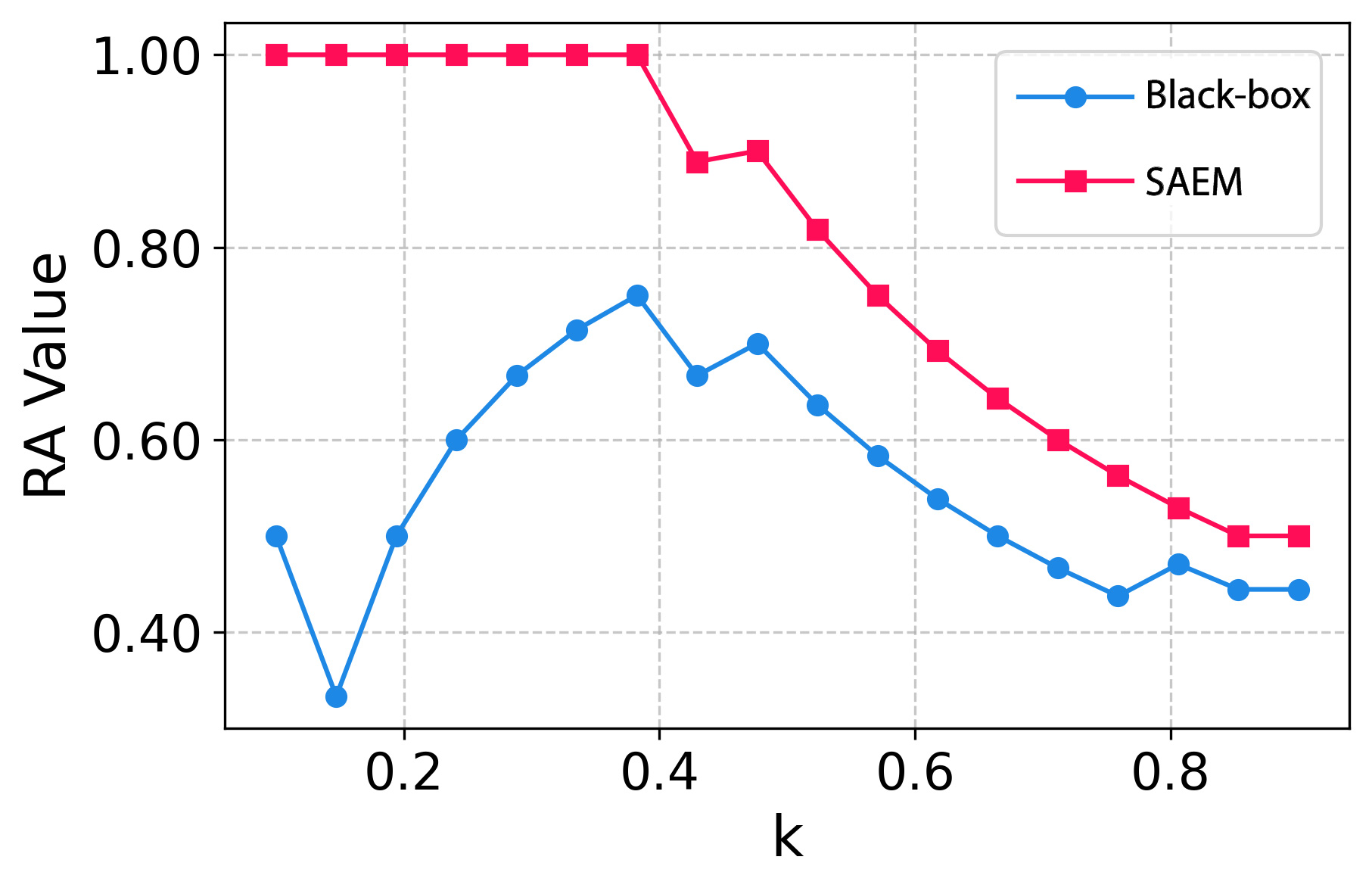}
        % \caption{After}
        \end{subfigure}
        \hfill
        \begin{subfigure}[t]{0.24\textwidth}
        \centering
        \includegraphics[width=\textwidth]{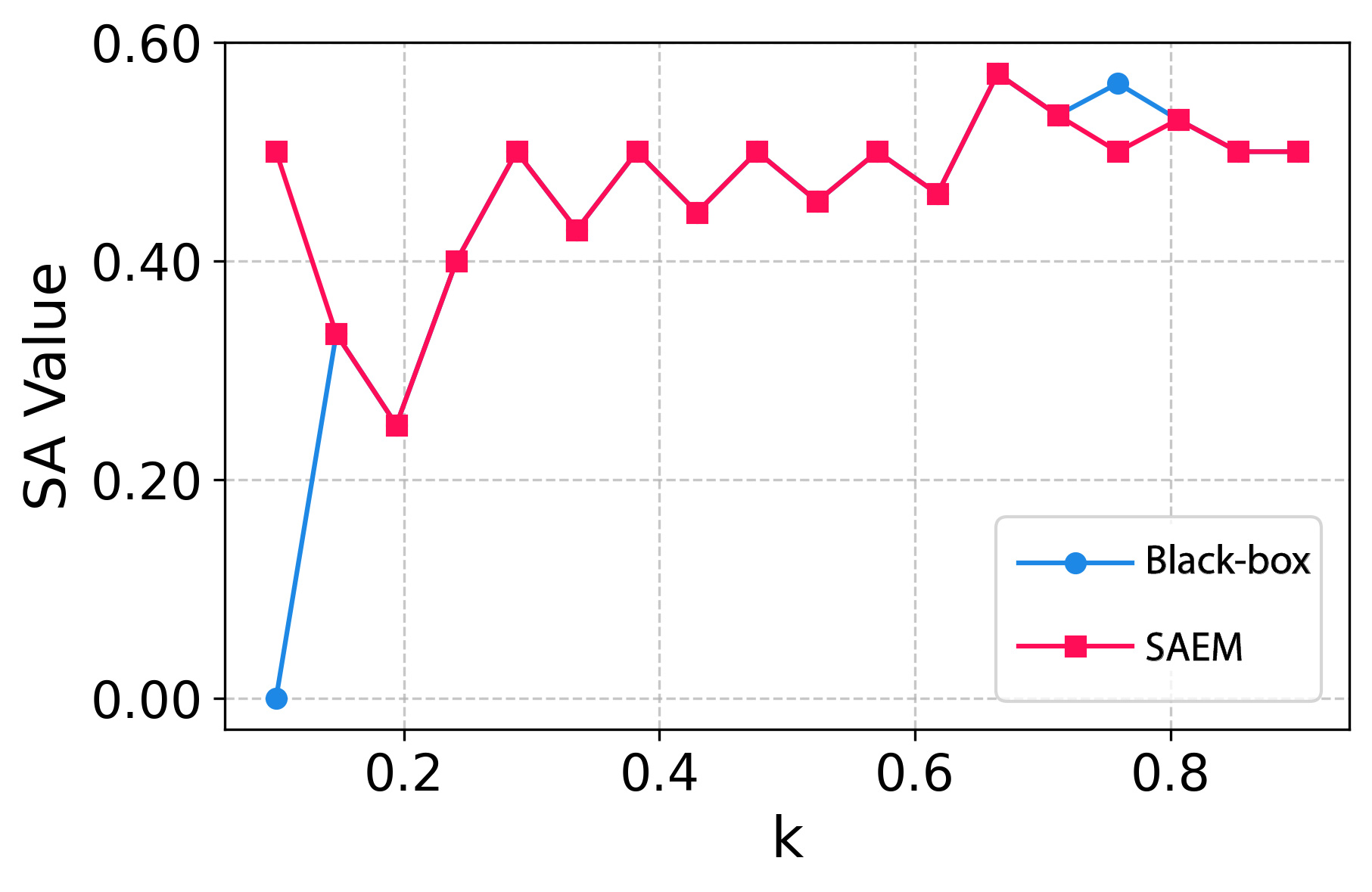}
        % \caption{After}
        \end{subfigure}
        \hfill
        \begin{subfigure}[t]{0.24\textwidth}
        \centering
        \includegraphics[width=\textwidth]{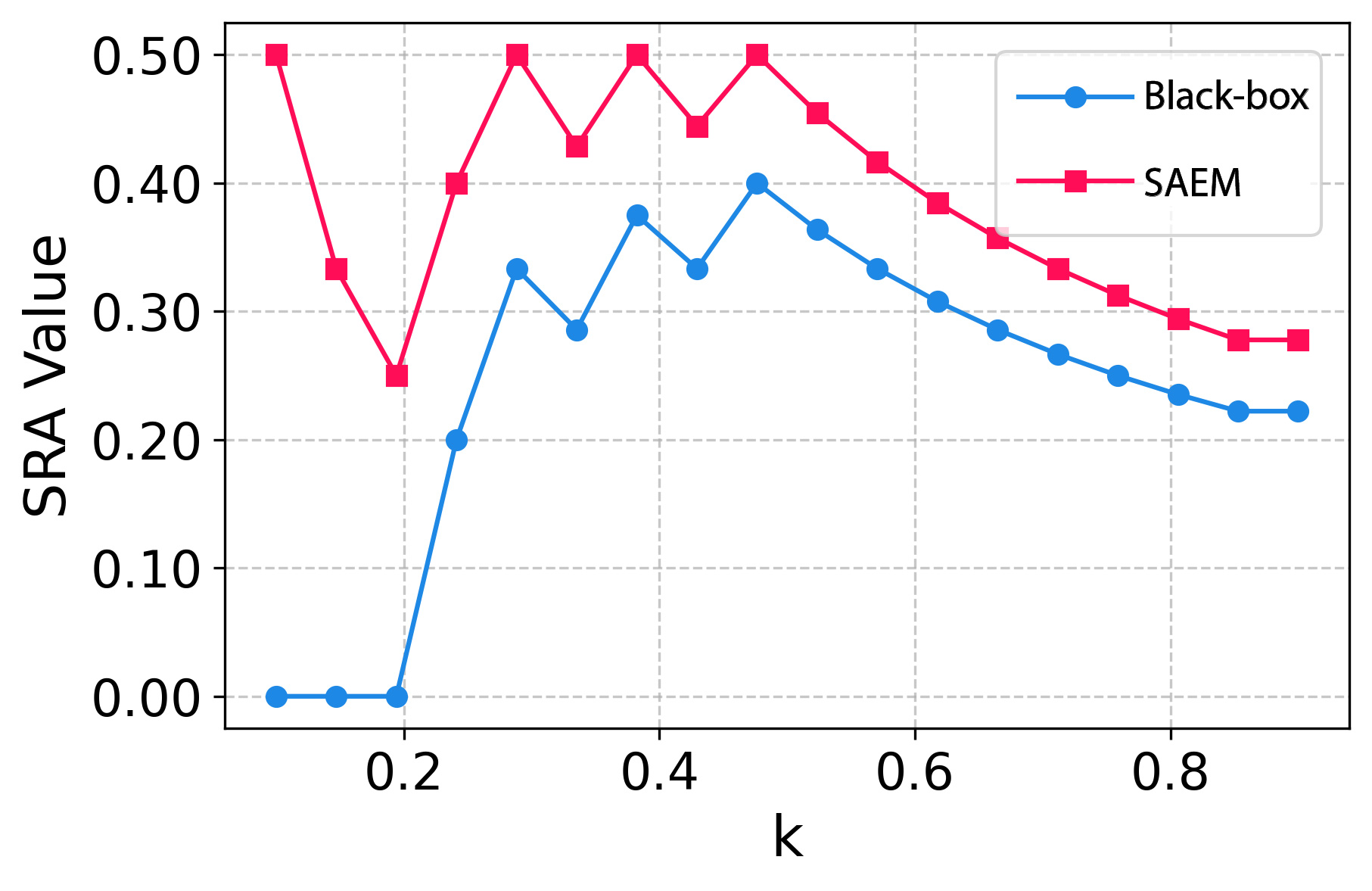}
        \end{subfigure}
        \begin{subfigure}[t]{0.24\textwidth}
            \centering
            \includegraphics[width=\textwidth]{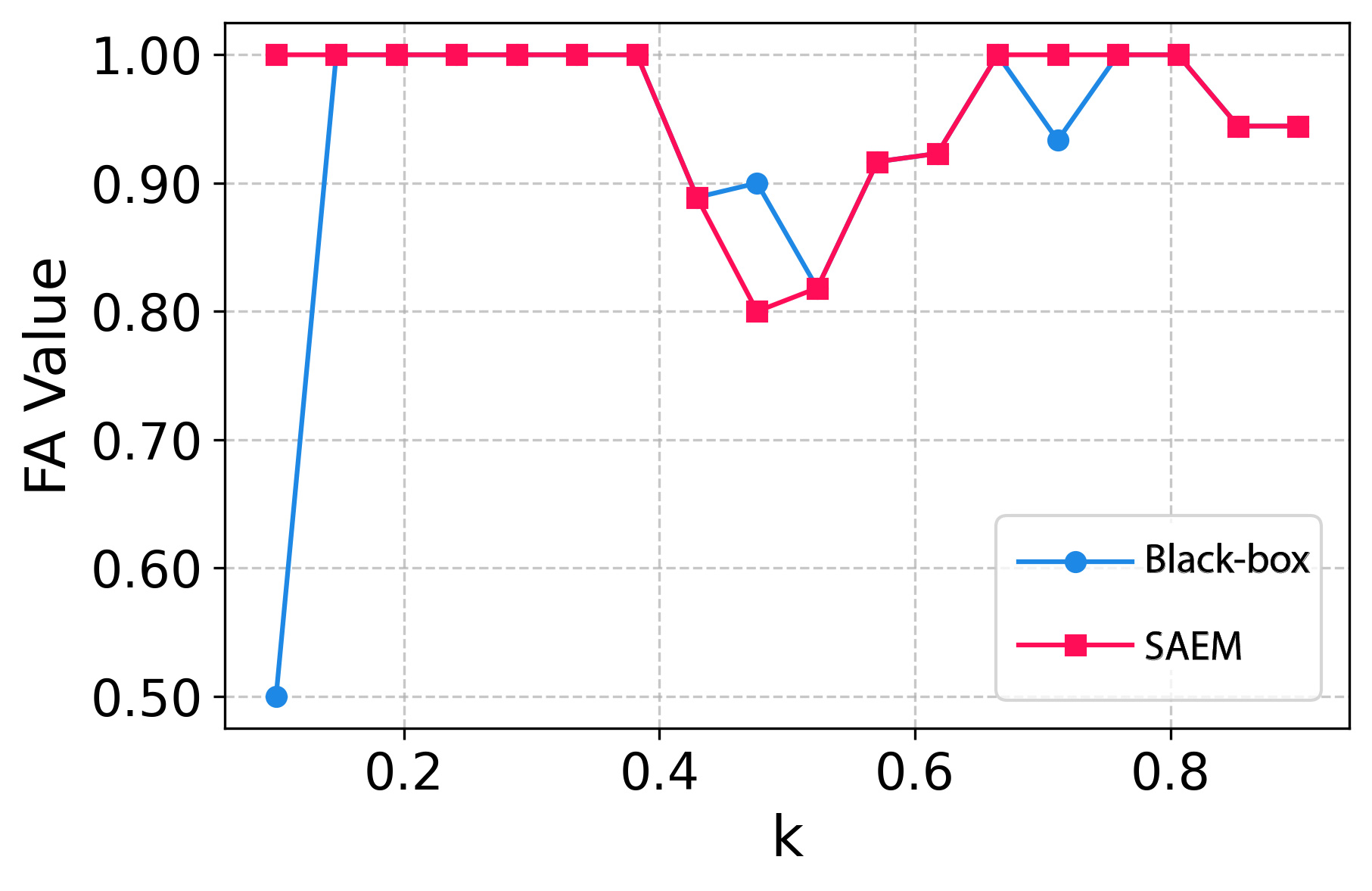}
            % \caption{FA improvement on Ground Truth}
        \end{subfigure}%
        \hfill
        \begin{subfigure}[t]{0.24\textwidth}
            \centering
            \includegraphics[width=\textwidth]{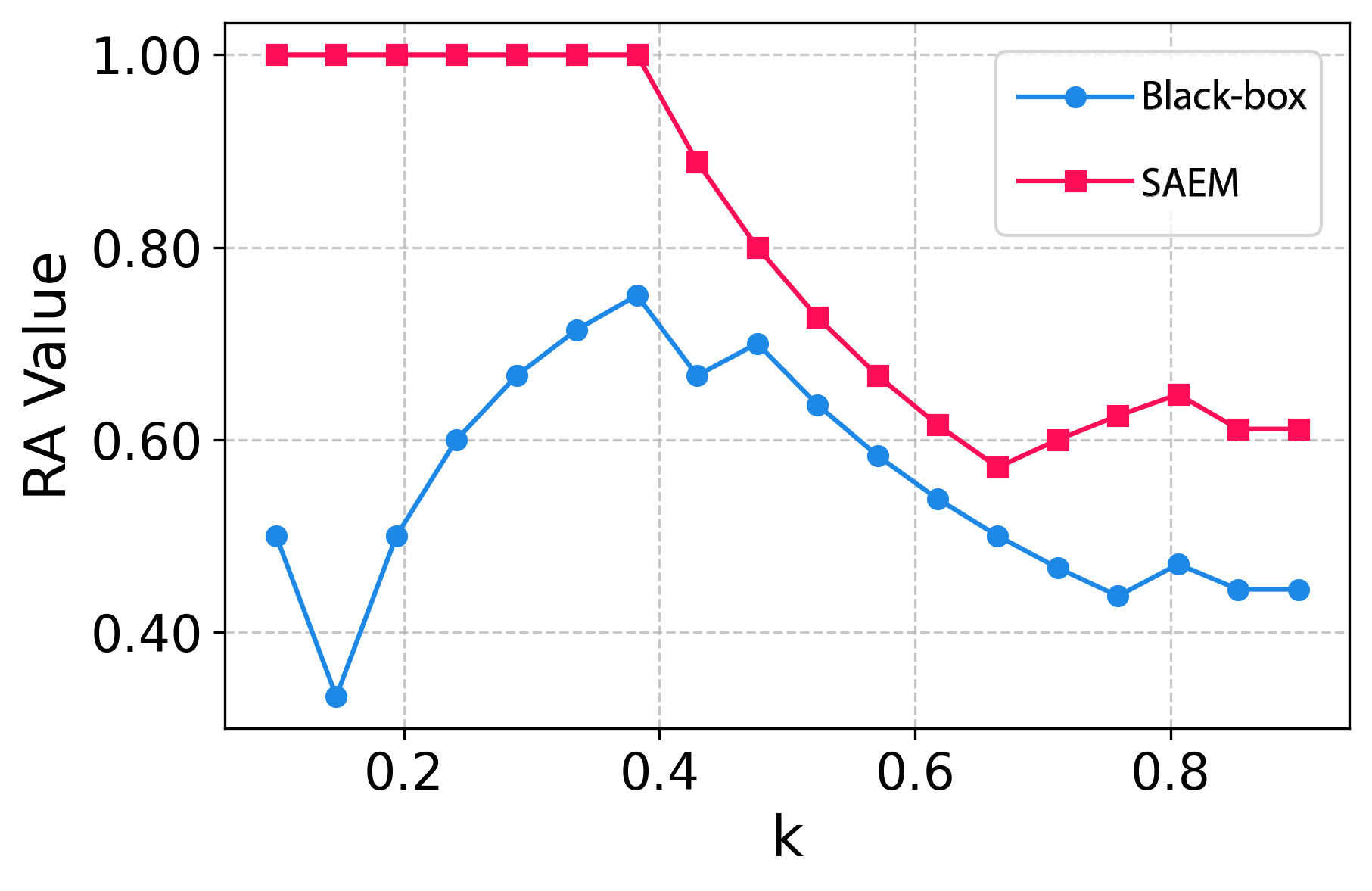}
            % \caption{RA improvement on Ground Truth}
        \end{subfigure}
        \hfill
        \begin{subfigure}[t]{0.24\textwidth}
            \centering
            \includegraphics[width=\textwidth]{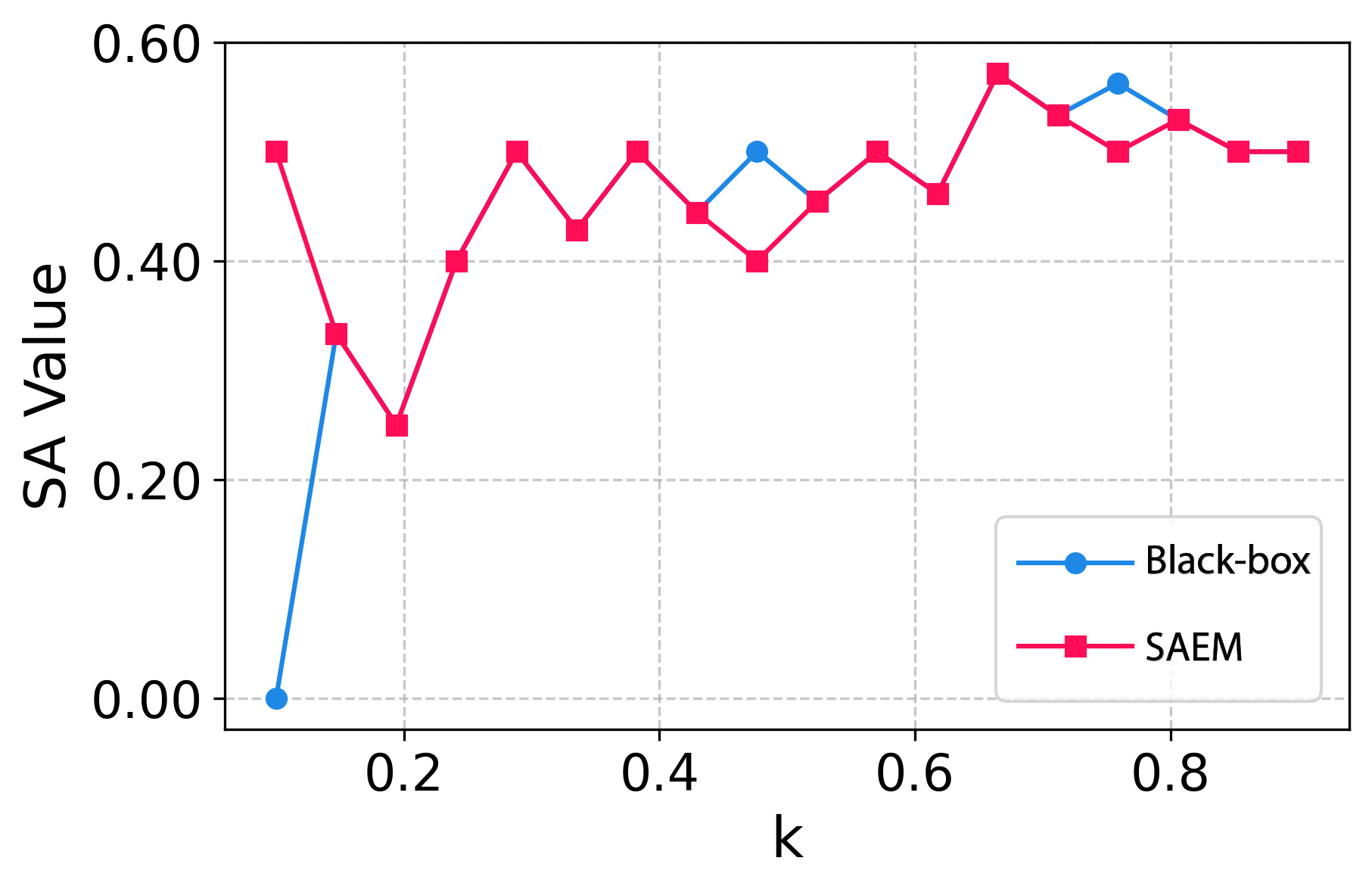}
            % \caption{SA improvement on Ground Truth}
        \end{subfigure}
        \hfill
        \begin{subfigure}[t]{0.24\textwidth}
            \centering
            \includegraphics[width=\textwidth]{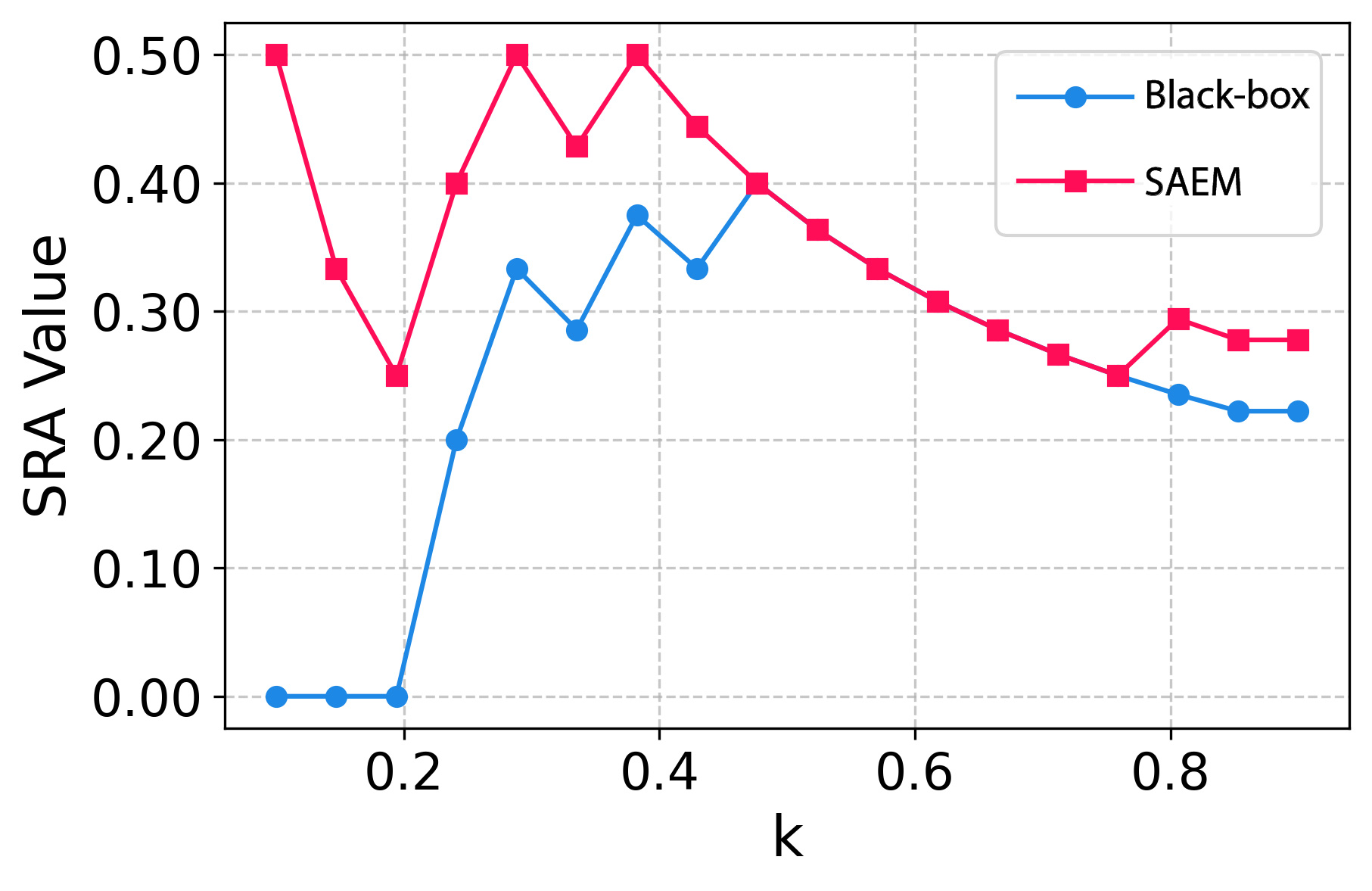}
            % \caption{SRA improvement on Ground Truth}
        \end{subfigure}
    \caption{Synthetic Dataset with model LR}
    \end{subfigure}
    \vfill
        \begin{subfigure}{\textwidth}
        \centering
       \begin{subfigure}[t]{0.24\textwidth}
        \centering
        \includegraphics[width=\textwidth]{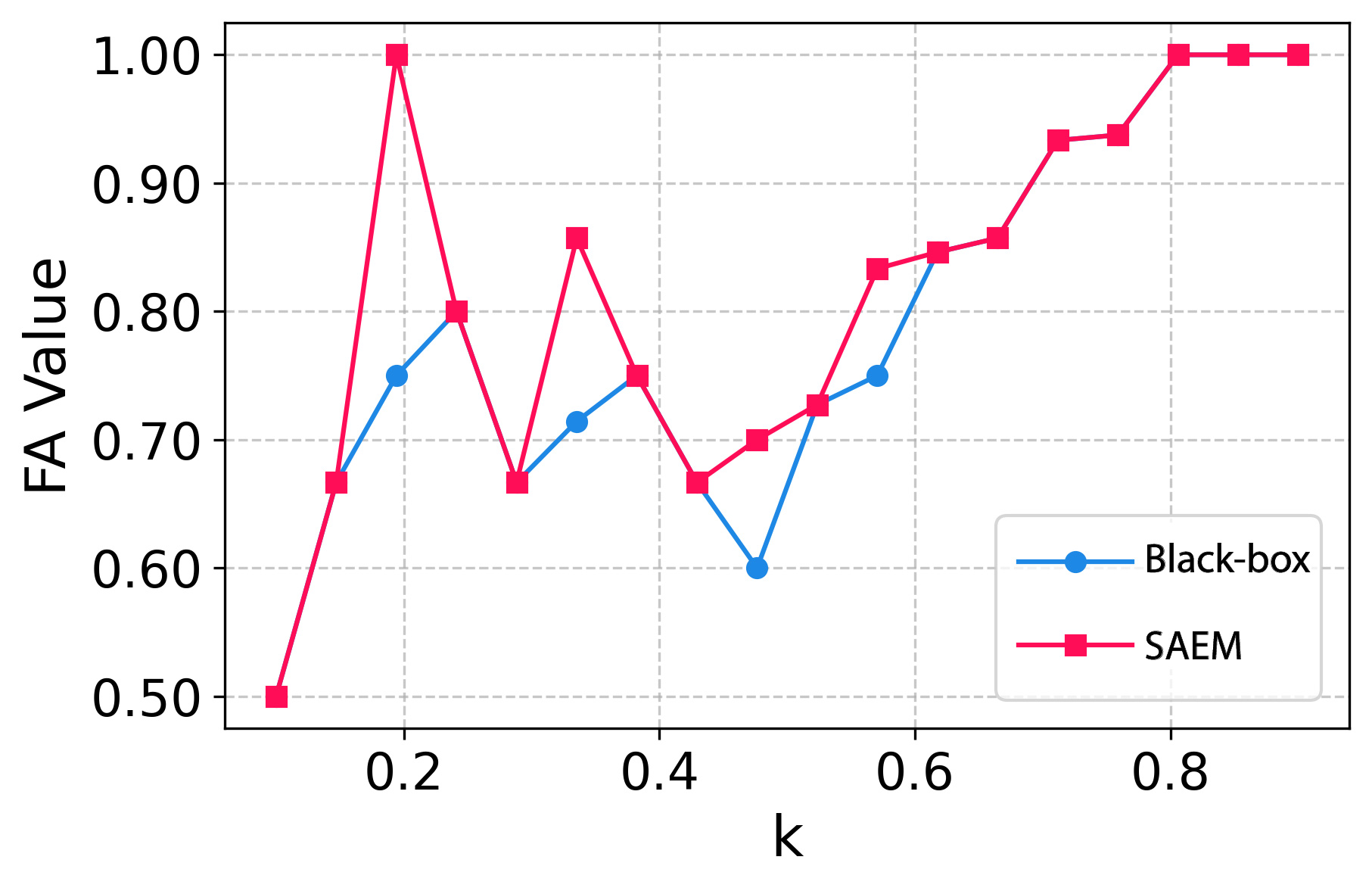}
        % \caption{FA improvement on Ground Truth}
        \end{subfigure}%
        \hfill
        \begin{subfigure}[t]{0.24\textwidth}
        \centering
        \includegraphics[width=\textwidth]{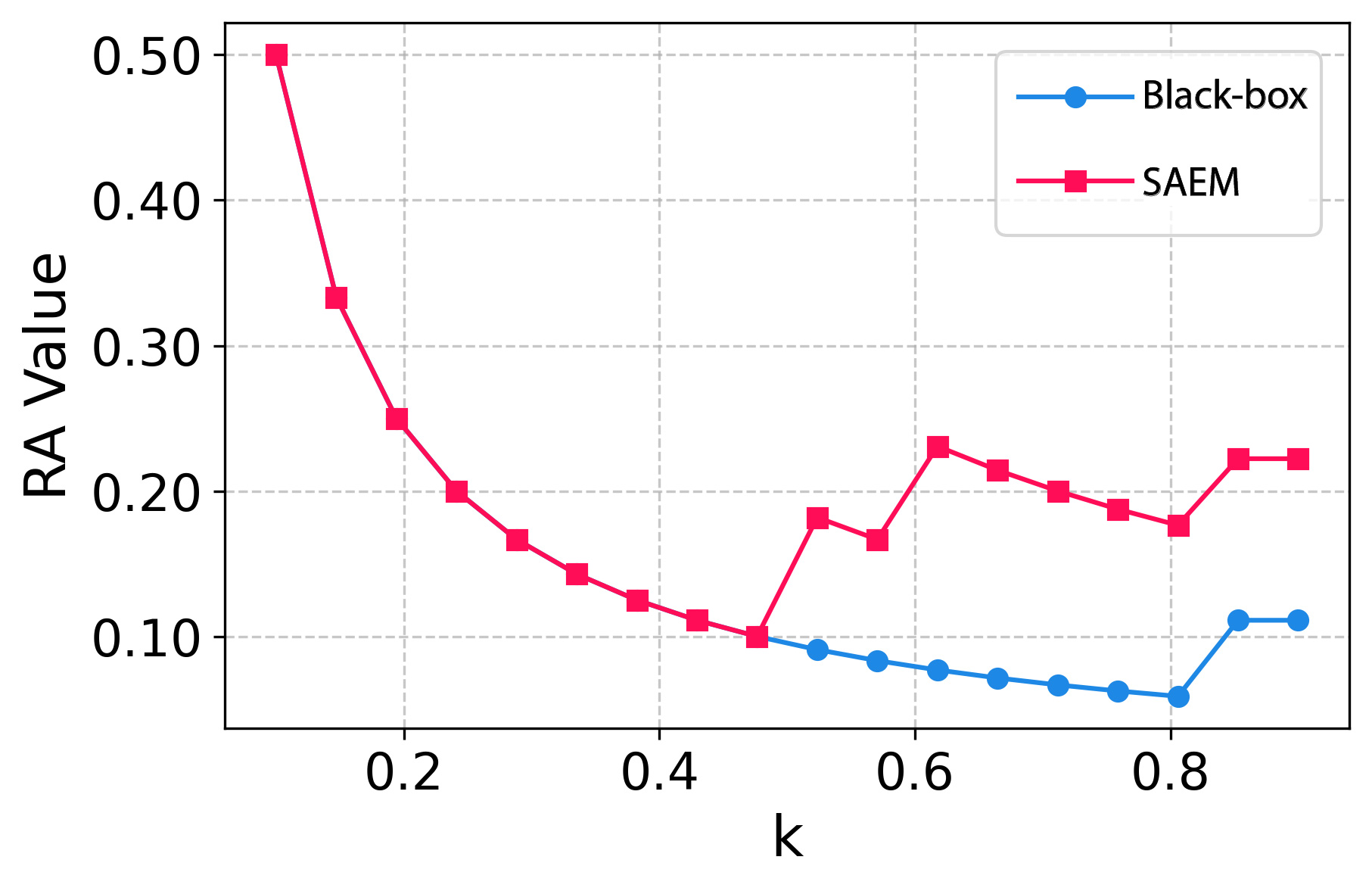}
        % \caption{RA improvement on Ground Truth}
        \end{subfigure}
        \hfill
        \begin{subfigure}[t]{0.24\textwidth}
        \centering
        \includegraphics[width=\textwidth]{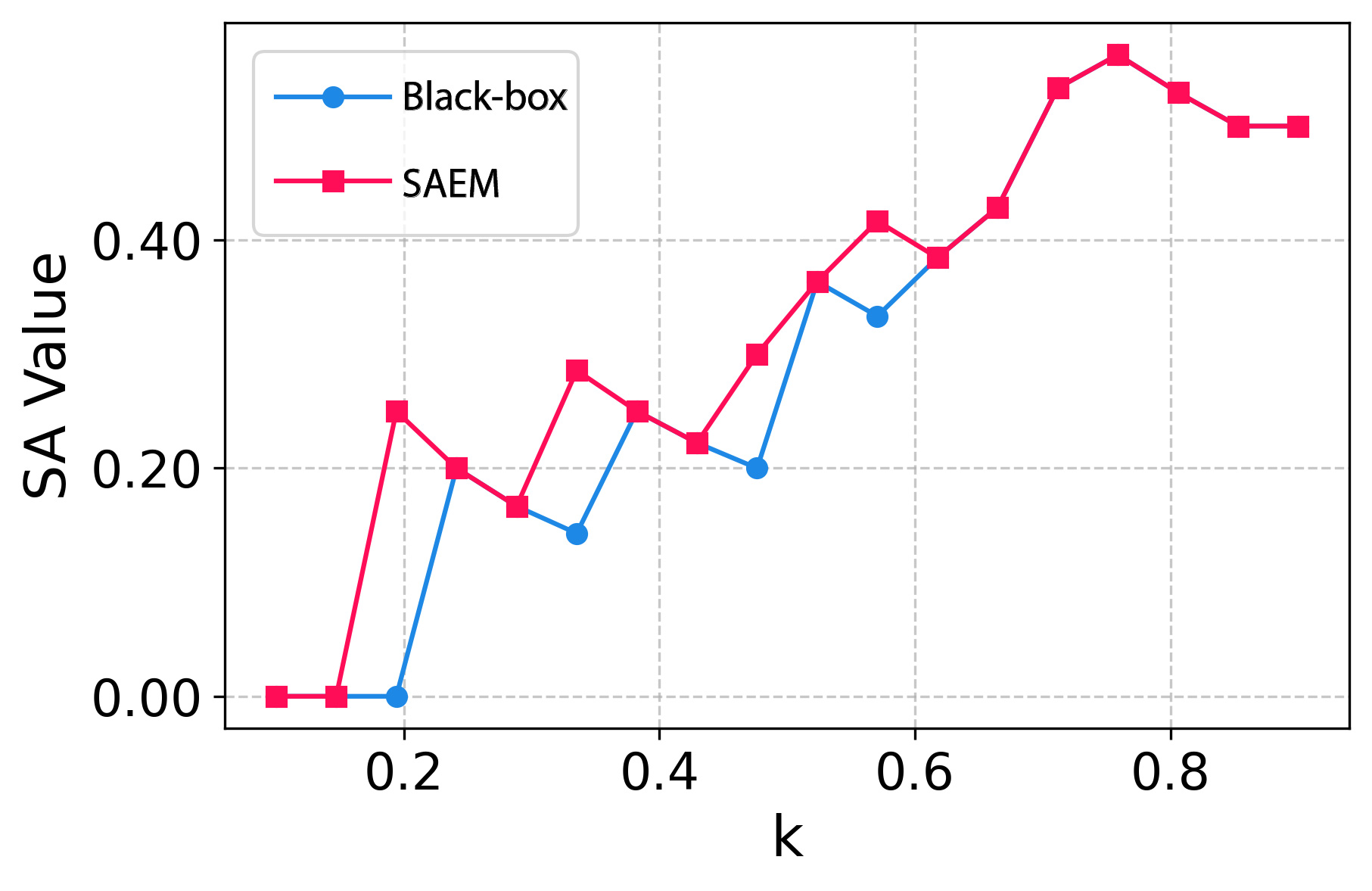}
        % \caption{SA improvement on Ground Truth}
        \end{subfigure}
        \hfill
        \begin{subfigure}[t]{0.24\textwidth}
        \centering
        \includegraphics[width=\textwidth]{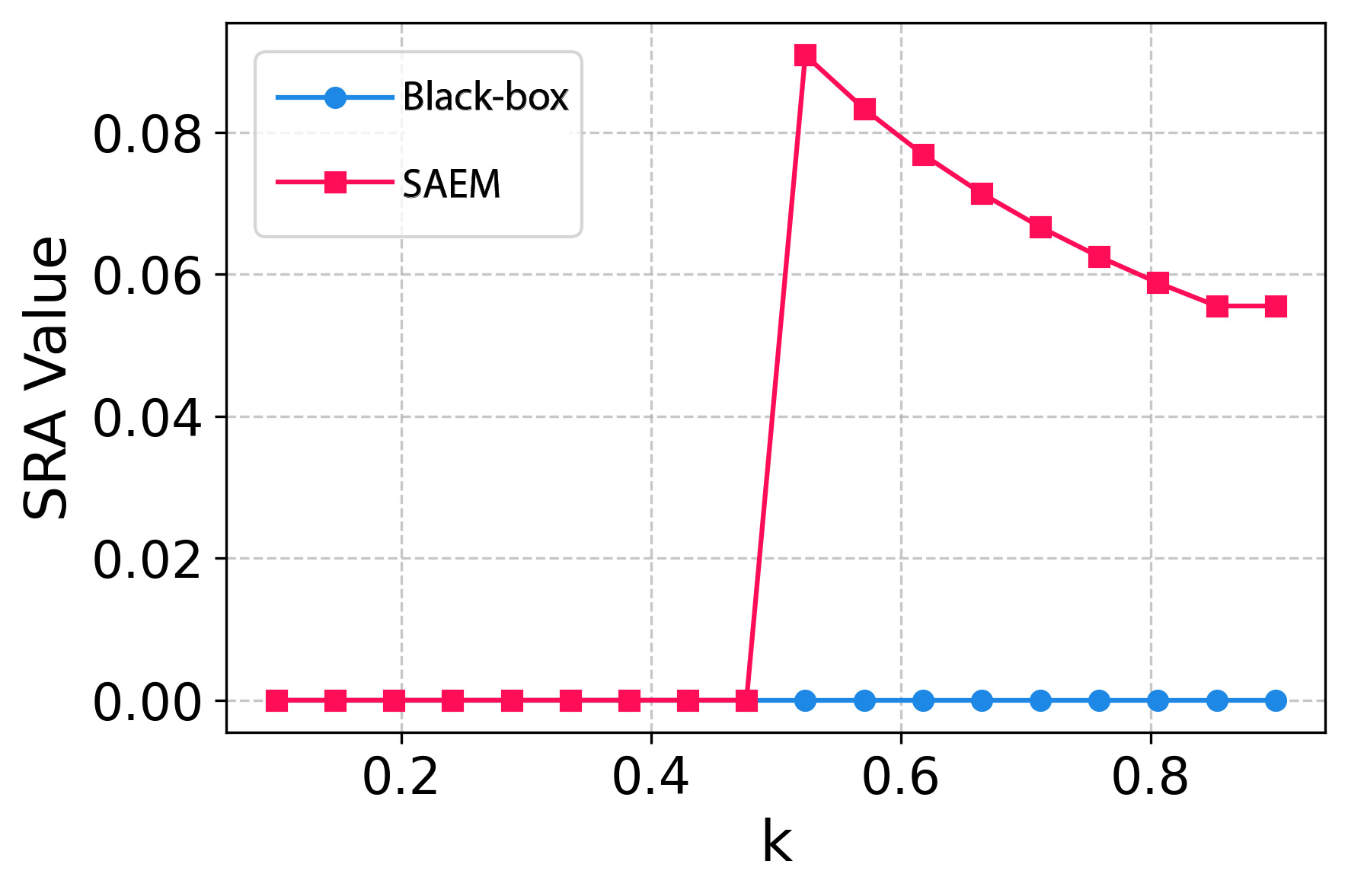}
        % \caption{SRA improvement on Ground Truth}
        \end{subfigure}
        \vfill
        \begin{subfigure}[t]{0.24\textwidth}
        \centering
        \includegraphics[width=\textwidth]{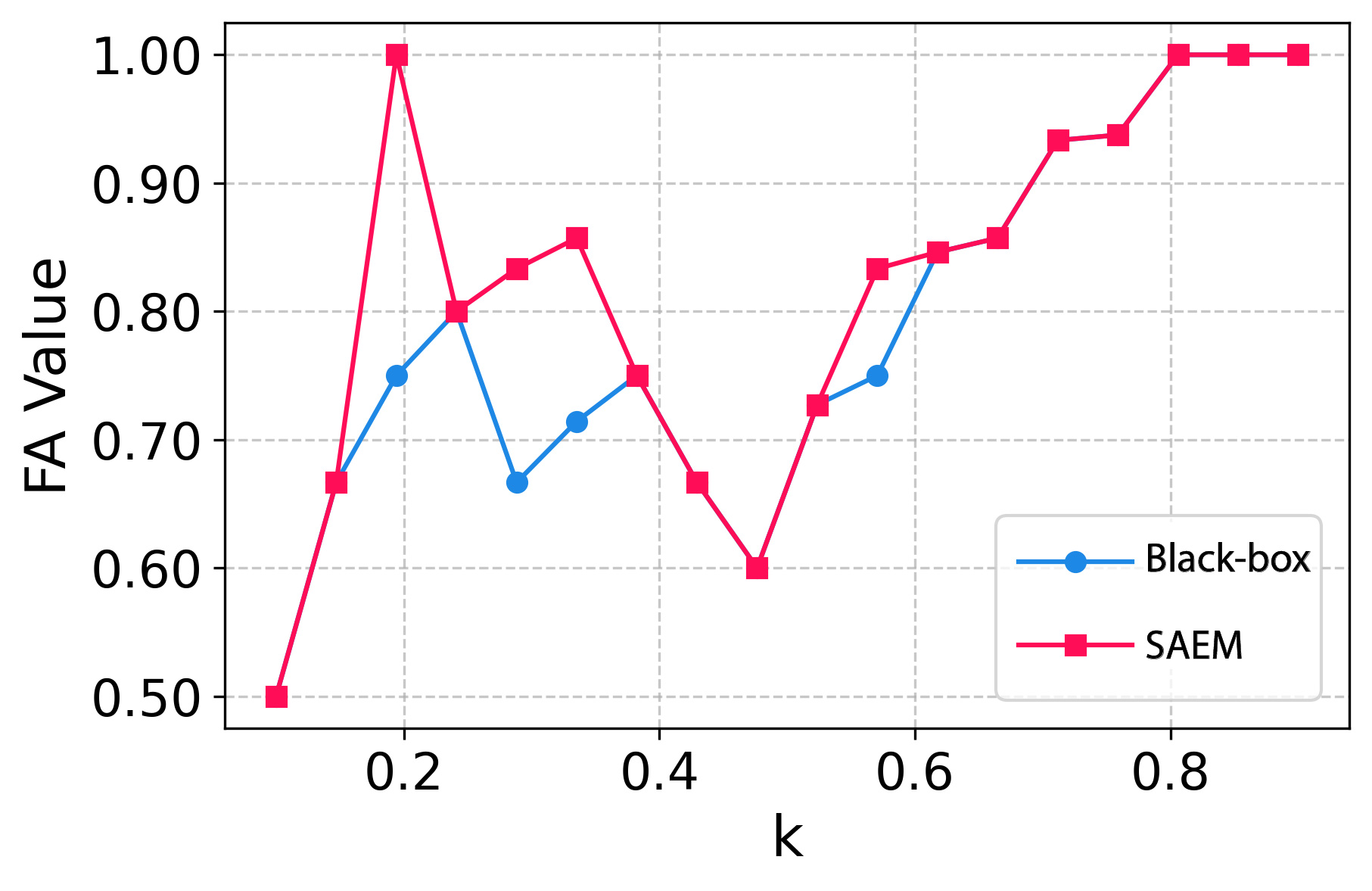}
        % \caption{Before}
        \end{subfigure}%
        \hfill
        \begin{subfigure}[t]{0.24\textwidth}
        \centering
        \includegraphics[width=\textwidth]{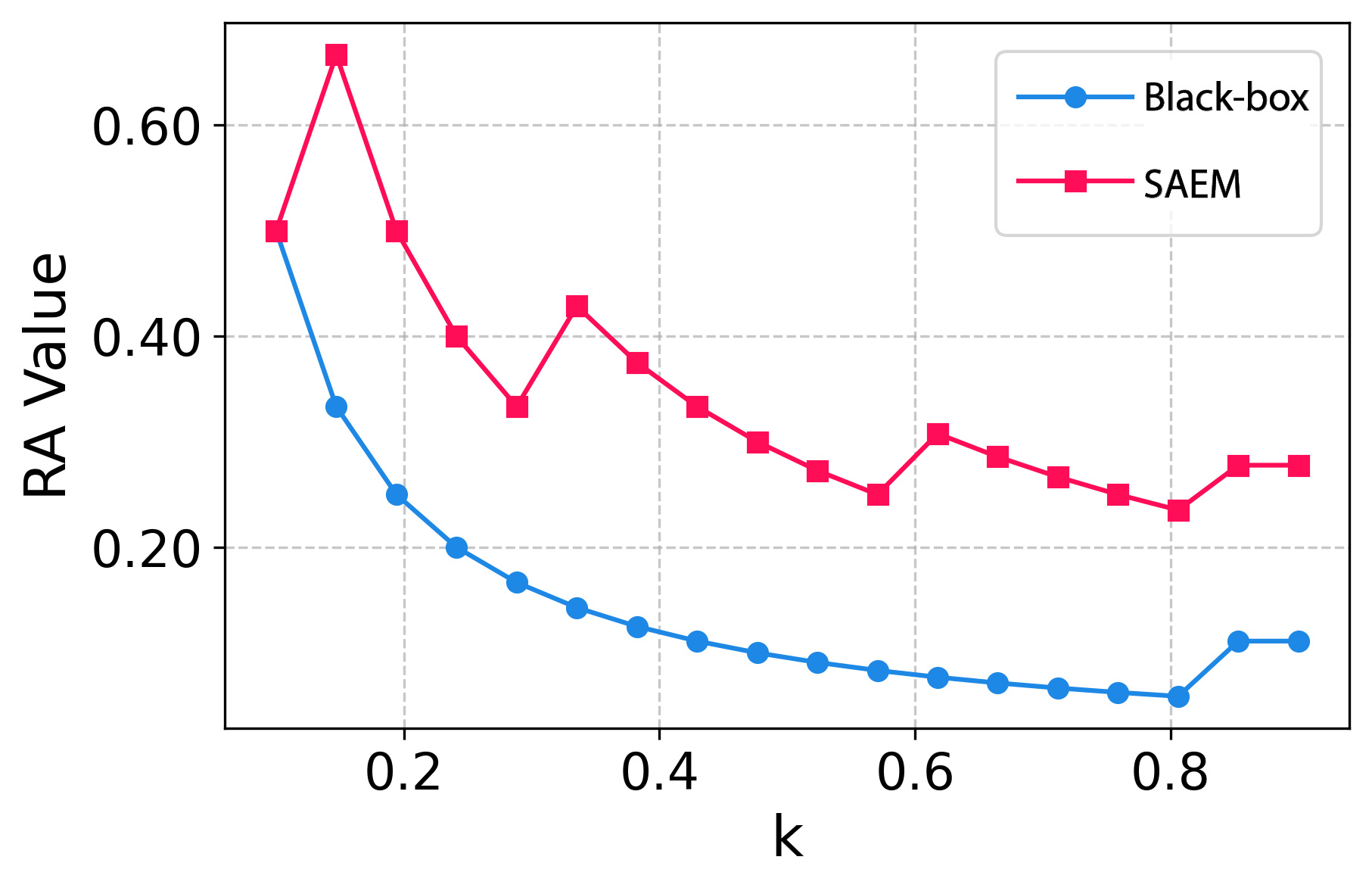}
        % \caption{After}
        \end{subfigure}
        \hfill
        \begin{subfigure}[t]{0.24\textwidth}
        \centering
        \includegraphics[width=\textwidth]{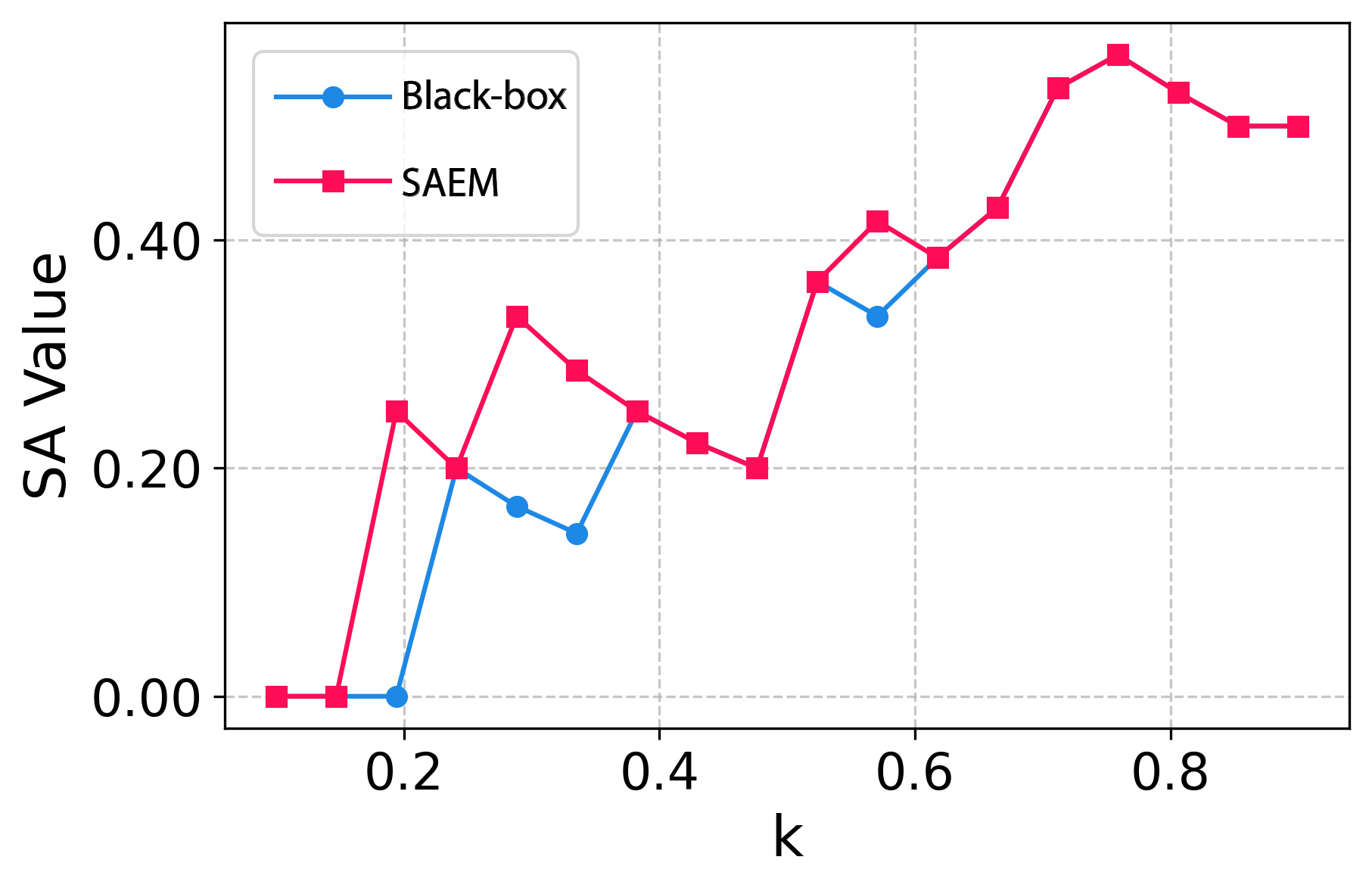}
        % \caption{After}
        \end{subfigure}
        \hfill
        \begin{subfigure}[t]{0.24\textwidth}
        \centering
        \includegraphics[width=\textwidth]{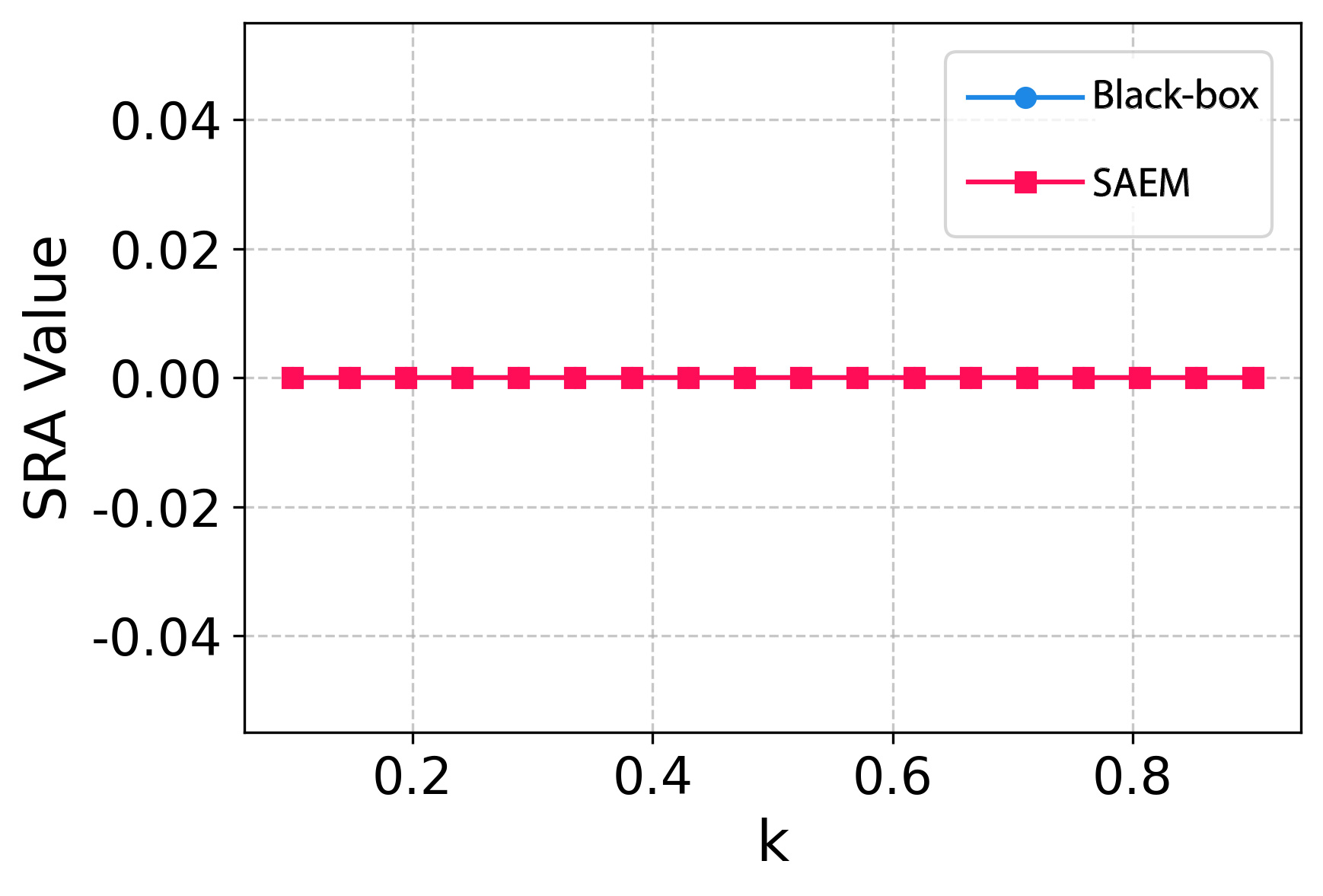}
        \end{subfigure}
        \begin{subfigure}[t]{0.24\textwidth}
            \centering
            \includegraphics[width=\textwidth]{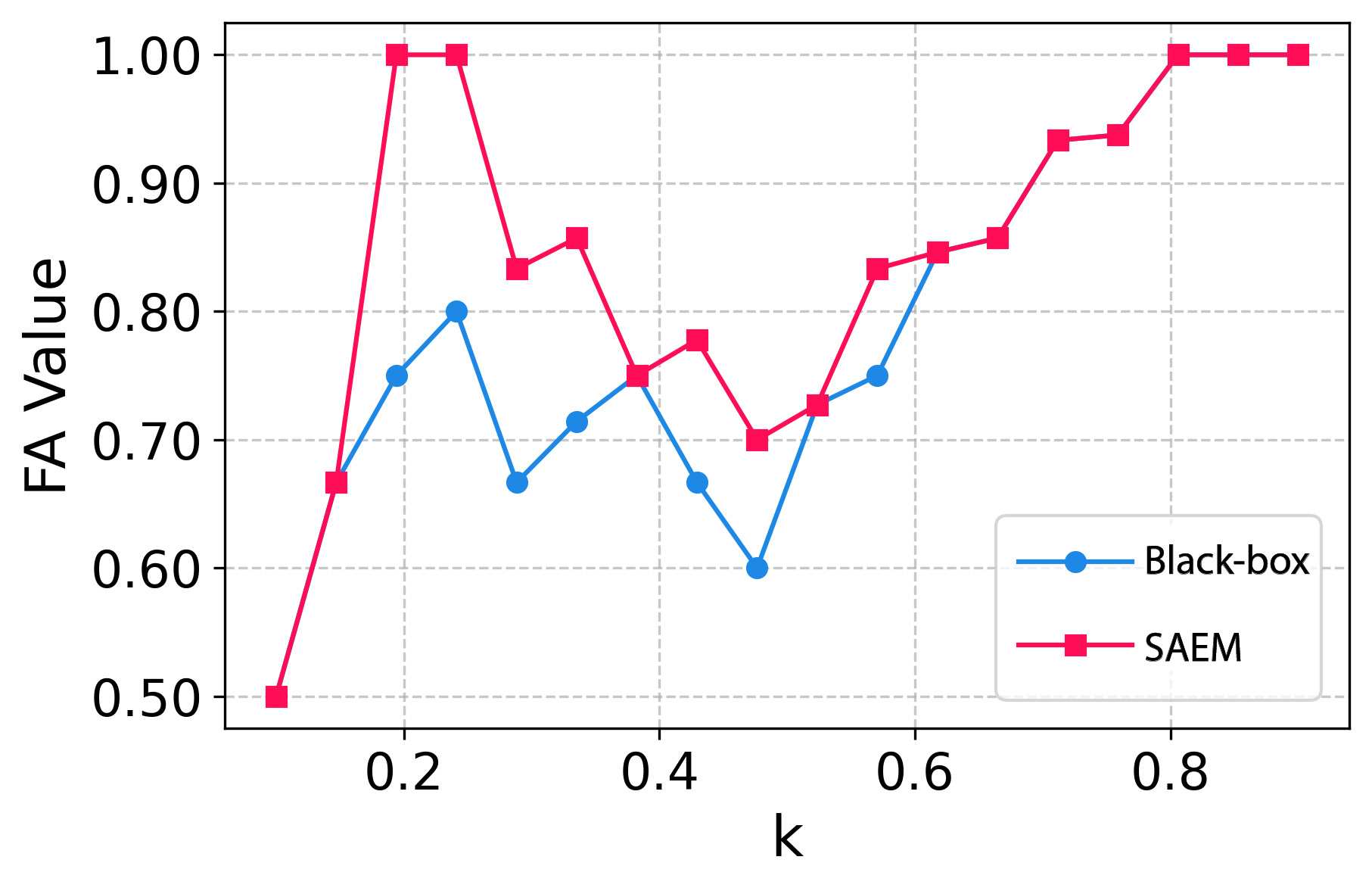}
            % \caption{FA improvement on Ground Truth}
        \end{subfigure}%
        \hfill
        \begin{subfigure}[t]{0.24\textwidth}
            \centering
            \includegraphics[width=\textwidth]{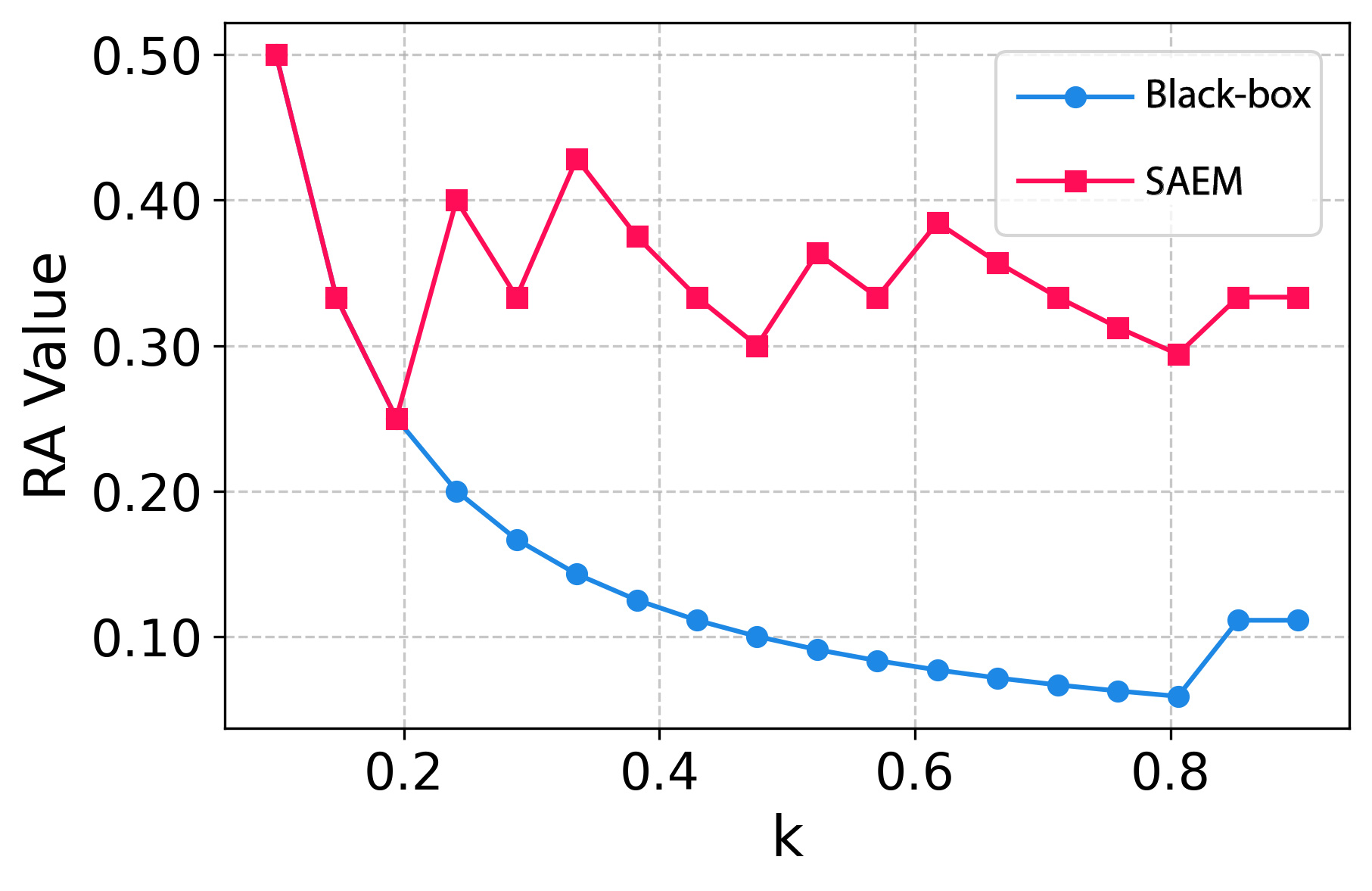}
            % \caption{RA improvement on Ground Truth}
        \end{subfigure}
        \hfill
        \begin{subfigure}[t]{0.24\textwidth}
            \centering
            \includegraphics[width=\textwidth]{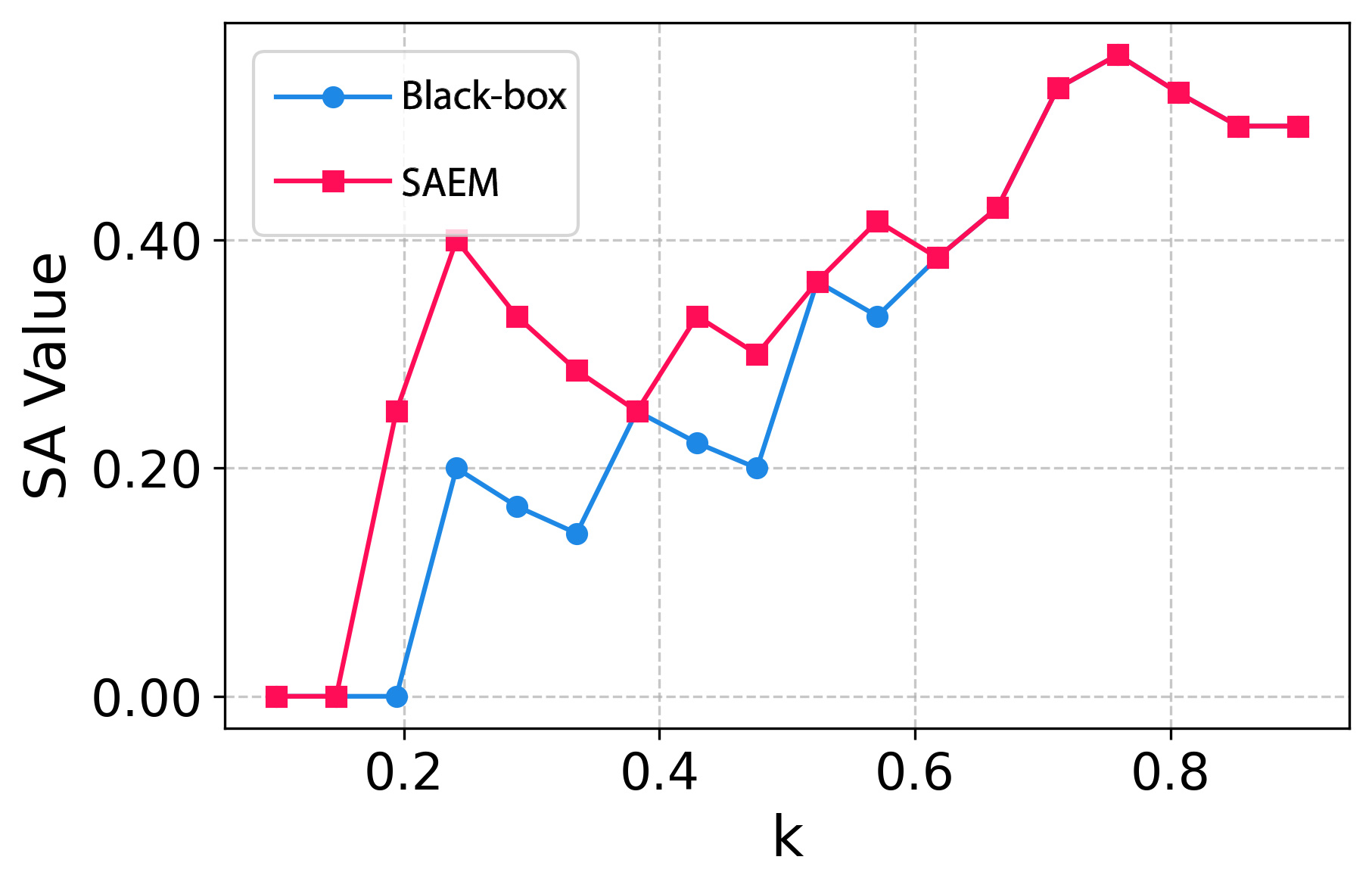}
            % \caption{SA improvement on Ground Truth}
        \end{subfigure}
        \hfill
        \begin{subfigure}[t]{0.24\textwidth}
            \centering
            \includegraphics[width=\textwidth]{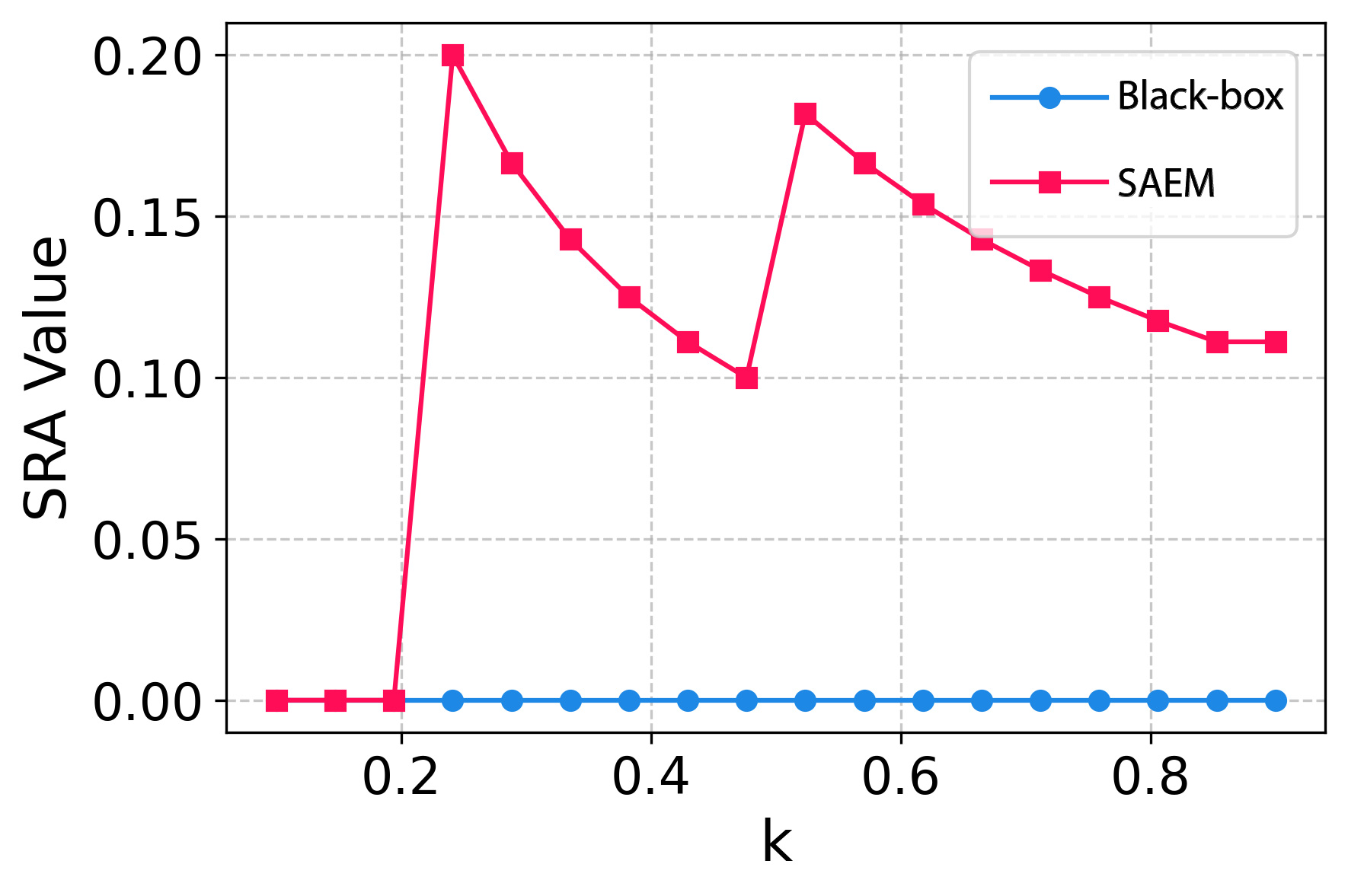}
            % \caption{SRA improvement on Ground Truth}
        \end{subfigure}
        \caption{Synthetic Dataset with model ANN}
    \end{subfigure}
\caption{\label{fig:ablation}Comparison of agreement metrics (FA, RA, SA, SRA) between black box models and their corresponding SAEM for varying $\epsilon$ values on the Synthetic dataset. From top to bottom, $\epsilon$ is 0.05, 0.1, and 0.2, respectively.}
\end{figure*}

\textbf{Results and Discussion}:
As illustrated in Fig. \ref{fig:ablation}, increasing $\epsilon$ generally enhances EXAGREE's capacity to identify models with improved explanation agreement. This is evidenced by consistent improvements in FA and RA metrics across all $k$ values as $\epsilon$ increases for both LR and ANN models. 
However, the results for SA and SRA are not substantially improved:
For the LR model, SA and SRA show mixed results, with improvements at certain $k$ values but not others. For the ANN model, all metrics, including SA and SRA, demonstrate clear improvements as $\epsilon$ increases from 0.05 to 0.2, except SRA when $\epsilon=0.1$.

These findings generally support the hypothesis that larger Rashomon sets facilitate better stakeholder alignment. However, the lack of improvement in some cases (e.g., SRA in Fig. \ref{fig:ablation} (a)) aligns with the work of \citet{li2024practical}, which suggests that a larger Rashomon set does not necessarily guarantee a greater range of feature attributions. This observation potentially explains the uneven improvements in agreement metrics.
In practice, decision-makers must carefully consider the trade-off between performance tolerance (determined by $\epsilon$) and the increased opportunities to find SAEMs. 

\section{Summary of Information}\label{sec:summary_of_information}

We summarized notations used in the work, detained datasets information, pseudocode of the algorithm, and evaluation metrics in this section. 
\begin{table*}[hpb!]
\small
\centering
\caption{Summary of Evaluation Metrics From OpenXAI \cite{agarwal2022openxai}}
\label{tab:metrics_summary}
\begin{tabular}{p{6cm}p{7.5cm}}
\toprule
\textbf{Metric} & \textbf{Description} \\
\midrule
Feature Agreement (FA) & Measures agreement in feature importance \\

Rank Agreement (RA) & Assesses agreement in feature ranking \\

Sign Agreement (SA) & Evaluates agreement in feature attribution signs \\

Signed Rank Agreement (SRA) & Combines sign and rank agreement \\

Pairwise Rank Agreement (PRA) & Measures pairwise rank consistency \\

Rank Correlation (RC) & Quantifies correlation between feature rankings \\
\midrule
Prediction Gap on Important feature perturbation (PGI) & Measures Impact of perturbing important features on model predictions \\

Prediction Gap on Unimportant feature perturbation (PGU) & Measures impact of perturbing unimportant features on model predictions \\
\midrule
Fairness & Compares all above metric values across majority and minority subgroups \\ \bottomrule

\end{tabular}
\end{table*}

\begin{algorithm*}
\caption{EXAGREE Framework Pseudocode (from stage 2)}
\begin{algorithmic}[1]
\REQUIRE A task $\D_{task}$ and different requirements $\{i \in \mathcal{S} \mid \bfs{r}^{i}\}$ from stakeholders $\mathcal{S}$. 
\STATE Initialize pre-trained $f_{\text{ref}}$, $f_{\text{dman}}$, and $f_{\text{diffsort}}$.
\FOR{$i = 0, 1, 2, \cdots$ \COMMENT{Find solutions for each stakeholder group}}
\STATE  $\bfs{r}^{\text{target}} \gets f_{\text{diffsort}}(|\bfs{a}^{i}|)$, 
$\text{sign}(\bfs{a}^{i}) \gets \bfs{a}^{i}$
\STATE $\bfs{r}^{\text{ref}} \gets f_{\text{diffsort}}(|\bfs{a}|)$ \COMMENT{Compute target and reference rankings}
\STATE $\bfs{p}_{\text{top\_k}} \gets$ Sort$(\bfs{r}^{\text{ref}}, k)$ 
\STATE $\bfs{p}_{\text{diff\_k}} \gets$ Diff$(\bfs{r}^{\text{ref}}, \bfs{r}^{\text{target}}, k)$
\COMMENT{Identify important features that make differences}
\STATE  $\bfs{M} \gets$ initialize\_mask$(h, \bfs{p}_{\text{top\_k}}, \bfs{p}_{\text{diff\_k}})$ \COMMENT{Initialize $h$ masks with attentions}
\STATE
\COMMENT{Note: each head corresponds to a mask, and all masks are validated in the initialization}
% \STATE all\_heads.initializer(diff\_positions)
% \STATE attention\_weights.initializer(topk\_positions)
% \STATE Initialize optimizer and scheduler
% \STATE MultiHeadResMaskNet.train()
\FOR{epoch = $0, 1, \cdots, num\_epochs$}
    % \STATE optimizer.zero\_grad()
    \STATE $\bfs{M}$.update()
    \COMMENT{Update all masks with states and keep a record of valid masks}
    \STATE $\bfs{i}_{\text{active}}$ $\gets \bfs{M}.states$
    \COMMENT{Ensure masks within the $\R_{\epsilon}$}
    % \STATE active\_heads $\gets$ validate\_mask(all\_heads) \COMMENT{check validity of in the $\R$}
    % \If{active\_heads is empty} \COMMENT{stop if all heads inactive}
    %     \STATE break\algorithmiccomment{This is a comment}
    % \EndIf
    \FOR{$\bfs{m} \in \bfs{M}_{\text{active}}$ \COMMENT{ Update the active masks only}}
        \STATE $\bfs{a}$ $\gets$ $f_{\text{dman}}(\bfs{m})$ \COMMENT{Approximate attributions of features}
        \STATE $\bfs{r}^{\bfs{m}, \varphi} \gets f_{\text{diffsort}}(|\bfs{a}|), \text{sign}(\bfs{a}) \gets \bfs{a}$
        \COMMENT{Calculate the ranking from DiffSortNet}
        \STATE $l_{\text{rank}}$ $\gets$ spearman\_cor($\bfs{r}^{\bfs{m}, \varphi}, \bfs{r}^{\text{target}}$)
        \COMMENT{Calculate the Spearman's correlation}
        \STATE $l_{\text{sign}}$ $\gets \text{MSE}$($\text{sign}(\bfs{a}), \text{sign}(\bfs{a}^{i})$)
        \COMMENT{Calculate loss on sign}
        \STATE $\bfs{l}_{\text{rank}}$.append($l_{\text{rank}}$) 
        \STATE $\bfs{L}_{\text{sign}}$.append($l_{\text{sign}}$)
    \ENDFOR
    \STATE $\Loss_{\text{rank}}$ $\gets$ Avg($\bfs{l}_{\text{rank}}$), $\Loss_{\text{sign}}$ $\gets$ Avg($\bfs{L}_{\text{sign}}$)
    \STATE $\Loss_{\text{sparsity}}$, $\Loss_{\text{diversity}}$ $\gets$ Avg(norm($\bfs{M}_{\text{active}}$), dim=0)), Avg(norm($\bfs{M}_{\text{active}}$), dim=1))
    % \STATE $\Loss_{\text{diversity}}$ $\gets$ sum(norm($\bfs{M}_{\text{active}}$), dim=1))
    \STATE $\Loss_{\text{total}}$ $\gets$ $\Loss_{\text{rank}}$ + $\Loss_{\text{sign}}$ + $\lambda_{s}\times$ $\Loss_{\text{sparsity}}$ + $ \lambda_{d}\times$ $\Loss_{\text{diversity}}$ 
    \STATE $\Loss_{\text{total}}$.backward()
    \STATE $\bfs{M}_{\text{active}}$.update()
    \COMMENT{Update active masks only}
\ENDFOR
\STATE $\bfs{M}_{\text{saved}}$.evaluate() \COMMENT{Calculate true attributions and evaluate according to OpenXAI metrics}
\STATE $\bfs{m}^{i}$ = $\bfs{M}_{\text{saved}}$.solution() \\
\COMMENT{Found model $M = \bfs{m}^{i} \circ f_{\text{ref}}$ that improve the overall faithfulness for the stakeholder group $i$}
\STATE 
\ENDFOR
\end{algorithmic}
\label{alg:a1}
\end{algorithm*}

\subsection{Hyperparameter and Model Structure Summary}
We used pre-trained LR and ANN models from OpenXAI \citep{agarwal2022openxai}. The GRS sampling used an epsilon rate of 0.05 with log loss \citep{li2024practical}. Our DMAN used two hidden layers of 100 units each, with input and output sizes matching the feature length. The DiffSortNet used a bitonic sorting network with Cauchy interpolation. The MHMN employed 50 heads. Both DMAN and MHMN used Adam optimizer with learning rate scheduling. We also included a simple DT as a global surrogate explanation. An overall summary of hyperparameters is provided in Table \ref{tab:hyper}. It is important to note that while this configuration yielded robust results across most of our tested datasets, it may not universally produce optimal outcomes for all scenarios. We provide these parameters as a strong starting point for further fine-tuning and adaptation to specific use cases.

\begin{table*}[h]
\small
\centering
\caption{\label{tab:hyper}An Overall Summary of Hyperparameters in Experiments}
\begin{tabular}{lp{6cm}l}
\toprule
Component & Parameter & Value \\
\midrule
GRS & Epsilon rate ($\epsilon$) & 0.05, (0.1, 0.2 in Synthetic Dataset) \\
    & Loss function & Log loss \\
\midrule
DMAN & Model structure & [n\_features, 100, 100, n\_features] \\
     & Optimizer & Adam ($1\times10^{-4}$) \\
\midrule
DiffSortNet & \multicolumn{2}{l}{Type = Bitonic, Steepness = 10, Interpolation = Cauchy, ART lambda ($\lambda$) = 0.25} \\
\midrule
MHMN & Number of heads ($h$) & 50 \\
     & Optimizer & Adam (0.01)\\
     & LR scheduler step size & 50 \\
     & LR scheduler gamma & 0.5 \\
     & Diversity weight \& Sparsity weight & 0.1 \\
\midrule
Decision Tree & \multicolumn{2}{l}{random state = 0, Max depth = 5, Min samples leaf = 5, Min samples split = 5} \\
\bottomrule
\end{tabular}
\end{table*}

% Pre-trained models : LR and ANN from OpenXAI 
% A table for hyperparameters: GRS: 'epsilon_rate': 0.05: loss_fn = log_loss  $f_{\text{dman}}$:
% 'hidden_layers_dman': [100,100]; input_size and output_size: length of features; optimizer : adam (0.0001 or $1\times10^{-4}$); f_{\text{diffsort}}:'bitonic',
%                          x.shape[1],
%                          steepness=10,
%                          interpolation_type='cauchy',
%                          art_lambda=0.25,
%                          device=x.device
% f_{MHMN}: 'num_heads': 50 optimizer : adam; scheduler : step 
% --lr_mul 0.05 --lr_scheduler_step_size 50 --lr_scheduler_gamma 0.5, diversity_weight=0.1, sparse_weight=0.1,

% additional interpretable model: DecisionTreeClassifier(random_state=0, max_depth=5, min_samples_leaf=5, min_samples_split=5)

\begin{table*}[]
\small
\caption{Summary of Datasets Information from OpenXAI}
\label{tab:summary_tables}
\centering
\begin{tabular}{lllp{2cm}p{3cm}l}
\toprule
Dataset        & Size  & \# Features & Feature Types        & Feature Information             & Balanced \\ \midrule
Synthetic Data & 5,000 & 20          & continuous           & synthetic                       & True     \\
German Credit       & 1,000   & 20 & discrete, continuous & demographic, personal, financial                       & False \\
HELOC          & 9,871 & 23          & continuous           & demographic, financial          & True     \\
Adult Income        & 45,222  & 13 & discrete, continuous & demographic, personal, education/employment, financial & False \\
COMPAS         & 6,172 & 7           & discrete, continuous & demographic, personal, criminal & False    \\
Give Me Some Credit & 102,209 & 10 & discrete, continuous & demographic, personal, financial                       & False \\ \bottomrule
\end{tabular}
\end{table*}

\begin{table*}[]
\centering
\caption{Comprehensive Notation Summary for EXAGREE Framework}
% \small
\label{tab:notation}
\begin{tabular}{ll}
\hline
Notation & Description \\
\hline
$x, y, z$ & Scalars \\
$\bfs{v}, \bfs{w}$ & Vectors \\
$\bfs{A}, \bfs{B}$ & Metrics \\
$\R, \Q, \D$ & Sets \\
$v_i$ & $i$-th element of vector $\bfs{v}$ \\
$a_{ij}$ & Element in $i$-th row and $j$-th column of matrix $\bfs{A}$ \\
$\bfs{a}_{i\cdot}$ & $i$-th row of matrix $\bfs{A}$ \\
$\bfs{a}_{\cdot j}$ & $j$-th column of matrix $\bfs{A}$ \\ \midrule
$(\bfs{X}, \bfs{y})$ & Dataset in $\mathbb{R}^{n \times (p+1)}$ \\
$\bfs{X}$ & Covariate input matrix in $\mathbb{R}^{n \times p}$ \\
$\bfs{y}$ & Output vector in $\mathbb{R}^n$ \\
$n$ & Number of instances \\
$p$ & Number of features \\
$h$ & Number of heads \\
$\mathcal{M}$ & Set of all considered models \\
$\Phi$ & Set of explanation methods \\
$\mathcal{S}$ & Set of stakeholders $\{s_1, s_2, s_3, \cdots\}$ \\
$\F$ & Set of features $\{\bfs{x}_{\cdot 1}, \bfs{x}_{\cdot 2}, \ldots, \bfs{x}_{\cdot p}\}$ \\
$\mathcal{M}_\I$ & Subset of interpretable models \\
$f$ & Predictive model: $\mathbb{R}^{n \times p} \to \mathbb{R}^n$ \\
$\Loss$ & Loss function: $\mathbb{R}^n \times \mathbb{R}^n \to \mathbb{R}$ \\
$\bfs{X}_{\setminus i}$ & Input matrix with $i$-th feature replaced \\
$a_i$ & Attribution measure for feature $i$ \\
$\bfs{a}$ & Vector of attributions for all features $\{a_0, a_1, \cdots, a_p\}$ \\
% $\mathcal{A}$ & Set of attribution vectors\\
$\bfs{r}$ & Ranking of features based on attributions $\{r_0, r_1, \cdots, r_p\}$ \\
$\bfs{a}^{M}_{\varphi}$ & Attribution vector for model $M$ and explanation method $\varphi$ \\
$\bfs{r}^{M}_{\varphi}$ & Feature ranking for model $M$ and explanation method $\varphi$ \\
$\bfs{r}^{k}_{\varphi}$ & Arbitrary stakeholder-grounded feature ranking for stakeholder $k$ \\
$\bfs{a}^{M_\I}_{\text{true}}$ & Ground truth attribution vector for interpretable model $M_\I$ \\
$\bfs{r}^{M_\I}_{\text{true}}$ & Ground truth ranking for interpretable model $M_\I$ \\
\hline
\end{tabular}
\end{table*}

\section{Additional Agreement Metrics}\label{sec:additional-results}
In this section, we present a comprehensive evaluation of all explanation methods, including gradient-based techniques applicable to both ANN and LR models. Additionally, we incorporate local model-agnostic methods, such as LIME, to ensure a broad assessment across various explanation paradigms. To maintain consistency, we report the \textit{mean} values of all agreement metrics, following the OpenXAI \textit{workflow}. Performance curves for varying $k$ values and a fixed $k=25\%$ baseline are reported in this section \ref{sec:summary_of_information}. Gradient-based explanations consistently demonstrate excellent agreement on LR models across all datasets, aligning with expectations given that the ground truth explanations are derived from LR coefficients. Interestingly, when applied to ANNs, some of these methods also show promising agreement, suggesting that in certain cases, these black-box models may coincidentally make predictions in a manner similar to the ground truth LR model. 

\begin{table*}[hptb]
\small
\centering
\caption{\label{tab:results_adult} Comprehensive evaluation of ranking agreement ($k$=0.25) on the Adult Income dataset across various explanation methods applied to LR and ANN models. Delivered explanations correspond to post-hoc methods applied to LR and ANN ($\bfs{r}^{\text{LR}}_{\text{post}}$, $\bfs{r}^{\text{ANN}}_{\text{post}}$), while the stakeholder-grounded explanation remains fixed as $\bfs{r}^{k}_{\text{LR}}$. Higher values ($\uparrow$) indicate better agreement, while lower values ($\downarrow$) indicate weaker alignment with stakeholder expectations.}
\begin{tabular}{p{.4cm}lllllllll}
\toprule
& \textbf{Method} & \textbf{FA($\uparrow$)} & \textbf{RA($\uparrow$)} & \textbf{SA($\uparrow$)} & \textbf{SRA($\uparrow$)} & \textbf{PRA($\uparrow$)} & \textbf{PGI($\uparrow$)} & \textbf{PGU($\downarrow$)} \\ \midrule
\multirow{9}{*}{LR} & LIME & 1.00 & 1.00 & 0.00 & 0.00 & 0.99 & 0.15 & 0.04 \\ 
& SHAP & 0.50 & 0.25 & 0.00 & 0.00 & 0.69 & 0.08 & 0.13 \\ 
& Integrated Gradient & 1.00 & 1.00 & 0.00 & 0.00 & 1.00 & 0.15 & 0.04 \\ 
& Vanilla Gradient & 1.00 & 1.00 & 0.00 & 0.00 & 1.00 & 0.15 & 0.04 \\ 
& SmoothGrad & 1.00 & 1.00 & 0.00 & 0.00 & 1.00 & 0.15 & 0.04 \\ 
& Random & 0.75 & 0.00 & 0.50 & 0.00 & 0.55 & 0.13 & 0.06 \\ 
& Gradient x Input & 0.50 & 0.00 & 0.00 & 0.00 & 0.72 & 0.07 & 0.13 \\ 
& FIS\_LR & 0.75 & 0.00 & 0.50 & 0.00 & 0.82 & 0.14 & 0.05 \\
& FIS\_SAEM & 1.00 & 0.25 & 0.75 & 0.25 & 0.90 & 0.15 & 0.04 \\ \midrule
\multirow{9}{*}{ANN} & LIME & 0.50 & 0.25 & 0.00 & 0.00 & 0.74 & 0.23 & 0.06 \\ 
& SHAP & 0.75 & 0.00 & 0.50 & 0.00 & 0.81 & 0.24 & 0.06 \\ 
& Integrated Gradient & 0.75 & 0.75 & 0.00 & 0.00 & 0.72 & 0.24 & 0.06 \\ 
& Vanilla Gradient & 0.50 & 0.50 & 0.00 & 0.00 & 0.63 & 0.23 & 0.07 \\ 
& SmoothGrad & 0.50 & 0.25 & 0.00 & 0.00 & 0.74 & 0.23 & 0.06 \\ 
& Random & 0.75 & 0.00 & 0.50 & 0.00 & 0.55 & 0.24 & 0.08 \\ 
& Gradient x Input & 0.25 & 0.00 & 0.00 & 0.00 & 0.53 & 0.06 & 0.24 \\ 
& FIS\_ANN & 0.75 & 0.00 & 0.50 & 0.00 & 0.83 & 0.24 & 0.07 \\
& FIS\_SAEM & 0.75 & 0.25 & 0.50 & 0.25 & 0.84 & 0.24 & 0.06 \\ \midrule
DT & Intrinsic Explanation & 1.00 & 0.25 & 0.75 & 0.25 & 0.80 & 0.15 & 0.04 \\ \bottomrule
\end{tabular}
\end{table*}
\begin{table*}[hptb]
\small
\centering
\caption{\label{tab:results_synthetic} Comprehensive evaluation of ranking agreement ($k$=0.25) on the Synthetic dataset across various explanation methods applied to LR and ANN models. Delivered explanations correspond to post-hoc methods applied to LR and ANN ($\bfs{r}^{\text{LR}}_{\text{post}}$, $\bfs{r}^{\text{ANN}}_{\text{post}}$), while the stakeholder-grounded explanation remains fixed as $\bfs{r}^{k}_{\text{LR}}$. Higher values ($\uparrow$) indicate better agreement, while lower values ($\downarrow$) indicate weaker alignment with stakeholder expectations.}
\begin{tabular}{p{.4cm}lllllllll}
\toprule
& \textbf{Method} & \textbf{FA($\uparrow$)} & \textbf{RA($\uparrow$)} & \textbf{SA($\uparrow$)} & \textbf{SRA($\uparrow$)} & \textbf{PRA($\uparrow$)} & \textbf{PGI($\uparrow$)} & \textbf{PGU($\downarrow$)} \\ \midrule
\multirow{9}{*}{LR} 
& LIME & 1.00 & 1.00 & 1.00 & 1.00 & 0.98 & 0.13 & 0.07 \\ 
& SHAP & 1.00 & 0.20 & 0.40 & 0.20 & 0.93 & 0.13 & 0.07 \\ 
& Integrated Gradient & 1.00 & 1.00 & 1.00 & 1.00 & 1.00 & 0.13 & 0.07 \\ 
& Vanilla Gradient & 1.00 & 1.00 & 1.00 & 1.00 & 1.00 & 0.13 & 0.07 \\ 
& SmoothGrad & 1.00 & 1.00 & 1.00 & 1.00 & 1.00 & 0.13 & 0.07 \\ 
& Random & 0.20 & 0.00 & 0.00 & 0.00 & 0.44 & 0.07 & 0.13 \\ 
& Gradient x Input & 0.40 & 0.00 & 0.40 & 0.00 & 0.79 & 0.09 & 0.11 \\ 
& FIS\_LR & 1.00 & 0.60 & 0.40 & 0.20 & 0.94 & 0.13 & 0.07 \\ 
& FIS\_SAEM & 1.00 & 1.00 & 0.40 & 0.40 & 0.94 & 0.13 & 0.07 \\ \midrule

\multirow{9}{*}{ANN} 
& LIME & 0.40 & 0.00 & 0.20 & 0.00 & 0.72 & 0.11 & 0.14 \\ 
& SHAP & 0.60 & 0.20 & 0.00 & 0.00 & 0.74 & 0.11 & 0.13 \\ 
& Integrated Gradient & 0.80 & 0.20 & 0.00 & 0.00 & 0.74 & 0.11 & 0.13 \\ 
& Vanilla Gradient & 0.40 & 0.20 & 0.40 & 0.20 & 0.63 & 0.12 & 0.13 \\ 
& SmoothGrad & 0.20 & 0.00 & 0.20 & 0.00 & 0.69 & 0.10 & 0.14 \\ 
& Random & 0.20 & 0.00 & 0.00 & 0.00 & 0.44 & 0.06 & 0.16 \\ 
& Gradient x Input & 0.40 & 0.20 & 0.20 & 0.00 & 0.66 & 0.11 & 0.13 \\ 
& FIS\_ANN & 0.80 & 0.20 & 0.20 & 0.00 & 0.74 & 0.12 & 0.12 \\ 
& FIS\_SAEM & 0.80 & 0.20 & 0.20 & 0.00 & 0.78 & 0.12 & 0.12 \\ \midrule

DT & Intrinsic Explanation & 0.60 & 0.00 & 0.00 & 0.00 & 0.48 & 0.11 & 0.10 \\ 
\bottomrule
\end{tabular}
\end{table*}
\begin{table*}[hptb]
\small
\centering
\caption{\label{tab:results_compas} Comprehensive evaluation of ranking agreement ($k$=0.25) on the COMPAS dataset across various explanation methods applied to LR and ANN models. Delivered explanations correspond to post-hoc methods applied to LR and ANN ($\bfs{r}^{\text{LR}}_{\text{post}}$, $\bfs{r}^{\text{ANN}}_{\text{post}}$), while the stakeholder-grounded explanation remains fixed as $\bfs{r}^{k}_{\text{LR}}$. Higher values ($\uparrow$) indicate better agreement, while lower values ($\downarrow$) indicate weaker alignment with stakeholder expectations.}
\begin{tabular}{p{.4cm}lllllllll}
\toprule
& \textbf{Method} & \textbf{FA($\uparrow$)} & \textbf{RA($\uparrow$)} & \textbf{SA($\uparrow$)} & \textbf{SRA($\uparrow$)} & \textbf{PRA($\uparrow$)} & \textbf{PGI($\uparrow$)} & \textbf{PGU($\downarrow$)} \\ \midrule
\multirow{9}{*}{LR} 
& LIME & 1.00 & 1.00 & 1.00 & 1.00 & 1.00 & 0.06 & 0.05 \\ 
& SHAP & 0.50 & 0.00 & 0.50 & 0.00 & 0.57 & 0.05 & 0.06 \\ 
& Integrated Gradient & 1.00 & 1.00 & 1.00 & 1.00 & 1.00 & 0.06 & 0.05 \\ 
& Vanilla Gradient & 1.00 & 1.00 & 1.00 & 1.00 & 1.00 & 0.06 & 0.05 \\ 
& SmoothGrad & 1.00 & 1.00 & 1.00 & 1.00 & 1.00 & 0.06 & 0.05 \\ 
& Random & 0.50 & 0.50 & 0.50 & 0.50 & 0.76 & 0.05 & 0.06 \\ 
& Gradient x Input & 0.50 & 0.00 & 0.50 & 0.00 & 0.52 & 0.05 & 0.06 \\ 
& FIS\_LR & 1.00 & 1.00 & 0.50 & 0.50 & 0.86 & 0.06 & 0.05 \\ 
& FIS\_SAEM & 1.00 & 1.00 & 0.50 & 0.50 & 0.90 & 0.06 & 0.05 \\ \midrule

\multirow{9}{*}{ANN} 
& LIME & 1.00 & 0.00 & 1.00 & 0.00 & 0.86 & 0.10 & 0.03 \\ 
& SHAP & 0.50 & 0.00 & 0.50 & 0.00 & 0.52 & 0.07 & 0.07 \\ 
& Integrated Gradient & 1.00 & 0.00 & 0.00 & 0.00 & 0.71 & 0.10 & 0.03 \\ 
& Vanilla Gradient & 1.00 & 0.00 & 1.00 & 0.00 & 0.81 & 0.10 & 0.03 \\ 
& SmoothGrad & 1.00 & 0.00 & 1.00 & 0.00 & 0.86 & 0.10 & 0.03 \\ 
& Random & 0.50 & 0.50 & 0.50 & 0.50 & 0.76 & 0.08 & 0.07 \\ 
& Gradient x Input & 0.00 & 0.00 & 0.00 & 0.00 & 0.33 & 0.01 & 0.10 \\ 
& FIS\_ANN & 1.00 & 1.00 & 0.50 & 0.50 & 0.86 & 0.10 & 0.03 \\ 
& FIS\_SAEM & 1.00 & 1.00 & 0.50 & 0.50 & 0.96 & 0.10 & 0.03 \\ \midrule

DT & Intrinsic Explanation & 1.00 & 1.00 & 0.50 & 0.50 & 0.95 & 0.06 & 0.05 \\ 
\bottomrule
\end{tabular}
\end{table*}
\begin{table*}[hptb]
\small
\centering
\caption{\label{tab:results_german} Comprehensive evaluation of ranking agreement ($k$=0.25) on the German Credit dataset across various explanation methods applied to LR and ANN models. Delivered explanations correspond to post-hoc methods applied to LR and ANN ($\bfs{r}^{\text{LR}}_{\text{post}}$, $\bfs{r}^{\text{ANN}}_{\text{post}}$), while the stakeholder-grounded explanation remains fixed as $\bfs{r}^{k}_{\text{LR}}$. Higher values ($\uparrow$) indicate better agreement, while lower values ($\downarrow$) indicate weaker alignment with stakeholder expectations.}
\begin{tabular}{p{.4cm}lllllllll}
\toprule
& \textbf{Method} & \textbf{FA($\uparrow$)} & \textbf{RA($\uparrow$)} & \textbf{SA($\uparrow$)} & \textbf{SRA($\uparrow$)} & \textbf{PRA($\uparrow$)} & \textbf{PGI($\uparrow$)} & \textbf{PGU($\downarrow$)} \\ \midrule
\multirow{9}{*}{LR} 
& LIME & 1.00 & 0.87 & 1.00 & 0.87 & 1.00 & 0.04 & 0.03 \\ 
& SHAP & 0.53 & 0.00 & 0.53 & 0.00 & 0.68 & 0.04 & 0.03 \\ 
& Integrated Gradient & 1.00 & 1.00 & 1.00 & 1.00 & 1.00 & 0.04 & 0.03 \\ 
& Vanilla Gradient & 1.00 & 1.00 & 1.00 & 1.00 & 1.00 & 0.04 & 0.03 \\ 
& SmoothGrad & 1.00 & 1.00 & 1.00 & 1.00 & 1.00 & 0.04 & 0.03 \\ 
& Random & 0.27 & 0.00 & 0.13 & 0.00 & 0.45 & 0.02 & 0.05 \\ 
& Gradient x Input & 0.53 & 0.00 & 0.53 & 0.00 & 0.67 & 0.04 & 0.03 \\ 
& FIS\_LR & 0.53 & 0.07 & 0.33 & 0.07 & 0.76 & 0.04 & 0.03 \\ 
& FIS\_SAEM & 0.53 & 0.07 & 0.27 & 0.07 & 0.78 & 0.04 & 0.03 \\ \midrule

\multirow{9}{*}{ANN} 
& LIME & 0.40 & 0.00 & 0.40 & 0.00 & 0.62 & 0.05 & 0.10 \\ 
& SHAP & 0.33 & 0.00 & 0.33 & 0.00 & 0.49 & 0.07 & 0.10 \\ 
& Integrated Gradient & 0.40 & 0.00 & 0.40 & 0.00 & 0.59 & 0.06 & 0.10 \\ 
& Vanilla Gradient & 0.33 & 0.00 & 0.33 & 0.00 & 0.63 & 0.06 & 0.10 \\ 
& SmoothGrad & 0.33 & 0.00 & 0.33 & 0.00 & 0.62 & 0.06 & 0.10 \\ 
& Random & 0.27 & 0.00 & 0.13 & 0.00 & 0.45 & 0.05 & 0.11 \\ 
& Gradient x Input & 0.47 & 0.00 & 0.47 & 0.00 & 0.52 & 0.07 & 0.10 \\ 
& FIS\_ANN & 0.40 & 0.07 & 0.26 & 0.07 & 0.52 & 0.06 & 0.10 \\ 
& FIS\_SAEM & 0.33 & 0.00 & 0.13 & 0.00 & 0.59 & 0.06 & 0.10 \\ \midrule

DT & Intrinsic Explanation & 0.27 & 0.00 & 0.13 & 0.00 & 0.48 & 0.04 & 0.03 \\ 
\bottomrule
\end{tabular}
\end{table*}
\begin{table*}[hptb]
\small
\centering
\caption{\label{tab:results_heloc} Comprehensive evaluation of ranking agreement ($k$=0.25) on the HELOC dataset across various explanation methods applied to LR and ANN models. Delivered explanations correspond to post-hoc methods applied to LR and ANN ($\bfs{r}^{\text{LR}}_{\text{post}}$, $\bfs{r}^{\text{ANN}}_{\text{post}}$), while the stakeholder-grounded explanation remains fixed as $\bfs{r}^{k}_{\text{LR}}$. Higher values ($\uparrow$) indicate better agreement, while lower values ($\downarrow$) indicate weaker alignment with stakeholder expectations.}
\begin{tabular}{p{.4cm}lllllllll}
\toprule
& \textbf{Method} & \textbf{FA($\uparrow$)} & \textbf{RA($\uparrow$)} & \textbf{SA($\uparrow$)} & \textbf{SRA($\uparrow$)} & \textbf{PRA($\uparrow$)} & \textbf{PGI($\uparrow$)} & \textbf{PGU($\downarrow$)} \\ \midrule
\multirow{9}{*}{LR} 
& LIME & 1.00 & 1.00 & 1.00 & 1.00 & 0.97 & 0.09 & 0.05 \\ 
& SHAP & 0.33 & 0.17 & 0.17 & 0.00 & 0.68 & 0.06 & 0.09 \\ 
& Integrated Gradient & 1.00 & 1.00 & 1.00 & 1.00 & 1.00 & 0.09 & 0.05 \\ 
& Vanilla Gradient & 1.00 & 1.00 & 1.00 & 1.00 & 1.00 & 0.09 & 0.05 \\ 
& SmoothGrad & 1.00 & 1.00 & 1.00 & 1.00 & 1.00 & 0.09 & 0.05 \\ 
& Random & 0.17 & 0.00 & 0.17 & 0.00 & 0.40 & 0.05 & 0.09 \\ 
& Gradient x Input & 0.17 & 0.00 & 0.17 & 0.00 & 0.61 & 0.05 & 0.09 \\ 
& FIS\_LR & 0.33 & 0.17 & 0.33 & 0.17 & 0.51 & 0.07 & 0.08 \\ 
& FIS\_SAEM & 0.67 & 0.17 & 0.17 & 0.17 & 0.78 & 0.07 & 0.08 \\ \midrule

\multirow{9}{*}{ANN} 
& LIME & 0.67 & 0.17 & 0.67 & 0.17 & 0.76 & 0.10 & 0.07 \\ 
& SHAP & 0.17 & 0.00 & 0.17 & 0.00 & 0.47 & 0.06 & 0.10 \\ 
& Integrated Gradient & 0.83 & 0.00 & 0.00 & 0.00 & 0.84 & 0.09 & 0.08 \\ 
& Vanilla Gradient & 0.67 & 0.33 & 0.67 & 0.33 & 0.78 & 0.10 & 0.07 \\ 
& SmoothGrad & 0.67 & 0.17 & 0.67 & 0.17 & 0.77 & 0.10 & 0.07 \\ 
& Random & 0.17 & 0.00 & 0.17 & 0.00 & 0.40 & 0.05 & 0.11 \\ 
& Gradient x Input & 0.33 & 0.00 & 0.33 & 0.00 & 0.62 & 0.06 & 0.10 \\ 
& FIS\_ANN & 0.33 & 0.00 & 0.33 & 0.00 & 0.23 & 0.08 & 0.08 \\ 
& FIS\_SAEM & 0.33 & 0.00 & 0.17 & 0.00 & 0.31 & 0.08 & 0.09 \\ \midrule

DT & Intrinsic Explanation & 0.33 & 0.00 & 0.17 & 0.00 & 0.22 & 0.06 & 0.09 \\ 
\bottomrule
\end{tabular}
\end{table*}
\begin{table*}[hptb]
\small
\centering
\caption{\label{tab:results_gmsc} Comprehensive evaluation of ranking agreement ($k$=0.25) on the Give Me Some Credit (GMSC) dataset across various explanation methods applied to LR and ANN models. Delivered explanations correspond to post-hoc methods applied to LR and ANN ($\bfs{r}^{\text{LR}}_{\text{post}}$, $\bfs{r}^{\text{ANN}}_{\text{post}}$), while the stakeholder-grounded explanation remains fixed as $\bfs{r}^{k}_{\text{LR}}$. Higher values ($\uparrow$) indicate better agreement, while lower values ($\downarrow$) indicate weaker alignment with stakeholder expectations.}
\begin{tabular}{p{.4cm}lllllllll}
\toprule
& \textbf{Method} & \textbf{FA($\uparrow$)} & \textbf{RA($\uparrow$)} & \textbf{SA($\uparrow$)} & \textbf{SRA($\uparrow$)} & \textbf{PRA($\uparrow$)} & \textbf{PGI($\uparrow$)} & \textbf{PGU($\downarrow$)} \\ \midrule
\multirow{9}{*}{LR} 
& LIME & 1.00 & 1.00 & 1.00 & 1.00 & 1.00 & 0.03 & 0.01 \\ 
& SHAP & 0.00 & 0.00 & 0.00 & 0.00 & 0.67 & 0.01 & 0.03 \\ 
& Integrated Gradient & 1.00 & 1.00 & 1.00 & 1.00 & 1.00 & 0.03 & 0.01 \\ 
& Vanilla Gradient & 1.00 & 1.00 & 1.00 & 1.00 & 1.00 & 0.03 & 0.01 \\ 
& SmoothGrad & 1.00 & 1.00 & 1.00 & 1.00 & 1.00 & 0.03 & 0.01 \\ 
& Random & 0.33 & 0.33 & 0.00 & 0.00 & 0.42 & 0.01 & 0.03 \\ 
& Gradient x Input & 0.67 & 0.00 & 0.67 & 0.00 & 0.73 & 0.02 & 0.01 \\ 
& FIS\_LR & 0.33 & 0.00 & 0.00 & 0.00 & 0.42 & 0.03 & 0.01 \\ 
& FIS\_SAEM & 0.66 & 0.66 & 0.00 & 0.00 & 0.56 & 0.03 & 0.01 \\ \midrule

\multirow{9}{*}{ANN} 
& LIME & 1.00 & 1.00 & 1.00 & 1.00 & 0.91 & 0.11 & 0.01 \\ 
& SHAP & 0.33 & 0.33 & 0.33 & 0.33 & 0.67 & 0.10 & 0.02 \\ 
& Integrated Gradient & 0.67 & 0.67 & 0.67 & 0.67 & 0.96 & 0.11 & 0.01 \\ 
& Vanilla Gradient & 1.00 & 1.00 & 1.00 & 1.00 & 0.98 & 0.11 & 0.01 \\ 
& SmoothGrad & 1.00 & 1.00 & 1.00 & 1.00 & 0.91 & 0.11 & 0.01 \\ 
& Random & 0.33 & 0.33 & 0.00 & 0.00 & 0.42 & 0.02 & 0.11 \\ 
& Gradient x Input & 0.33 & 0.33 & 0.33 & 0.33 & 0.60 & 0.10 & 0.02 \\ 
& FIS\_ANN & 0.33 & 0.33 & 0.00 & 0.00 & 0.32 & 0.10 & 0.02 \\ 
& FIS\_SAEM & 0.67 & 0.33 & 0.33 & 0.33 & 0.73 & 0.11 & 0.01 \\ \midrule

DT & Intrinsic Explanation & 0.67 & 0.00 & 0.00 & 0.00 & 0.79 & 0.02 & 0.01 \\ 
\bottomrule
\end{tabular}
\end{table*}

\begin{figure*}[t!]
\centering
    \begin{subfigure}{\textwidth}
    \centering
        \begin{subfigure}[t]{0.24\textwidth}
        \centering
        \includegraphics[width=\textwidth]{Figures/gaussian_lr_FA_11.jpg}
        % \caption{FA improvement on Ground Truth}
    \end{subfigure}%
        \hfill
        \begin{subfigure}[t]{0.24\textwidth}
        \centering
        \includegraphics[width=\textwidth]{Figures/gaussian_lr_RA_11.jpg}
        % \caption{RA improvement on Ground Truth}
    \end{subfigure}
        \hfill
        \begin{subfigure}[t]{0.24\textwidth}
        \centering
        \includegraphics[width=\textwidth]{Figures/gaussian_lr_SA_11.jpg}
        % \caption{SA improvement on Ground Truth}
    \end{subfigure}
        \hfill
        \begin{subfigure}[t]{0.24\textwidth}
        \centering
        \includegraphics[width=\textwidth]{Figures/gaussian_lr_SRA_11.jpg}
        % \caption{SRA improvement on Ground Truth}
    \end{subfigure}
        \vfill
        \begin{subfigure}[t]{0.24\textwidth}
        \centering
        \includegraphics[width=\textwidth]{Figures/gaussian_ann_FA_10.jpg}
        % \caption{Before}
        \end{subfigure}%
        \hfill
        \begin{subfigure}[t]{0.24\textwidth}
        \centering
        \includegraphics[width=\textwidth]{Figures/gaussian_ann_RA_10.jpg}
        % \caption{After}
        \end{subfigure}
        \hfill
        \begin{subfigure}[t]{0.24\textwidth}
        \centering
        \includegraphics[width=\textwidth]{Figures/gaussian_ann_SA_10.jpg}
        % \caption{After}
    \end{subfigure}
        \hfill
        \begin{subfigure}[t]{0.24\textwidth}
        \centering
        \includegraphics[width=\textwidth]{Figures/gaussian_ann_SRA_10.jpg}
        \end{subfigure}
    \caption{Synthetic Dataset: LR (top row) and ANN (bottom row) in blue; corresponding SAEMs in red.}
    \end{subfigure}
    \vfill
    \begin{subfigure}{\textwidth}
    \centering
        \begin{subfigure}[t]{0.24\textwidth}
            \centering
            \includegraphics[width=\textwidth]{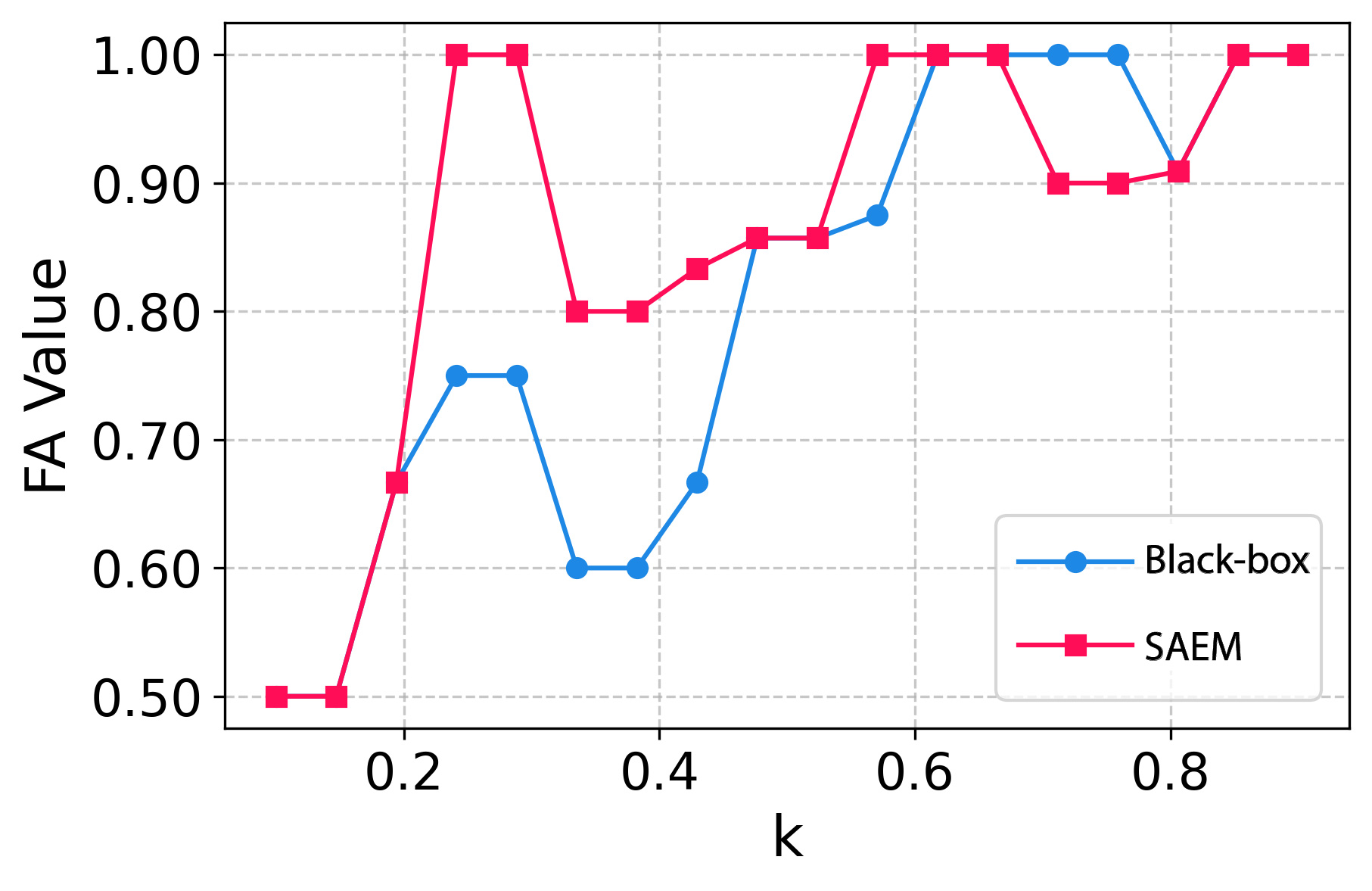}
            % \caption{FA improvement on Ground Truth}
        \end{subfigure}%
        \hfill
        \begin{subfigure}[t]{0.24\textwidth}
            \centering
            \includegraphics[width=\textwidth]{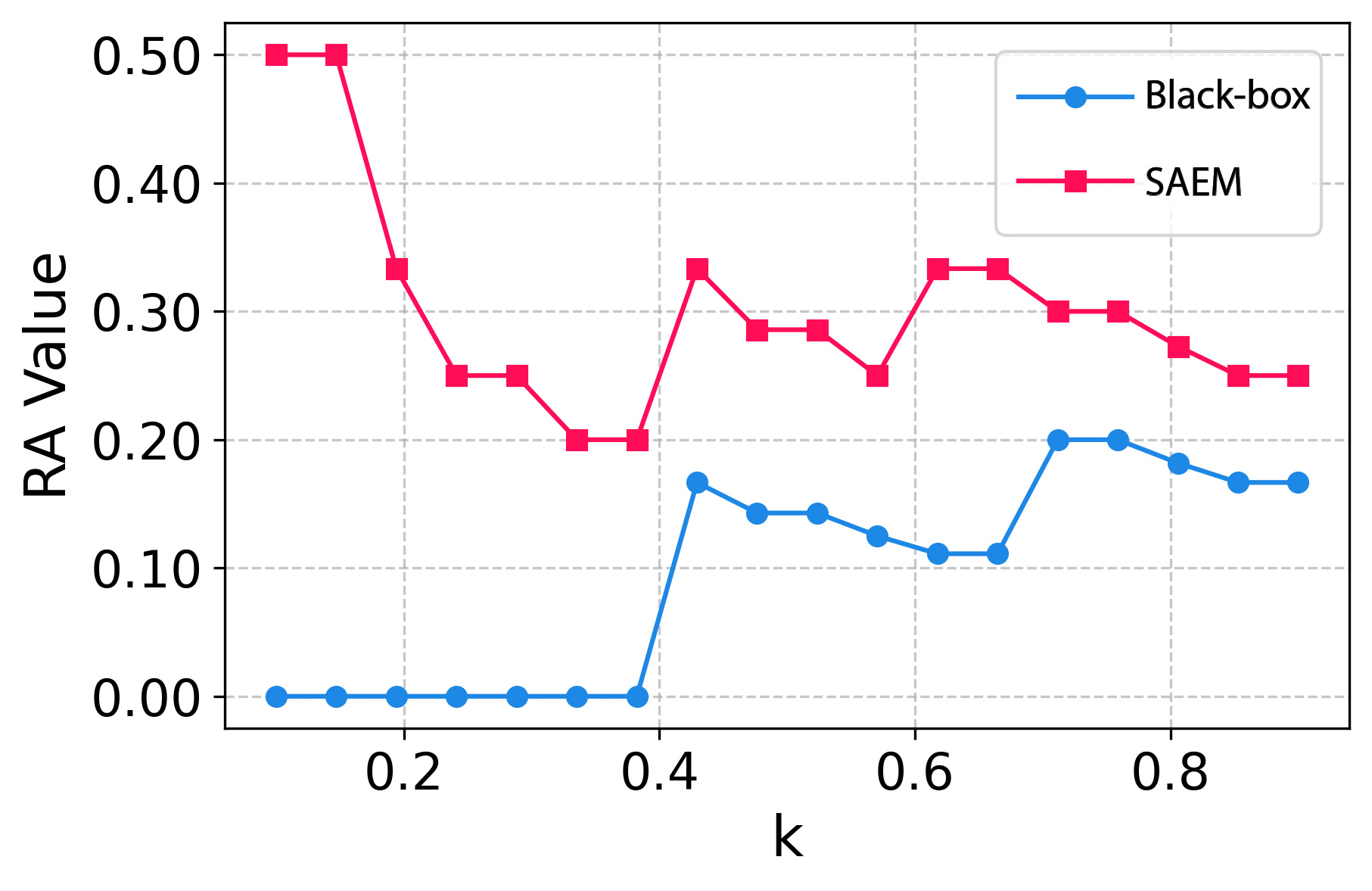}
            % \caption{RA improvement on Ground Truth}
        \end{subfigure}
        \hfill
        \begin{subfigure}[t]{0.24\textwidth}
            \centering
            \includegraphics[width=\textwidth]{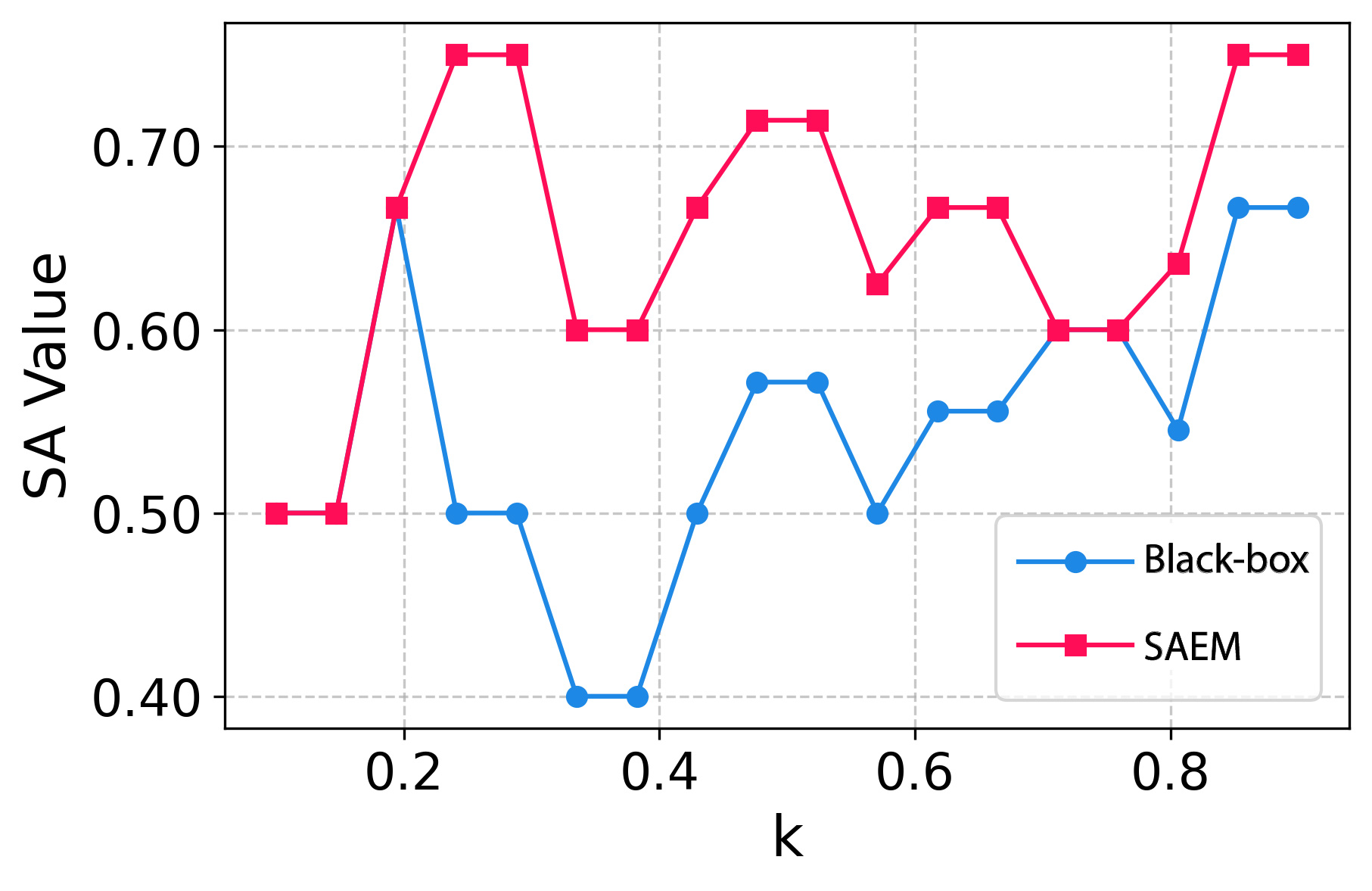}
            % \caption{SA improvement on Ground Truth}
        \end{subfigure}
        \hfill
        \begin{subfigure}[t]{0.24\textwidth}
            \centering
            \includegraphics[width=\textwidth]{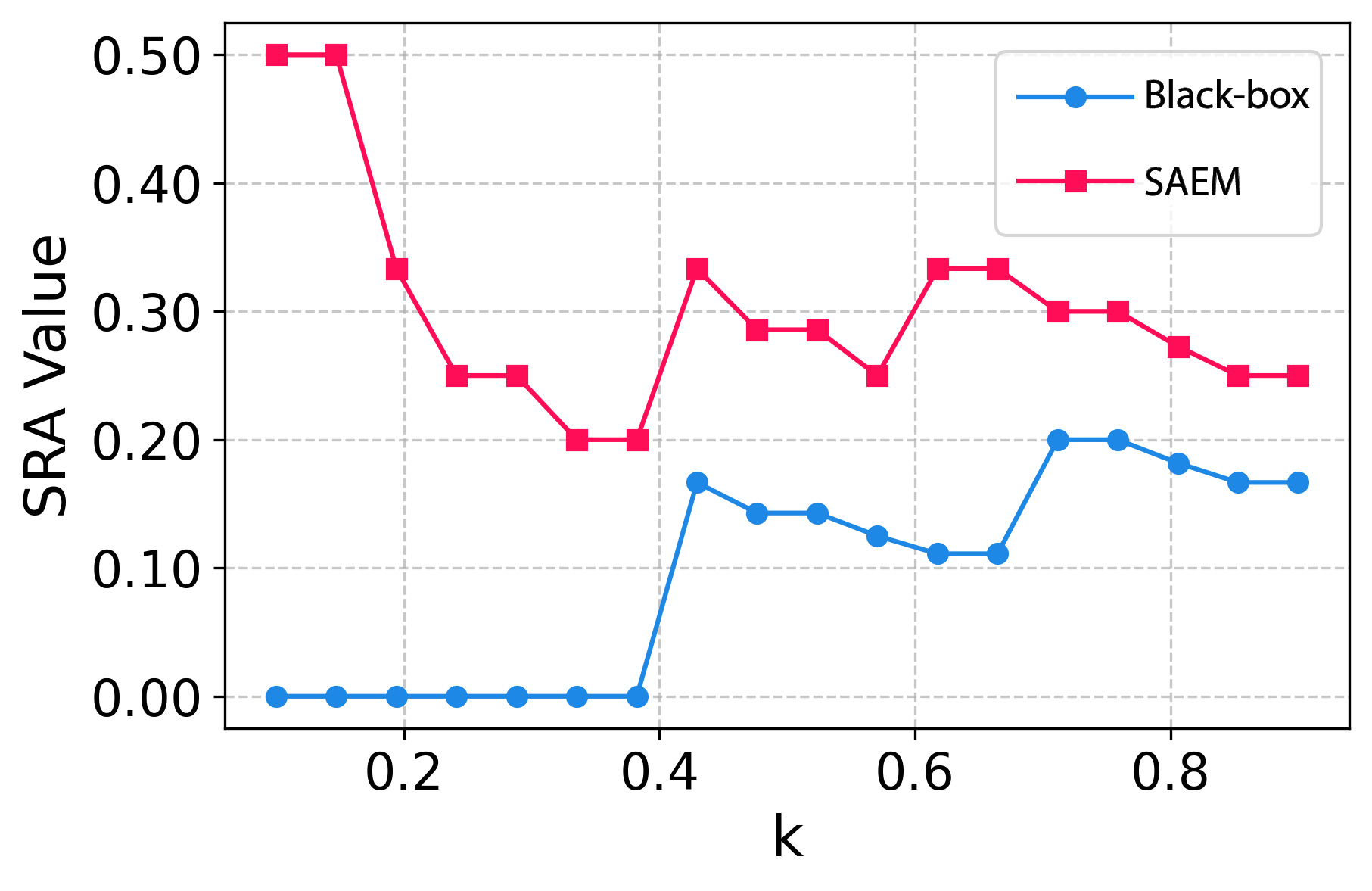}
            % \caption{SRA improvement on Ground Truth}
        \end{subfigure}
        \vfill
        \begin{subfigure}[t]{0.24\textwidth}
            \centering
            \includegraphics[width=\textwidth]{Figures/adult_ann_FA_14.jpg}
            % \caption{Before}
        \end{subfigure}%
        \hfill
        \begin{subfigure}[t]{0.24\textwidth}
            \centering
            \includegraphics[width=\textwidth]{Figures/adult_ann_RA_14.jpg}
            % \caption{After}
        \end{subfigure}
        \hfill
        \begin{subfigure}[t]{0.24\textwidth}
            \centering
            \includegraphics[width=\textwidth]{Figures/adult_ann_SA_14.jpg}
            % \caption{After}
        \end{subfigure}
        \hfill
        \begin{subfigure}[t]{0.24\textwidth}
            \centering
            \includegraphics[width=\textwidth]{Figures/adult_ann_SRA_14.jpg}
            % \caption{After}
        \end{subfigure}
    \caption{Adult Income Dataset: LR (top row) and ANN (bottom row) in blue; corresponding SAEMs in red.}
    \end{subfigure}
    \vfill
        \begin{subfigure}{\textwidth}
        \centering
        \begin{subfigure}[t]{0.24\textwidth}
        \centering
        \includegraphics[width=\textwidth]{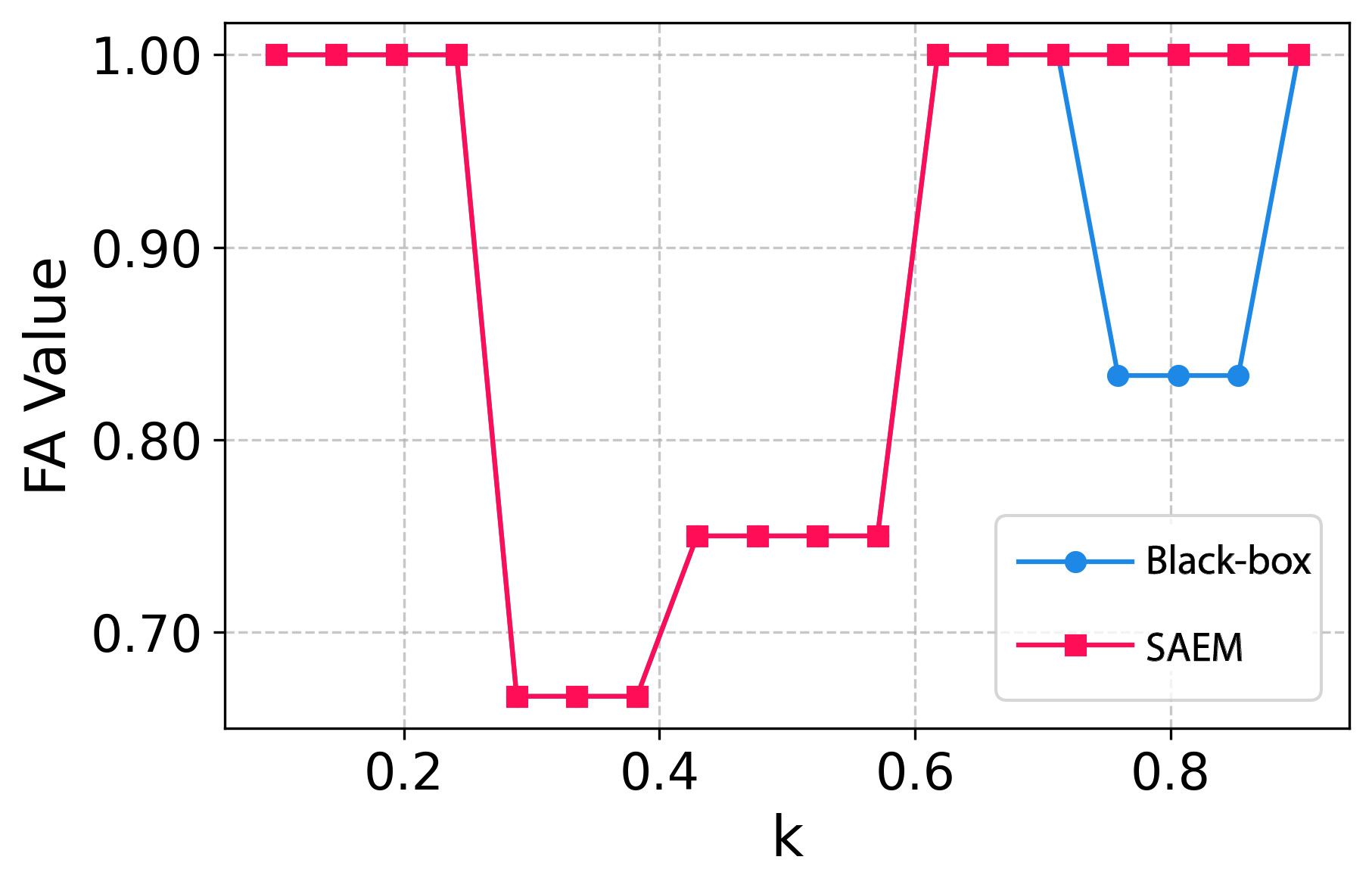}
        % \caption{FA improvement on Ground Truth}
    \end{subfigure}%
        \hfill
        \begin{subfigure}[t]{0.24\textwidth}
        \centering
        \includegraphics[width=\textwidth]{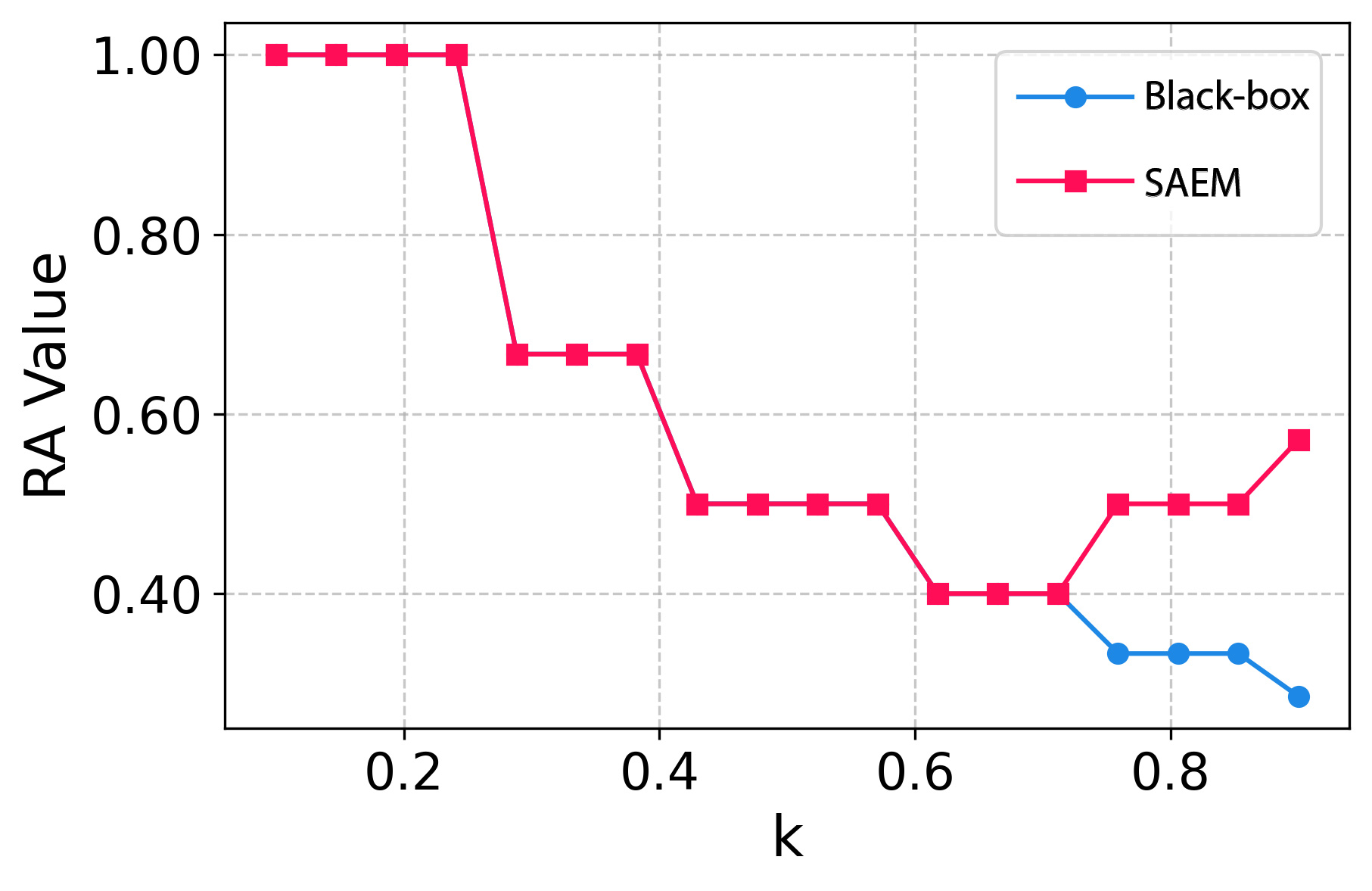}
        % \caption{RA improvement on Ground Truth}
    \end{subfigure}
        \hfill
        \begin{subfigure}[t]{0.24\textwidth}
        \centering
        \includegraphics[width=\textwidth]{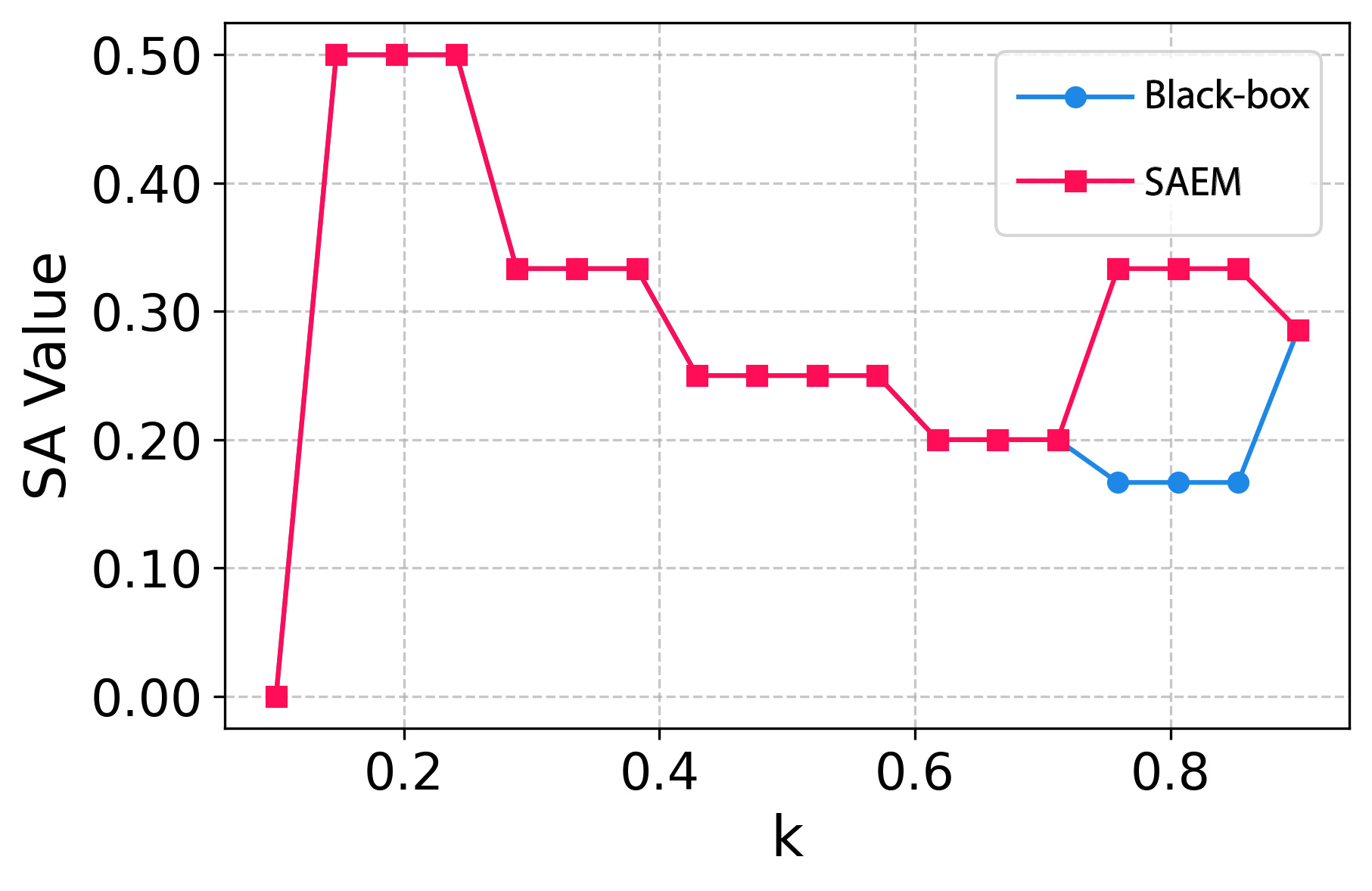}
        % \caption{SA improvement on Ground Truth}
    \end{subfigure}
        \hfill
        \begin{subfigure}[t]{0.24\textwidth}
        \centering
        \includegraphics[width=\textwidth]{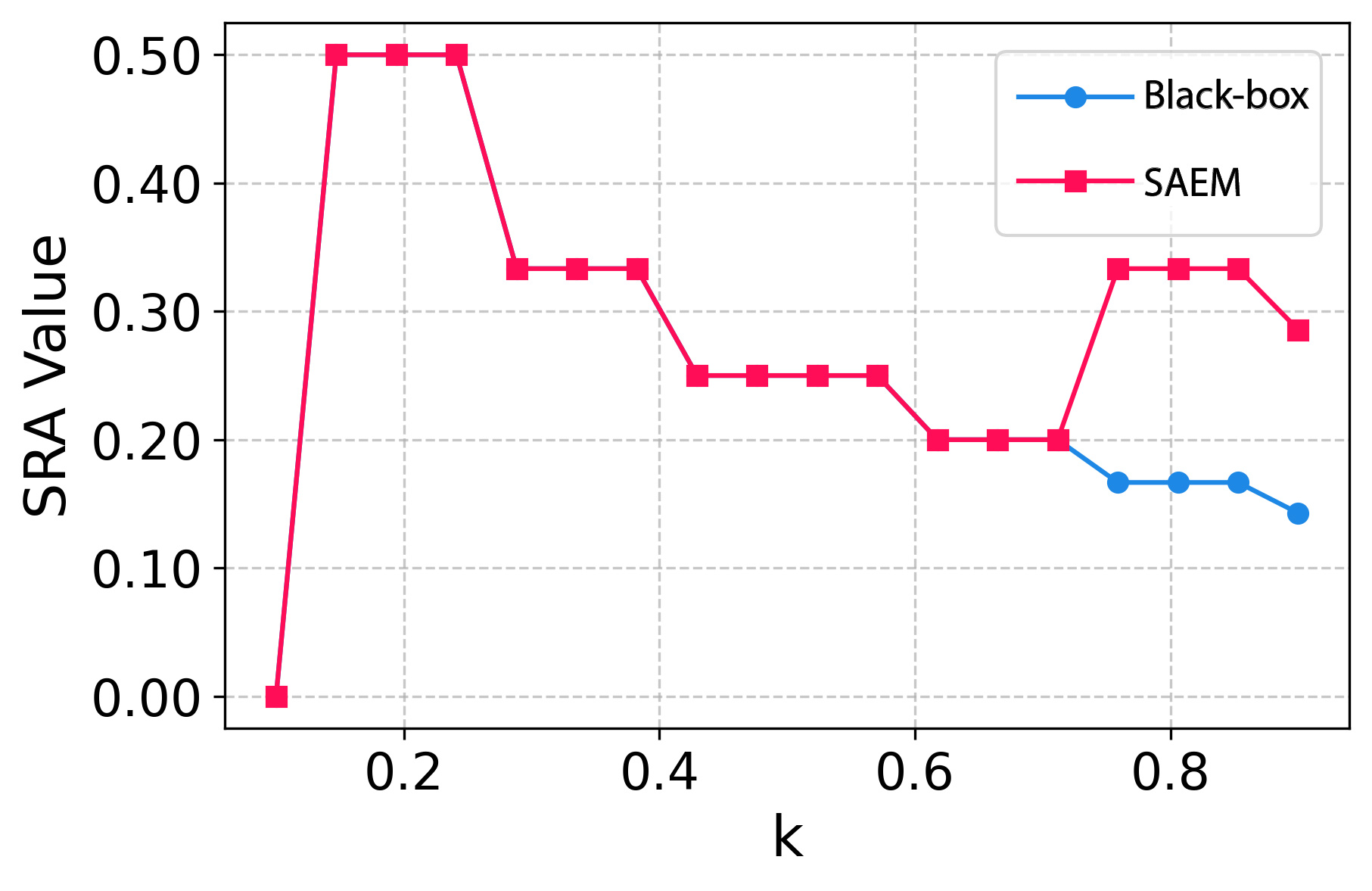}
        % \caption{SRA improvement on Ground Truth}
    \end{subfigure}
        \vfill
        \begin{subfigure}[t]{0.24\textwidth}
        \centering
        \includegraphics[width=\textwidth]{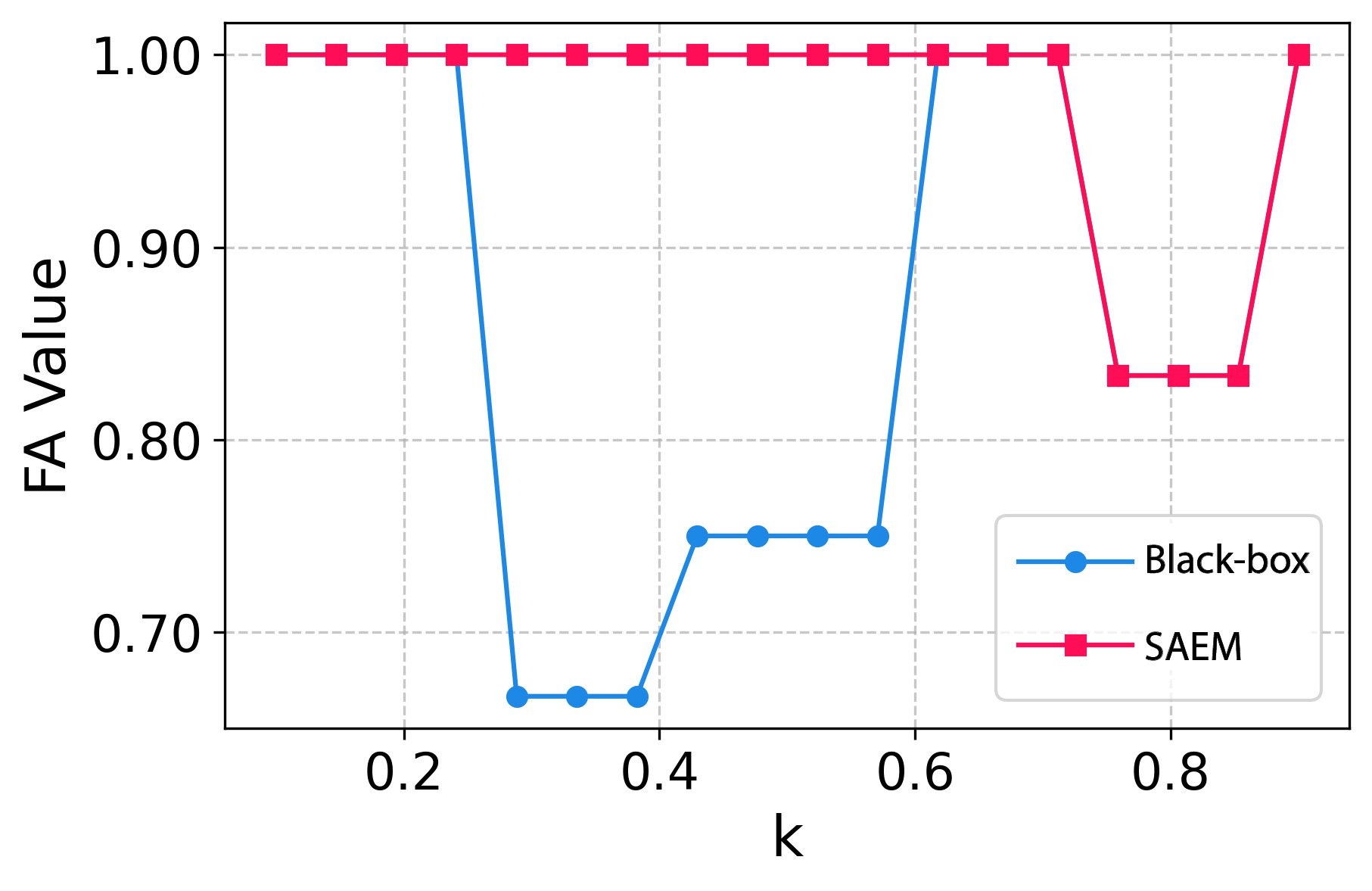}
        % \caption{Before}
    \end{subfigure}%
        \hfill
        \begin{subfigure}[t]{0.24\textwidth}
        \centering
        \includegraphics[width=\textwidth]{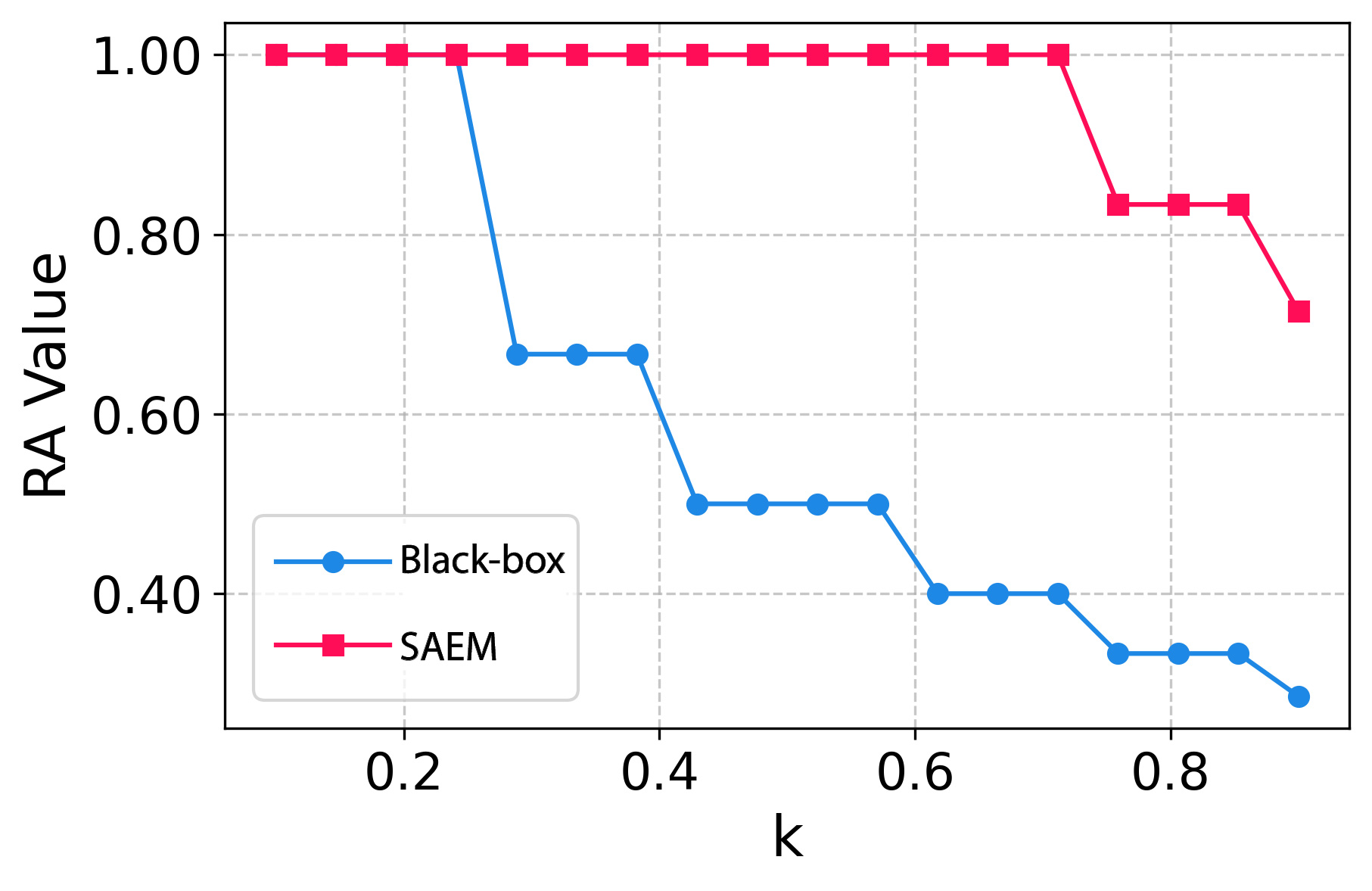}
        % \caption{After}
    \end{subfigure}
        \hfill
        \begin{subfigure}[t]{0.24\textwidth}
        \centering
        \includegraphics[width=\textwidth]{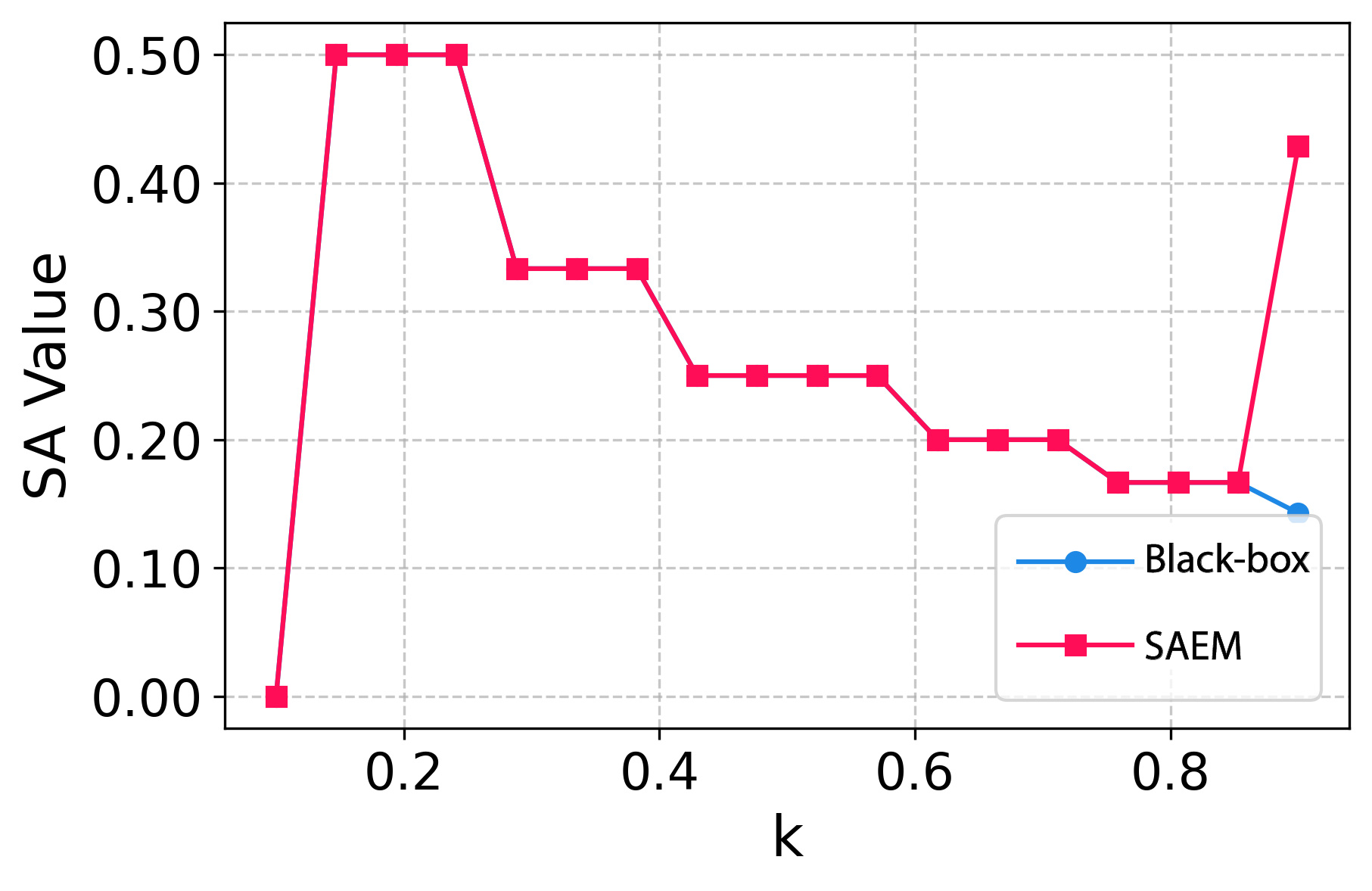}
        % \caption{After}
    \end{subfigure}
        \hfill
        \begin{subfigure}[t]{0.24\textwidth}
        \centering
        \includegraphics[width=\textwidth]{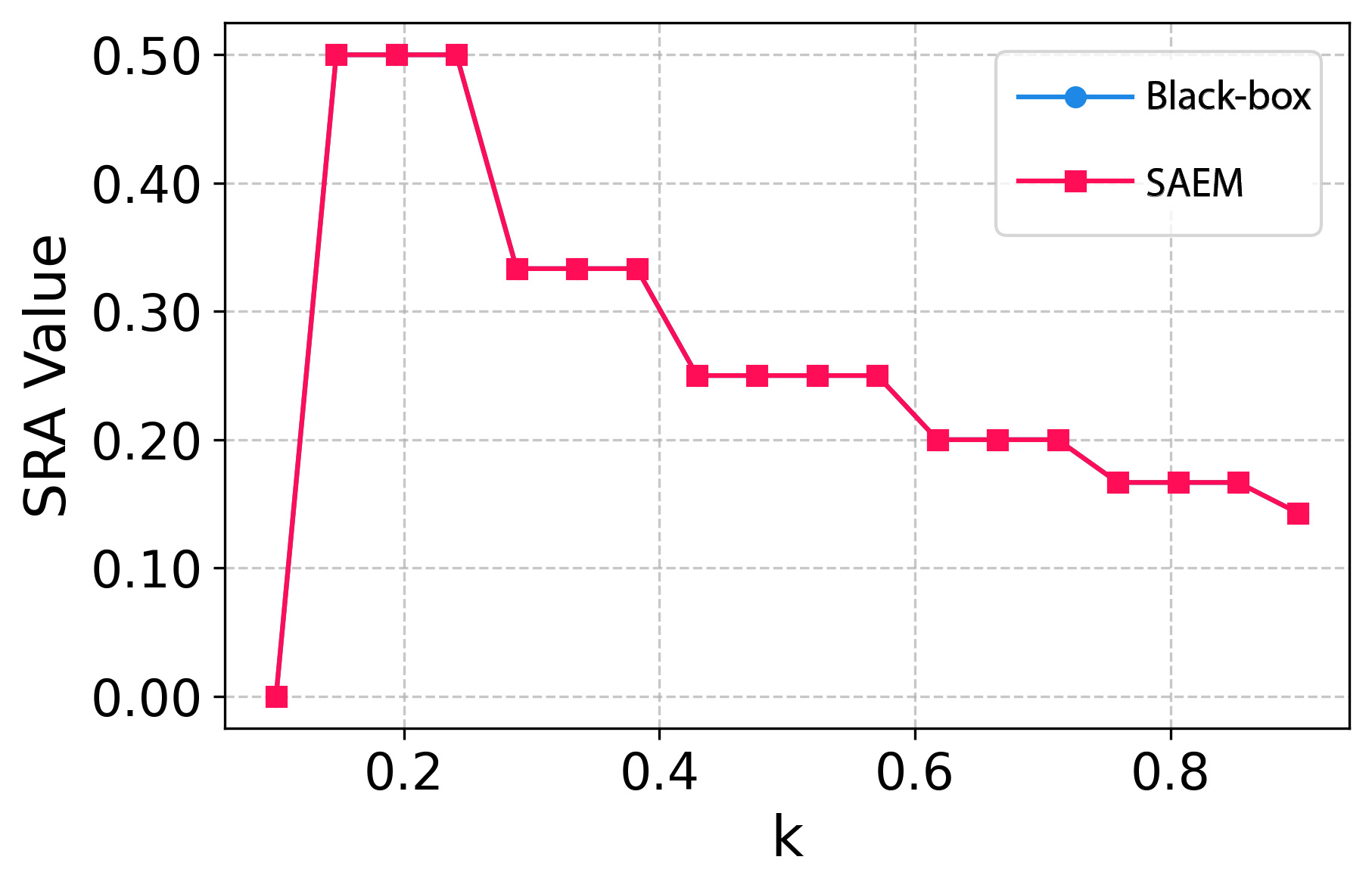}
    \end{subfigure}
    \caption{COMPAS Dataset: LR (top row) and ANN (bottom row) in blue; corresponding SAEMs in red.}
    \end{subfigure}
\caption{\label{fig:adult_income_comparison_k} Comparison of metrics (FA, RA, SA, SRA) between delivered models (LR and ANN) and the identified SAEMs for varying $k$ values on different datasets.} 
\end{figure*}

\begin{figure*}[]
\centering
    \begin{subfigure}{\textwidth}
        \centering
        \begin{subfigure}[t]{0.24\textwidth}
            \centering
            \includegraphics[width=\textwidth]{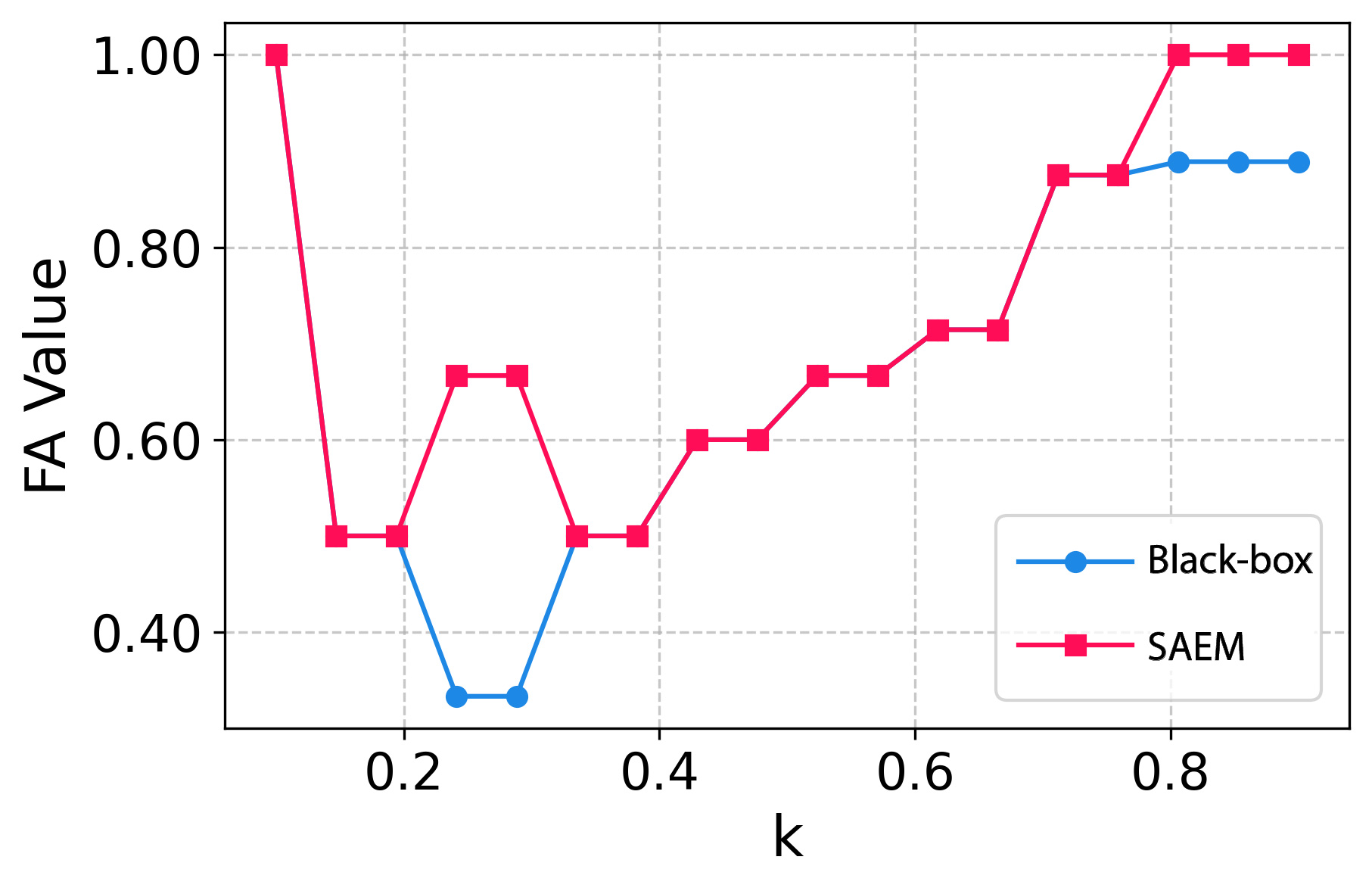}
            % \caption{FA improvement on Ground Truth}
        \end{subfigure}%
        \hfill
        \begin{subfigure}[t]{0.24\textwidth}
            \centering
            \includegraphics[width=\textwidth]{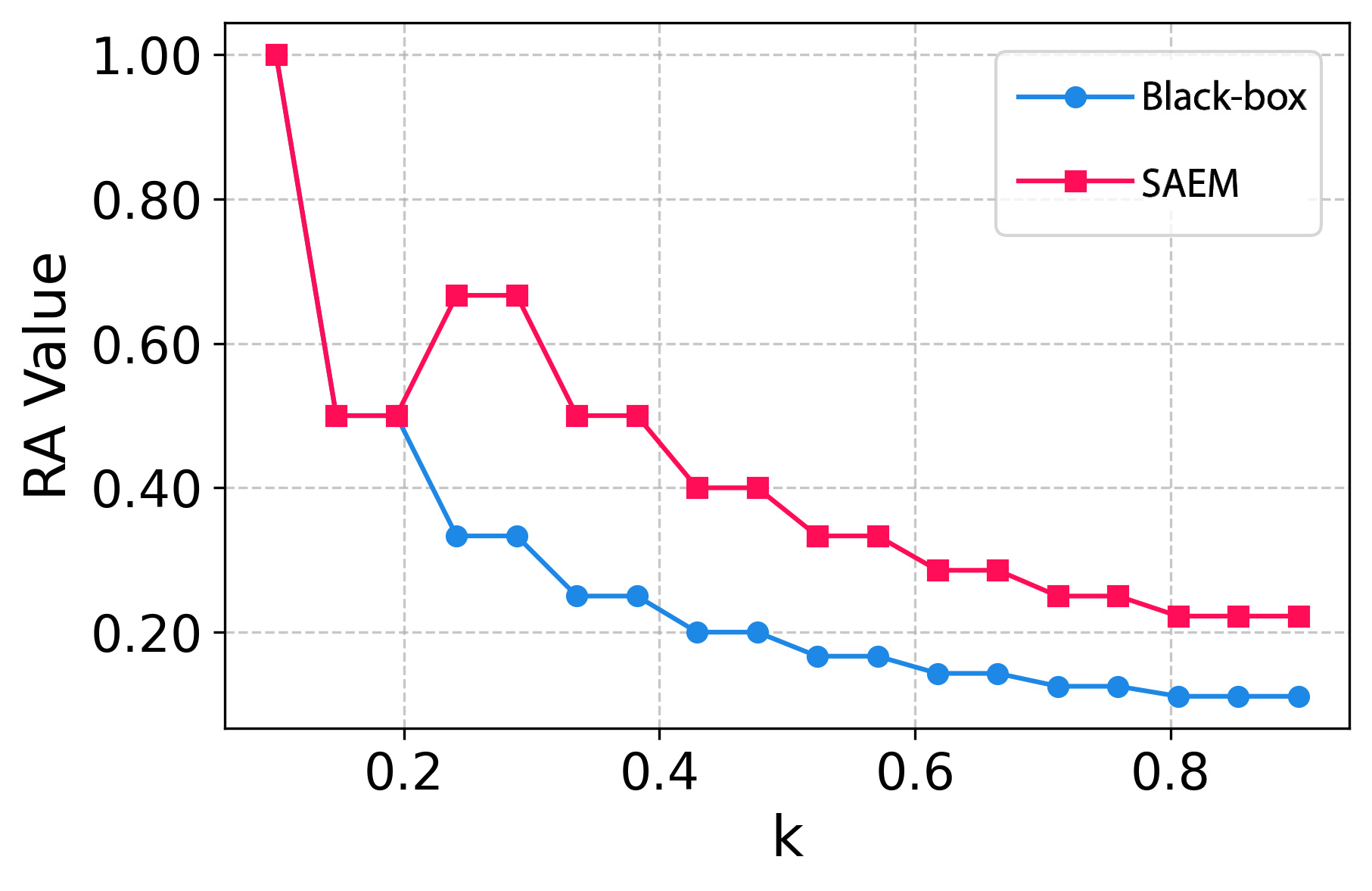}
            % \caption{RA improvement on Ground Truth}
        \end{subfigure}
        \hfill
        \begin{subfigure}[t]{0.24\textwidth}
            \centering
            \includegraphics[width=\textwidth]{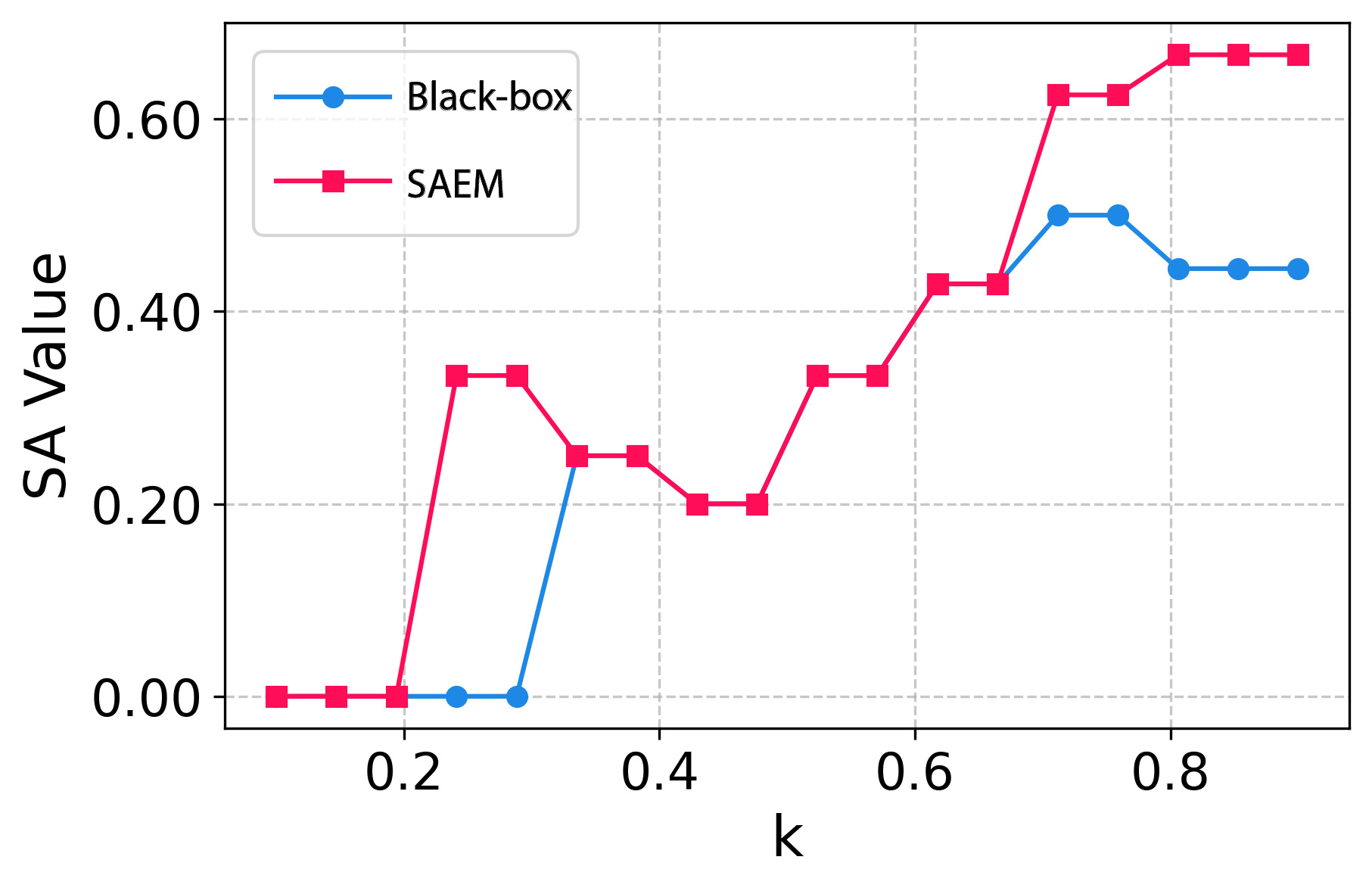}
            % \caption{SA improvement on Ground Truth}
        \end{subfigure}
        \hfill
        \begin{subfigure}[t]{0.24\textwidth}
            \centering
            \includegraphics[width=\textwidth]{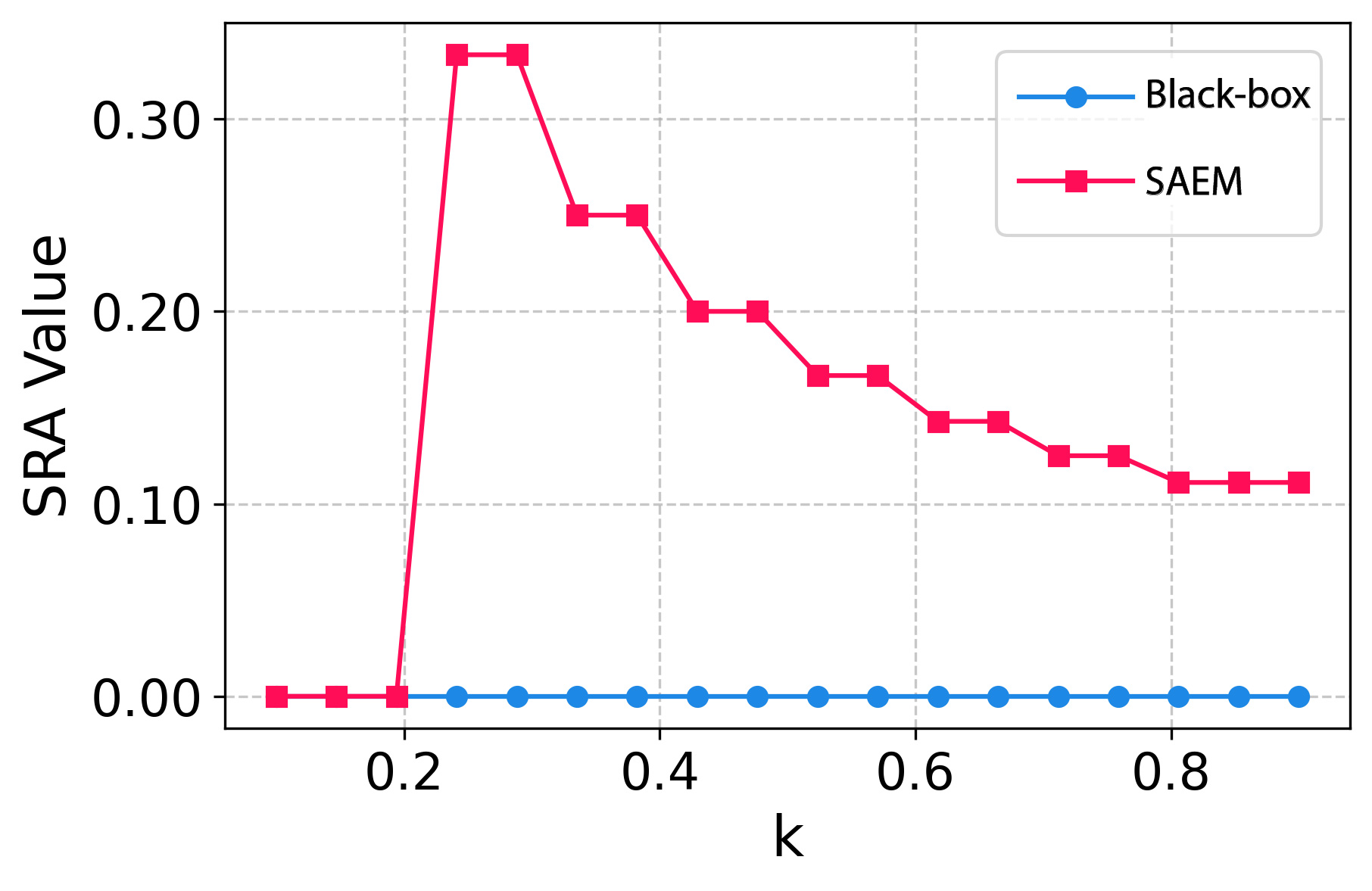}
            % \caption{SRA improvement on Ground Truth}
        \end{subfigure}
        \vfill
        \begin{subfigure}[t]{0.24\textwidth}
            \centering
            \includegraphics[width=\textwidth]{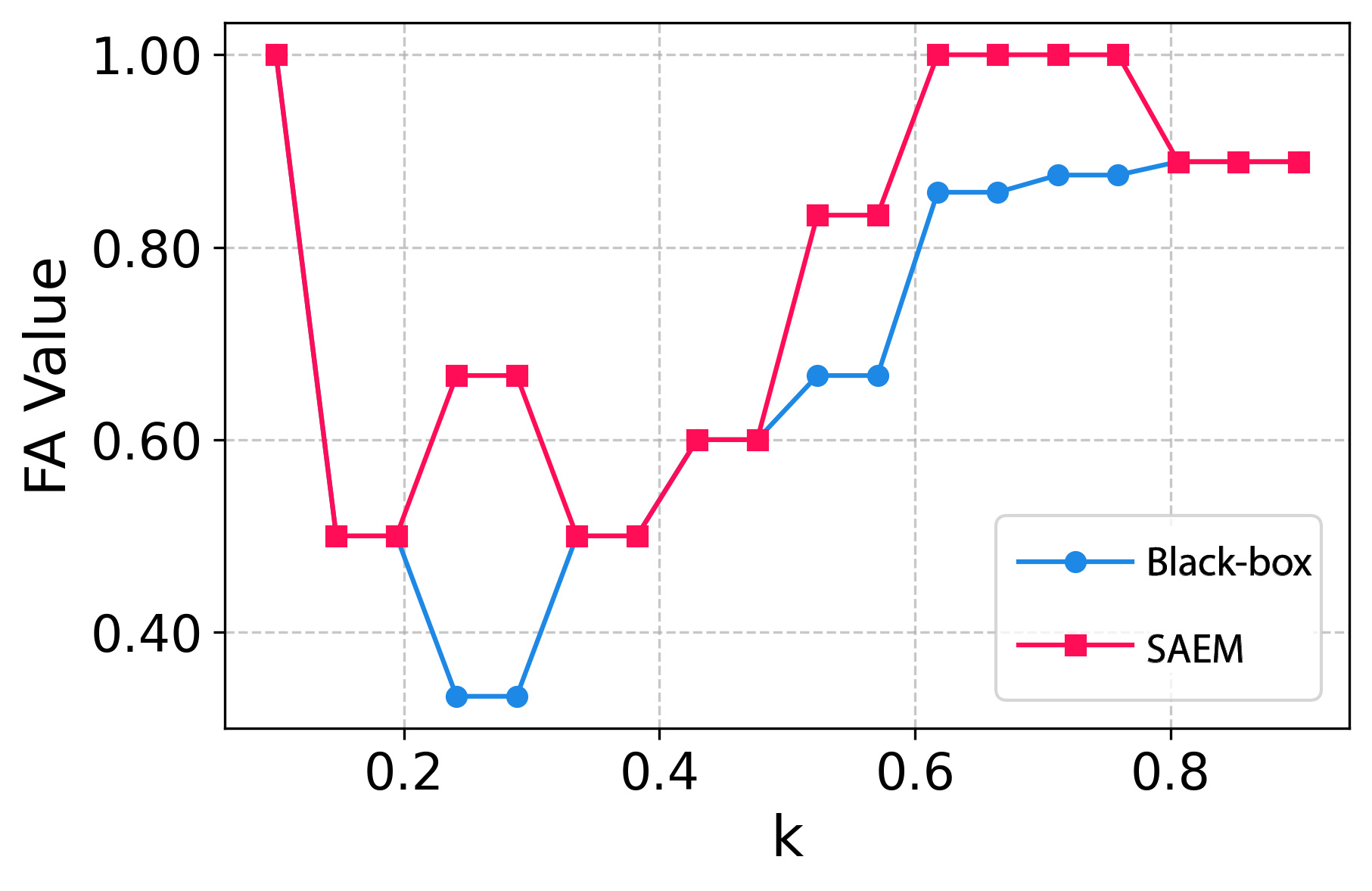}
            % \caption{Before}
        \end{subfigure}%
        \hfill
        \begin{subfigure}[t]{0.24\textwidth}
            \centering
            \includegraphics[width=\textwidth]{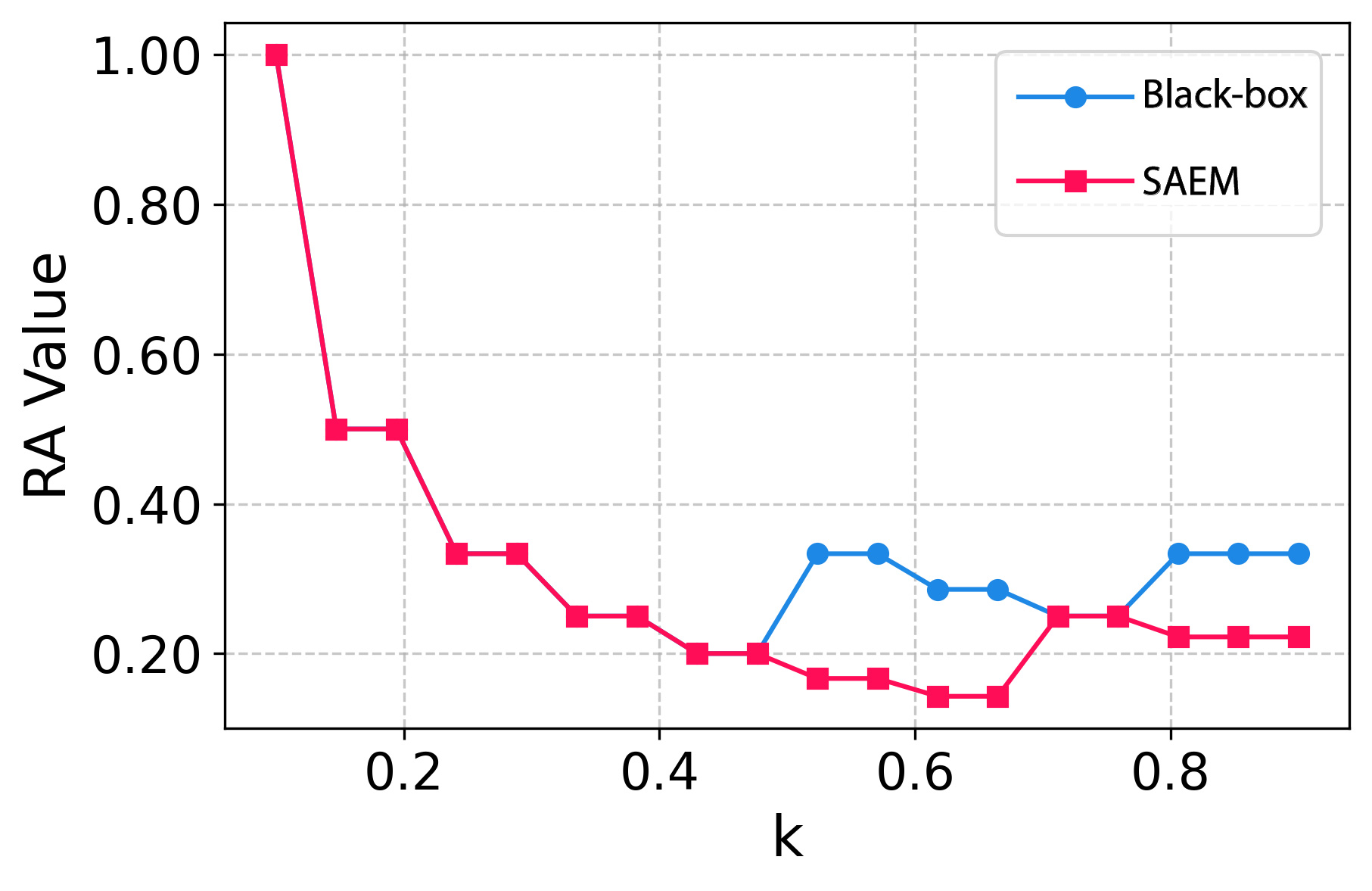}
            % \caption{After}
        \end{subfigure}
        \hfill
        \begin{subfigure}[t]{0.24\textwidth}
            \centering
            \includegraphics[width=\textwidth]{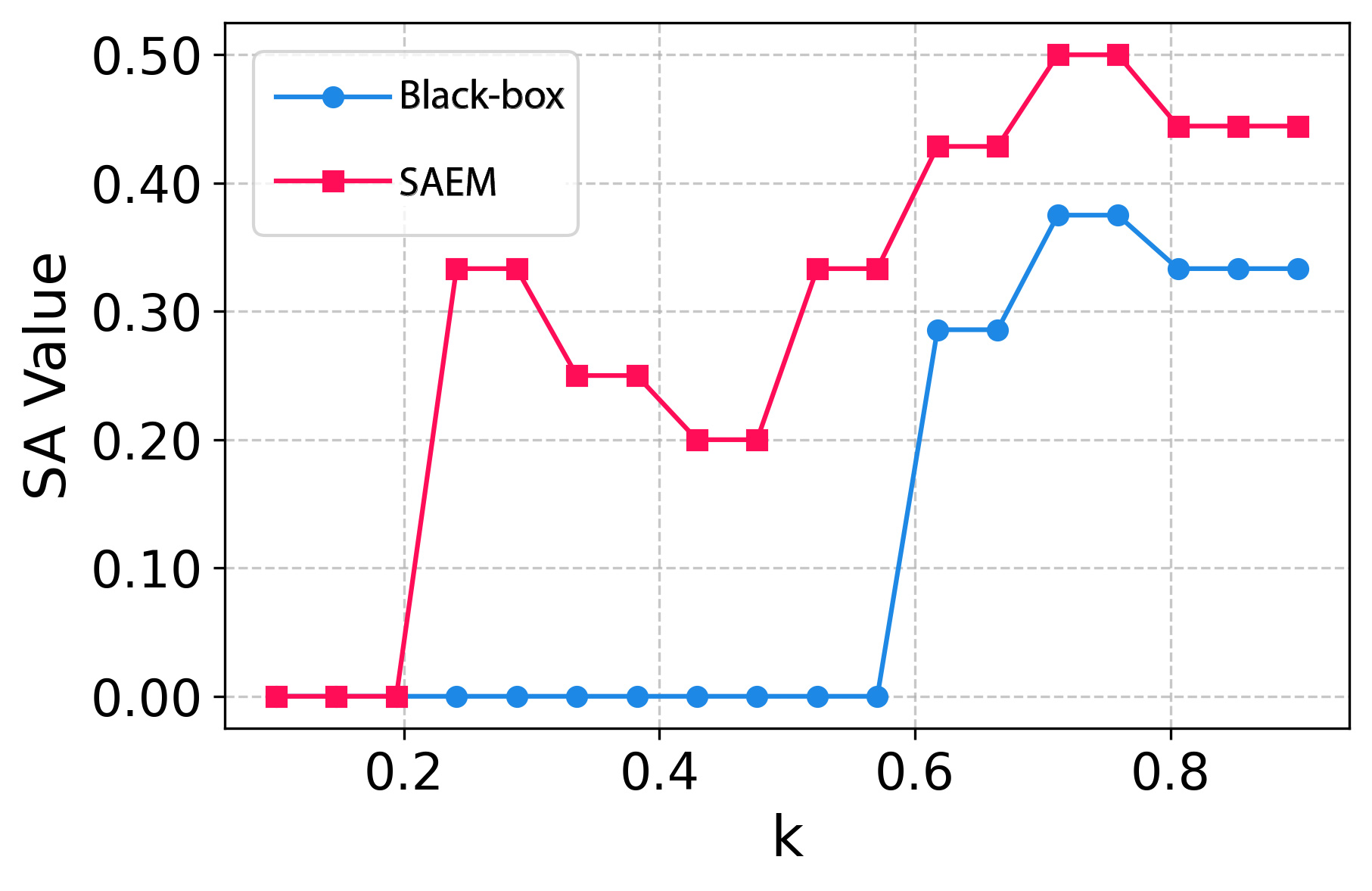}
            % \caption{After}
        \end{subfigure}
        \hfill
        \begin{subfigure}[t]{0.24\textwidth}
            \centering
            \includegraphics[width=\textwidth]{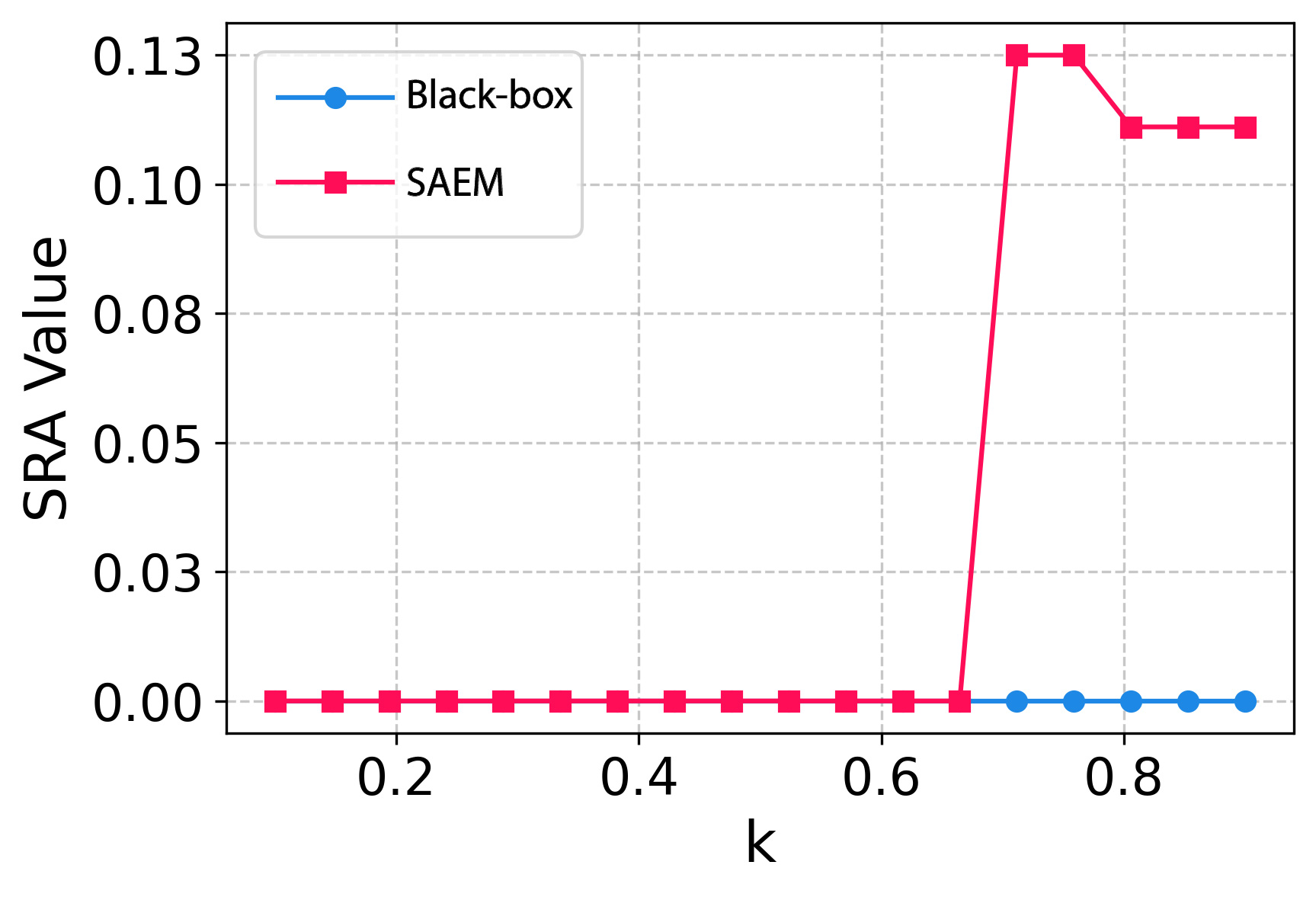}
            % \caption{After}
        \end{subfigure}
        \caption{Give Me Some Credit: LR (top row) and ANN (bottom row) in blue; corresponding SAEMs in red.}
    \end{subfigure}
        \vfill
    \begin{subfigure}{\textwidth}
        \centering
        \begin{subfigure}[t]{0.24\textwidth}
            \centering
            \includegraphics[width=\textwidth]{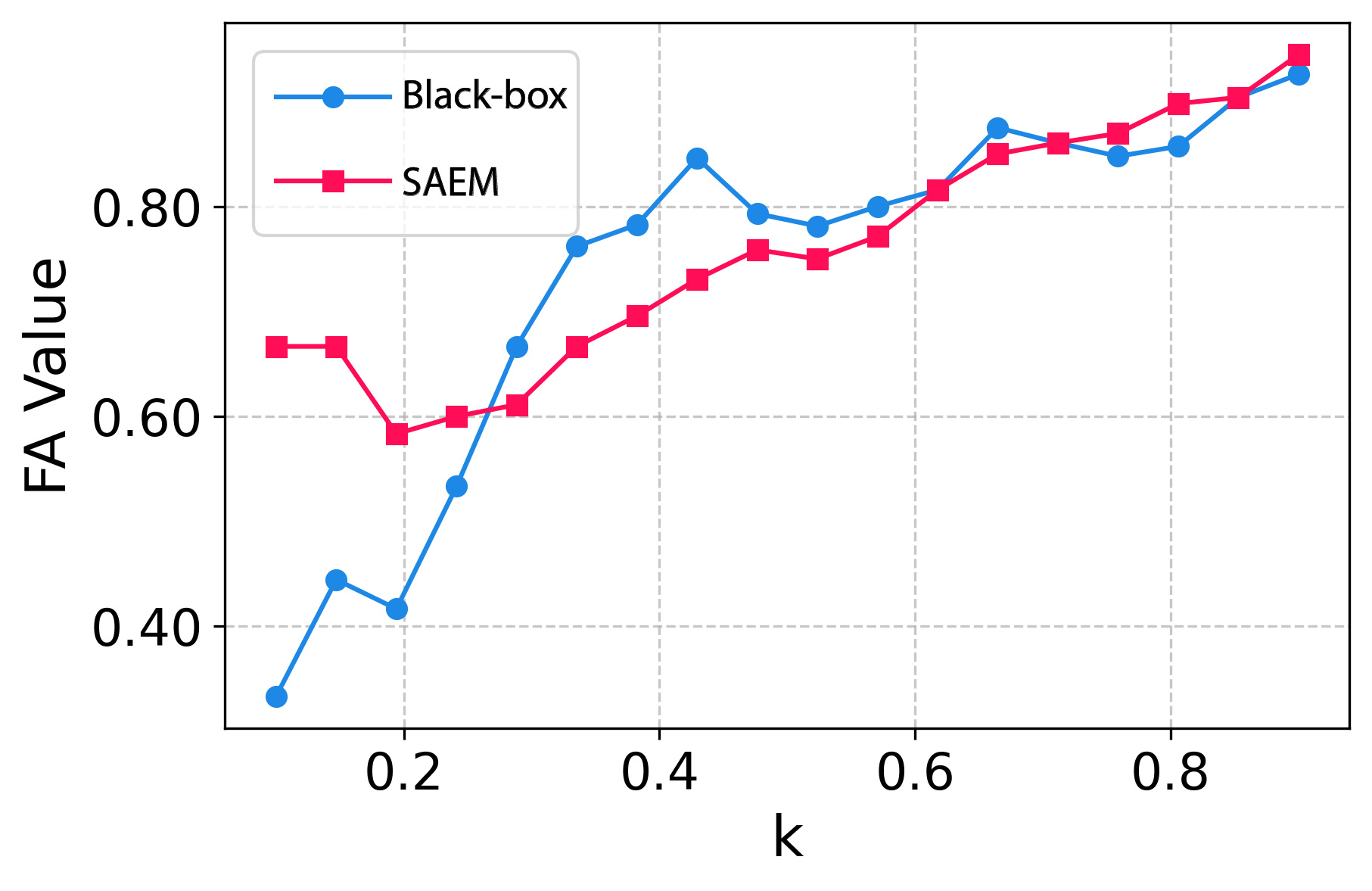}
            % \caption{FA improvement on Ground Truth}
        \end{subfigure}%
        \hfill
        \begin{subfigure}[t]{0.24\textwidth}
            \centering
            \includegraphics[width=\textwidth]{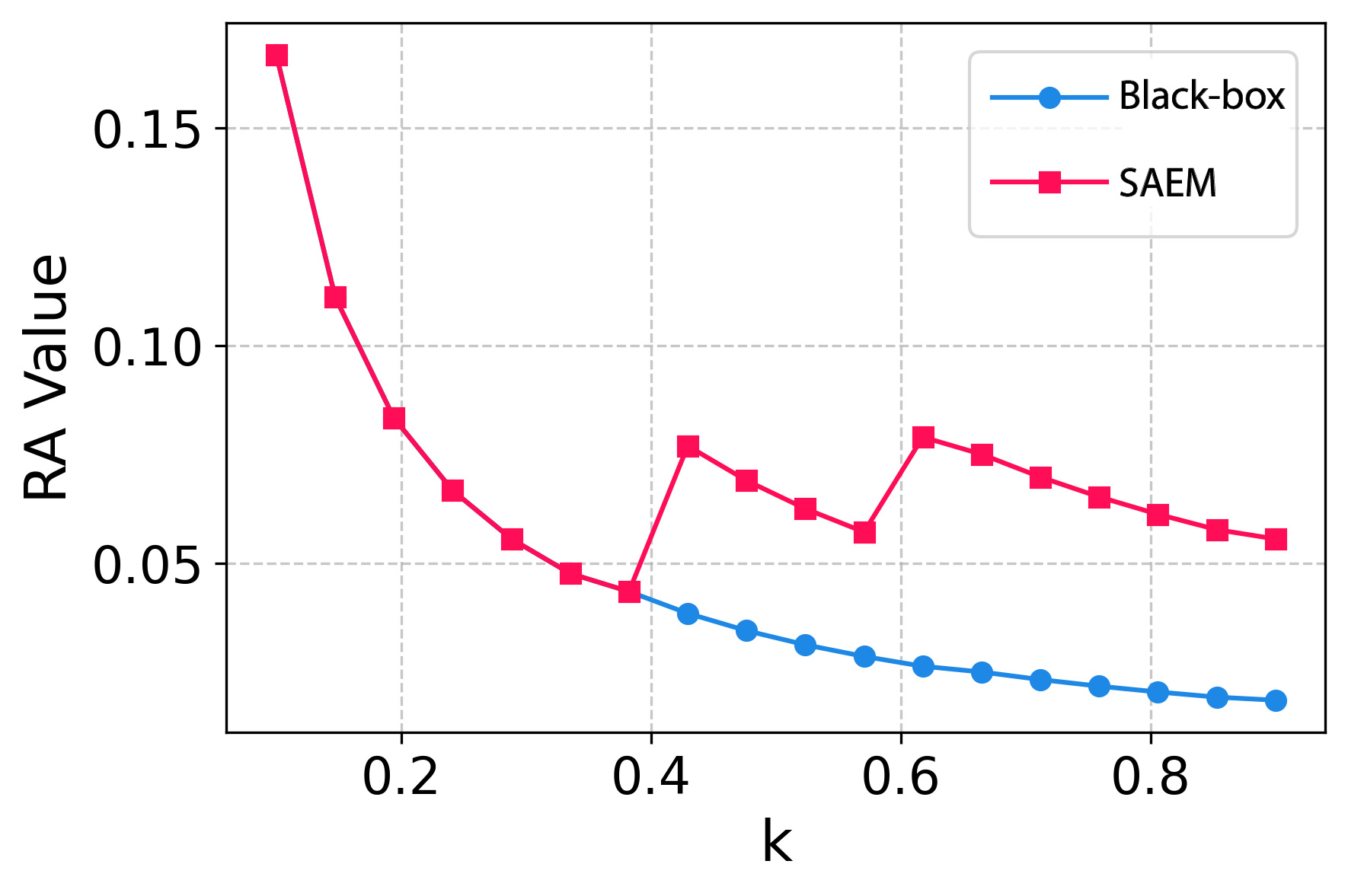}
            % \caption{RA improvement on Ground Truth}
        \end{subfigure}
        \hfill
        \begin{subfigure}[t]{0.24\textwidth}
            \centering
            \includegraphics[width=\textwidth]{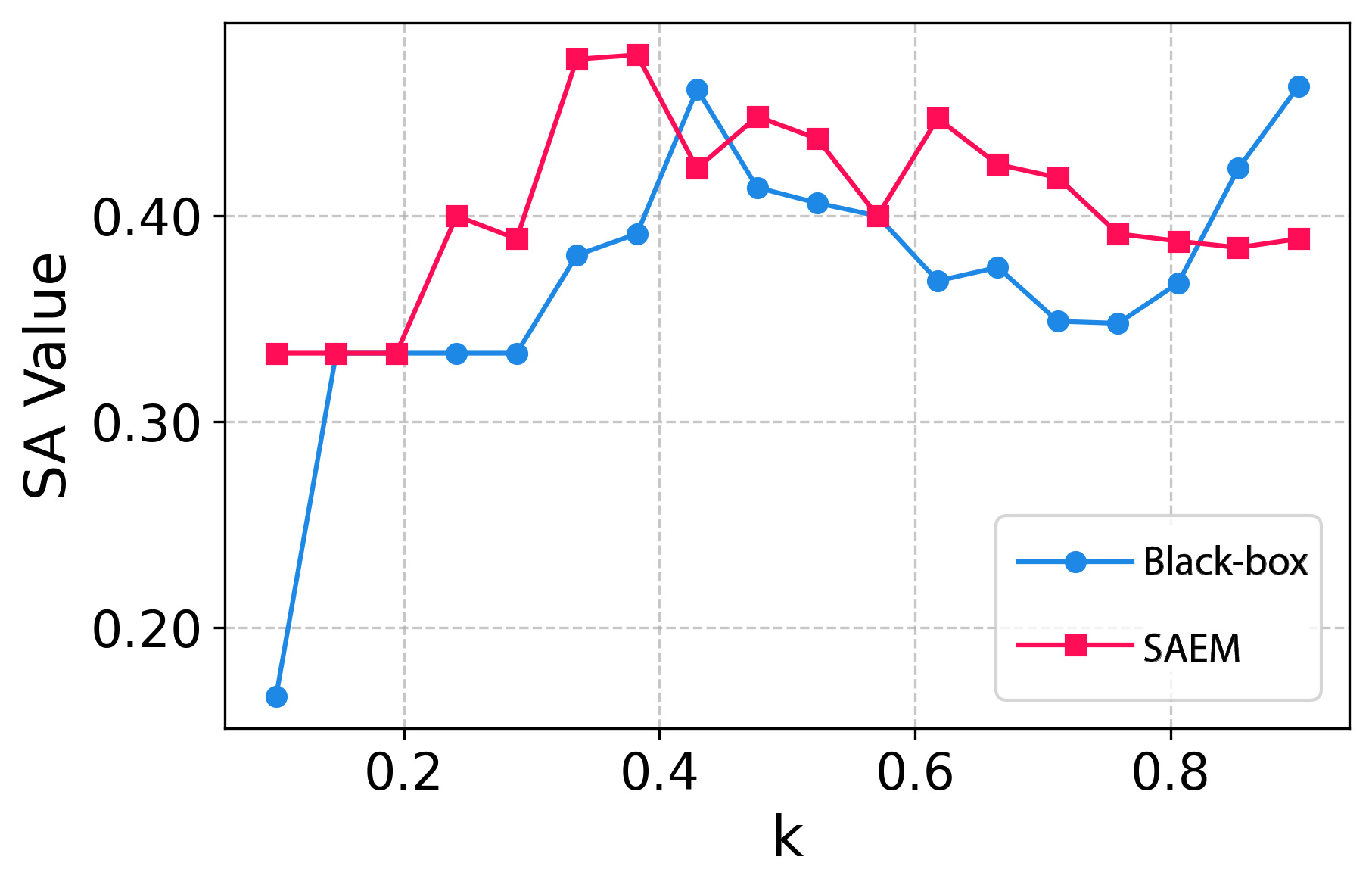}
            % \caption{SA improvement on Ground Truth}
        \end{subfigure}
        \hfill
        \begin{subfigure}[t]{0.24\textwidth}
            \centering
            \includegraphics[width=\textwidth]{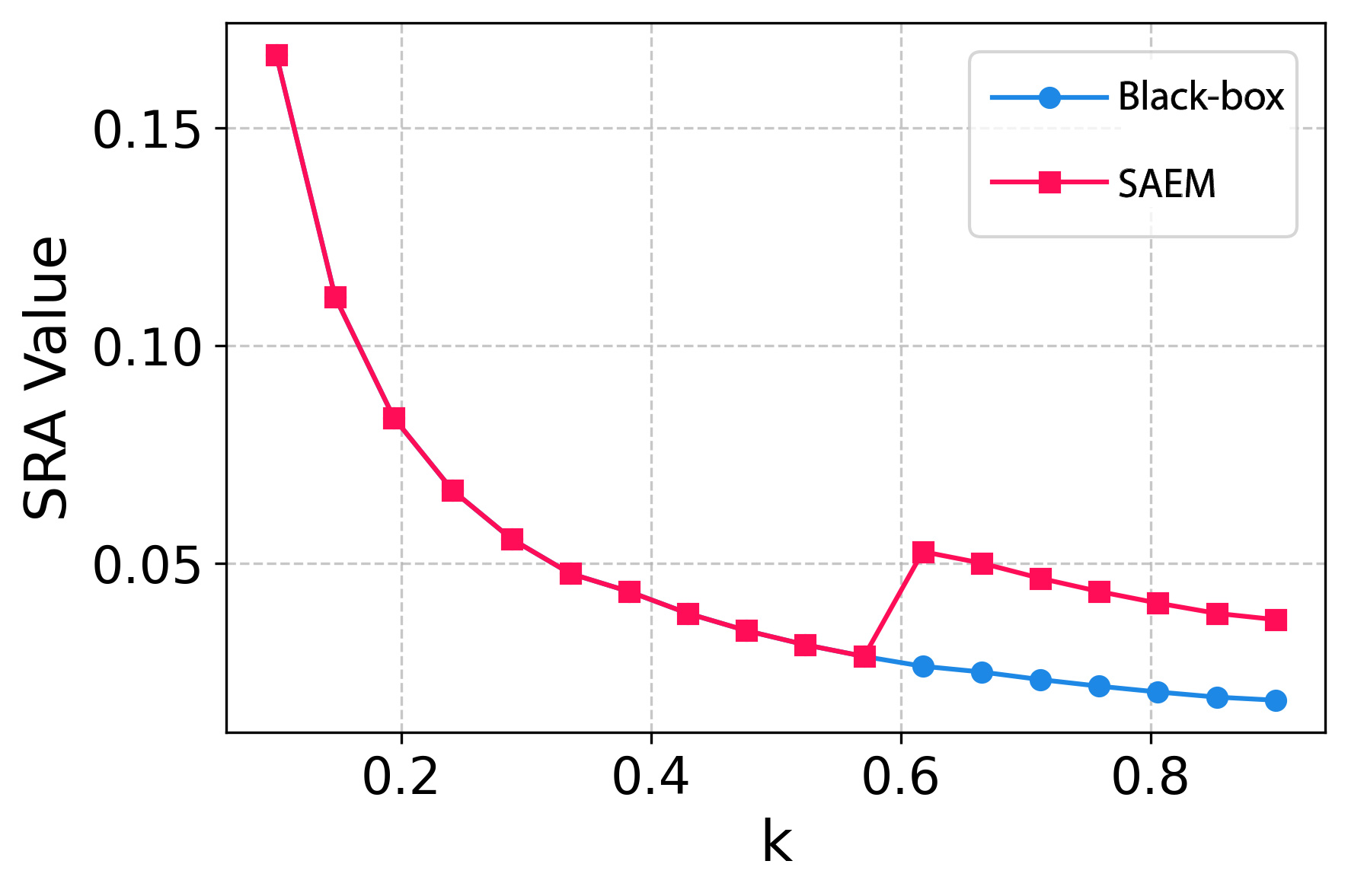}
            % \caption{SRA improvement on Ground Truth}
        \end{subfigure}
        \vfill
        \begin{subfigure}[t]{0.24\textwidth}
            \centering
            \includegraphics[width=\textwidth]{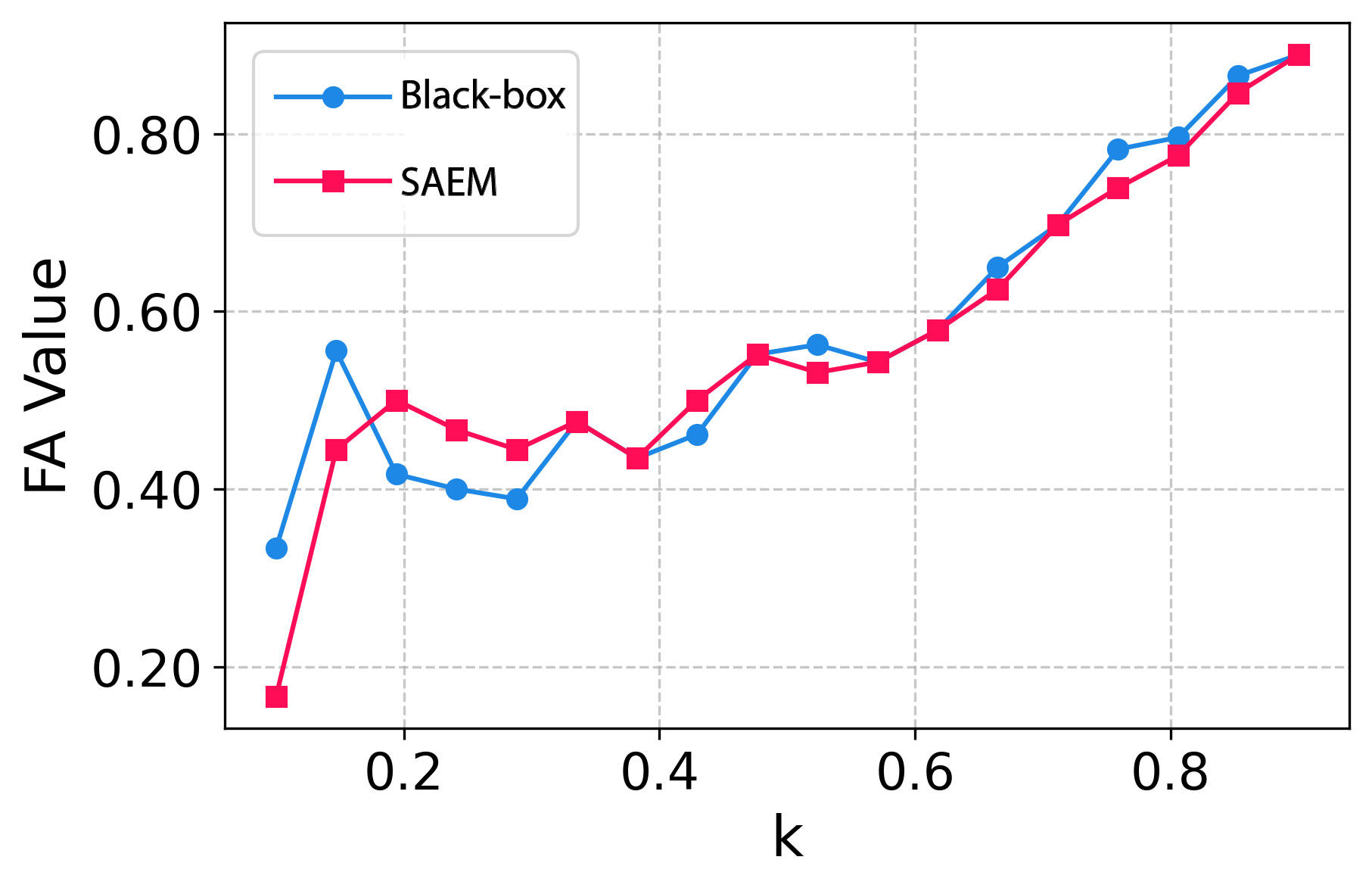}
            % \caption{Before}
        \end{subfigure}%
        \hfill
        \begin{subfigure}[t]{0.24\textwidth}
            \centering
            \includegraphics[width=\textwidth]{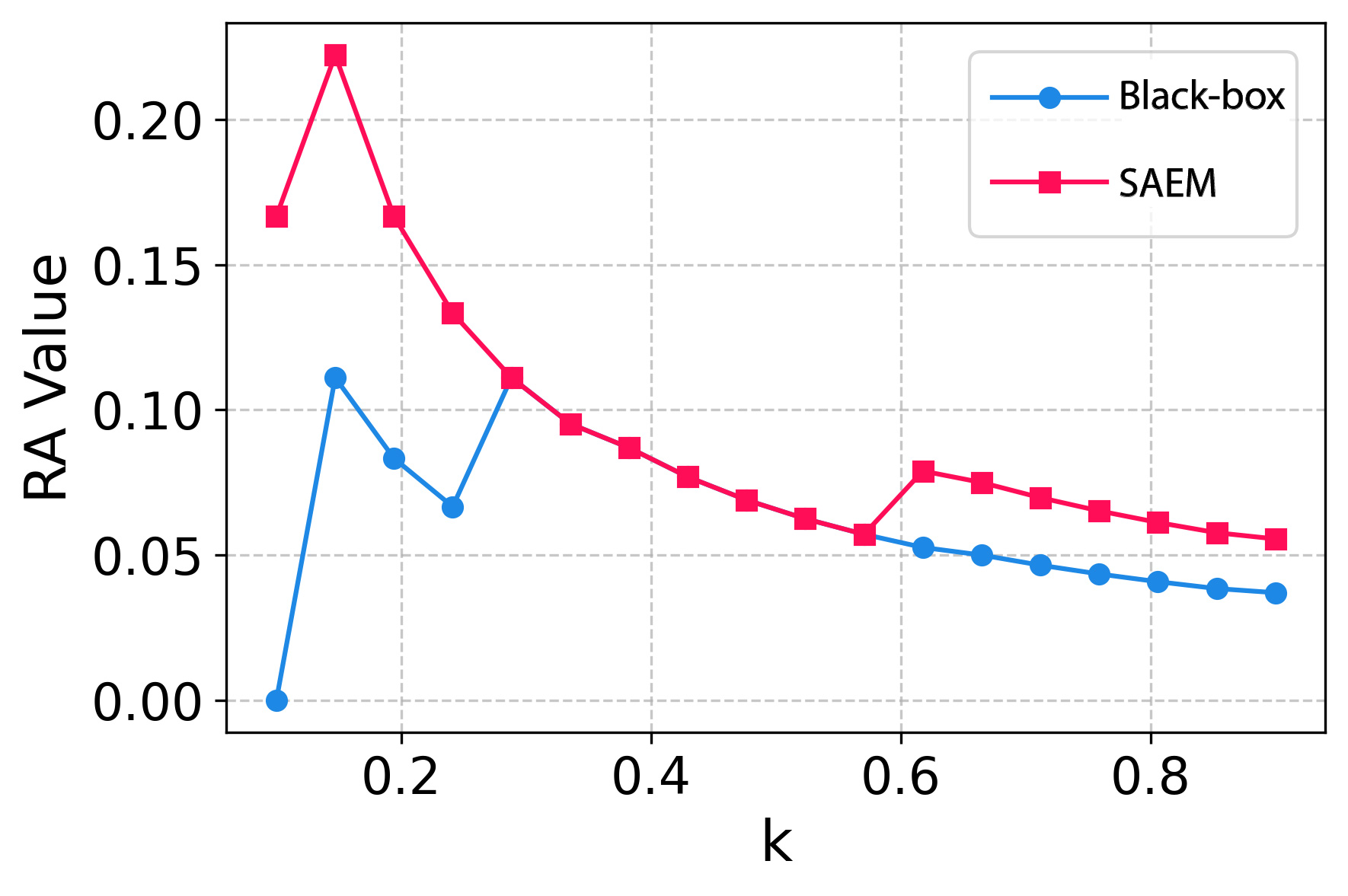}
            % \caption{After}
        \end{subfigure}
        \hfill
        \begin{subfigure}[t]{0.24\textwidth}
            \centering
            \includegraphics[width=\textwidth]{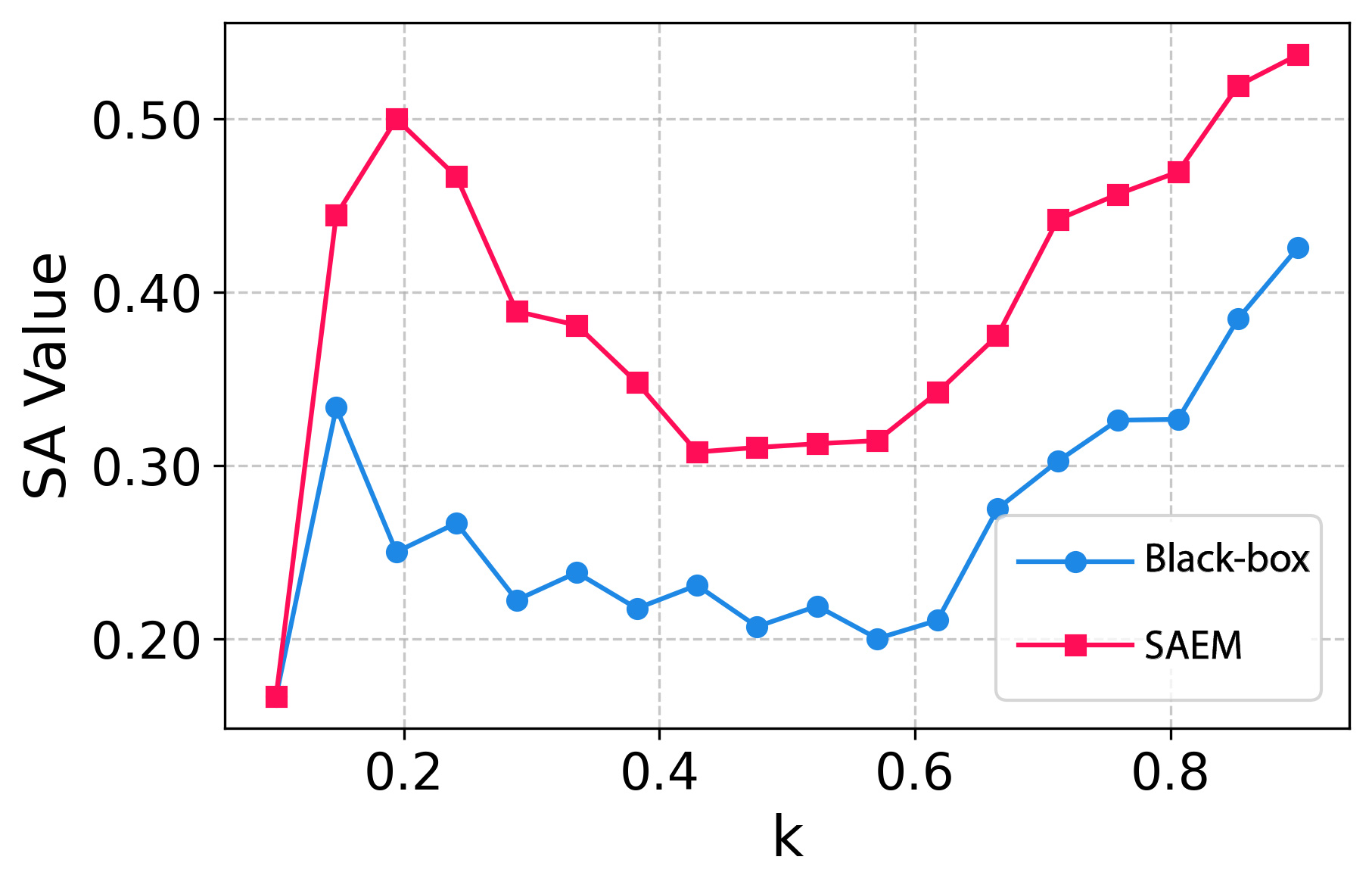}
            % \caption{After}
        \end{subfigure}
        \hfill
        \begin{subfigure}[t]{0.24\textwidth}
            \centering
            \includegraphics[width=\textwidth]{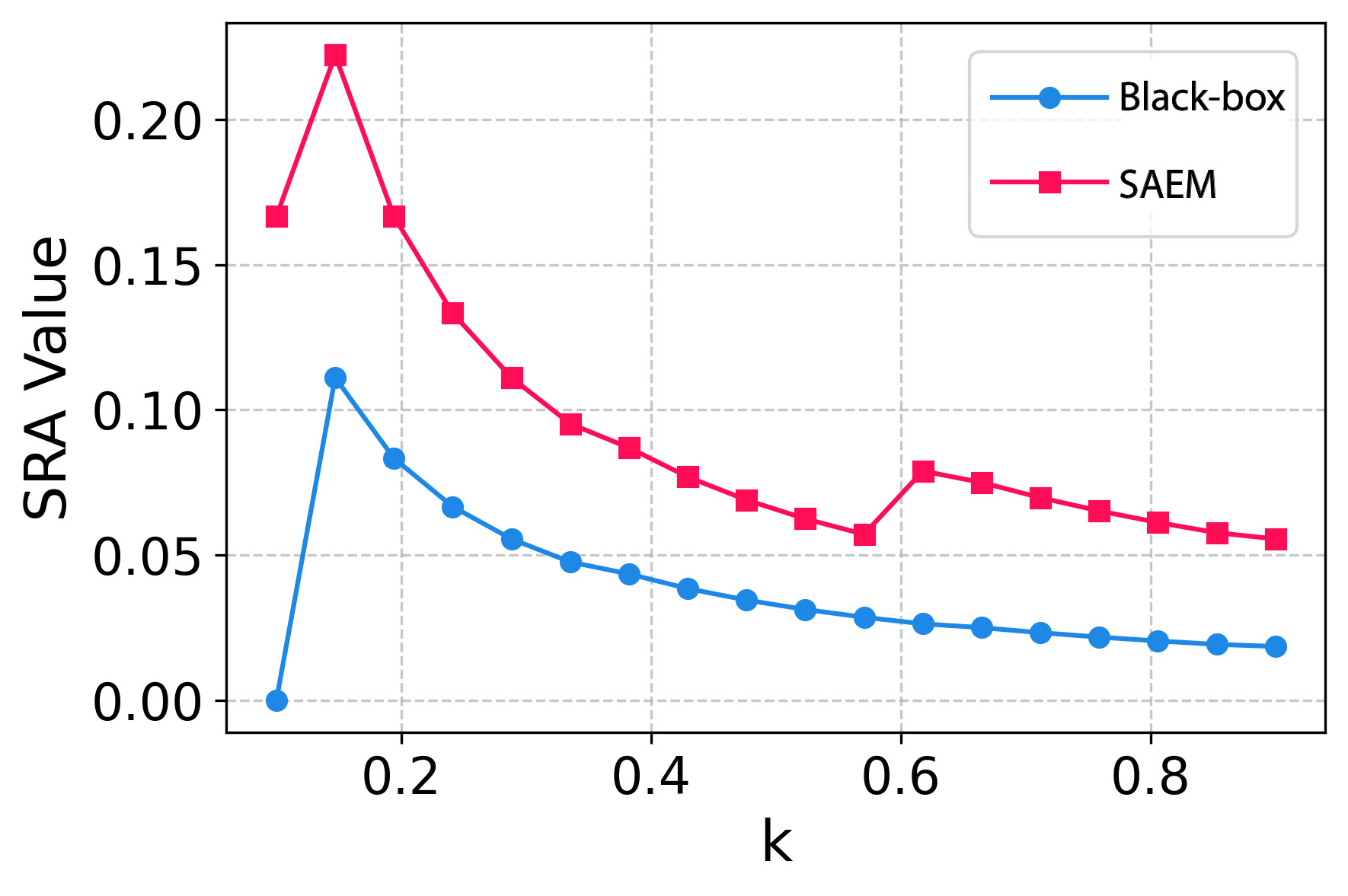}
            % \caption{After}
        \end{subfigure}
        \caption{German Credit: LR (top row) and ANN (bottom row) in blue; corresponding SAEMs in red.}
    \end{subfigure}
            \vfill
    \begin{subfigure}{\textwidth}
        \centering
        \begin{subfigure}[t]{0.24\textwidth}
            \centering
            \includegraphics[width=\textwidth]{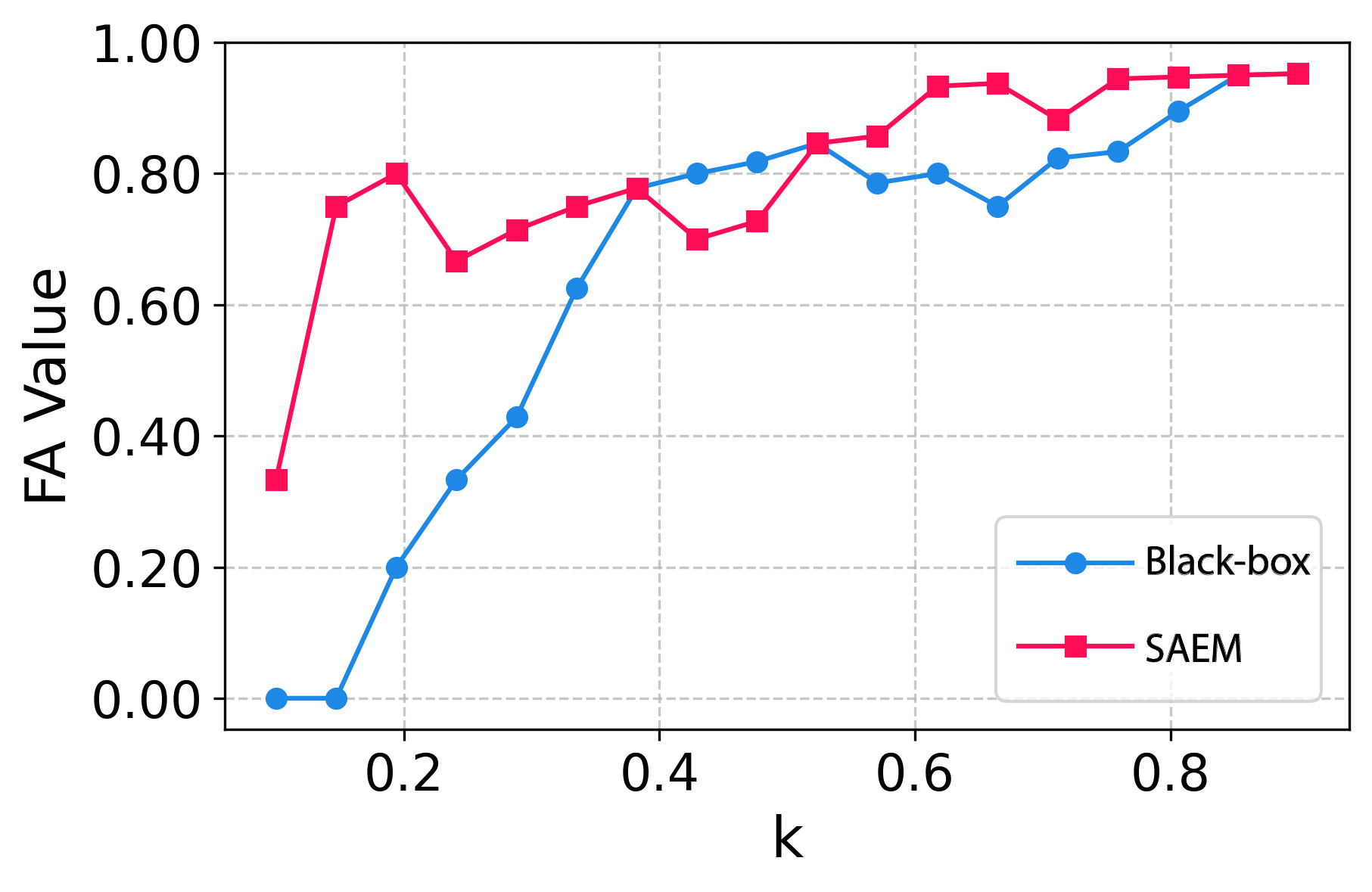}
            % \caption{FA improvement on Ground Truth}
        \end{subfigure}%
        \hfill
        \begin{subfigure}[t]{0.24\textwidth}
            \centering
            \includegraphics[width=\textwidth]{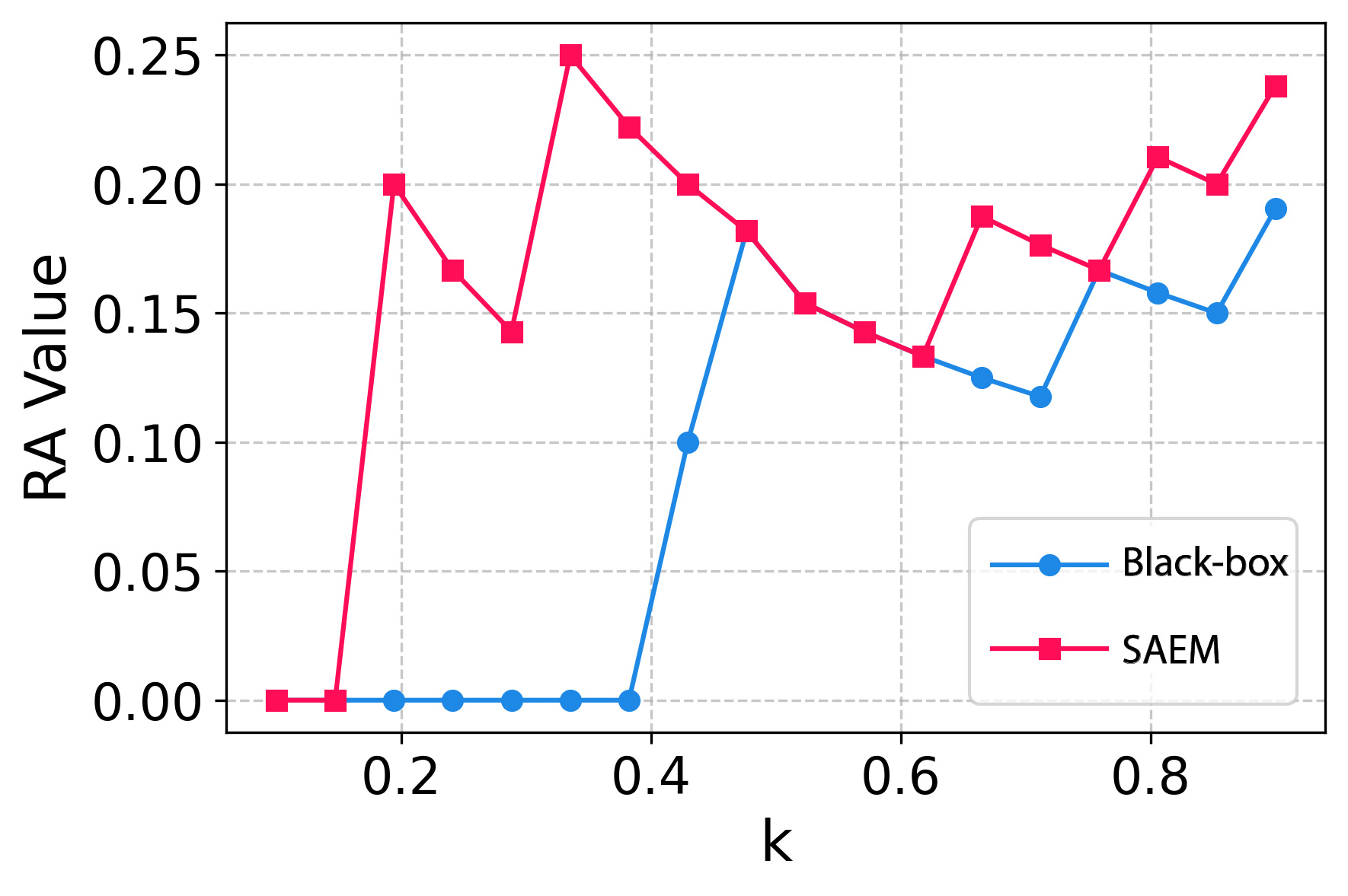}
            % \caption{RA improvement on Ground Truth}
        \end{subfigure}
        \hfill
        \begin{subfigure}[t]{0.24\textwidth}
            \centering
            \includegraphics[width=\textwidth]{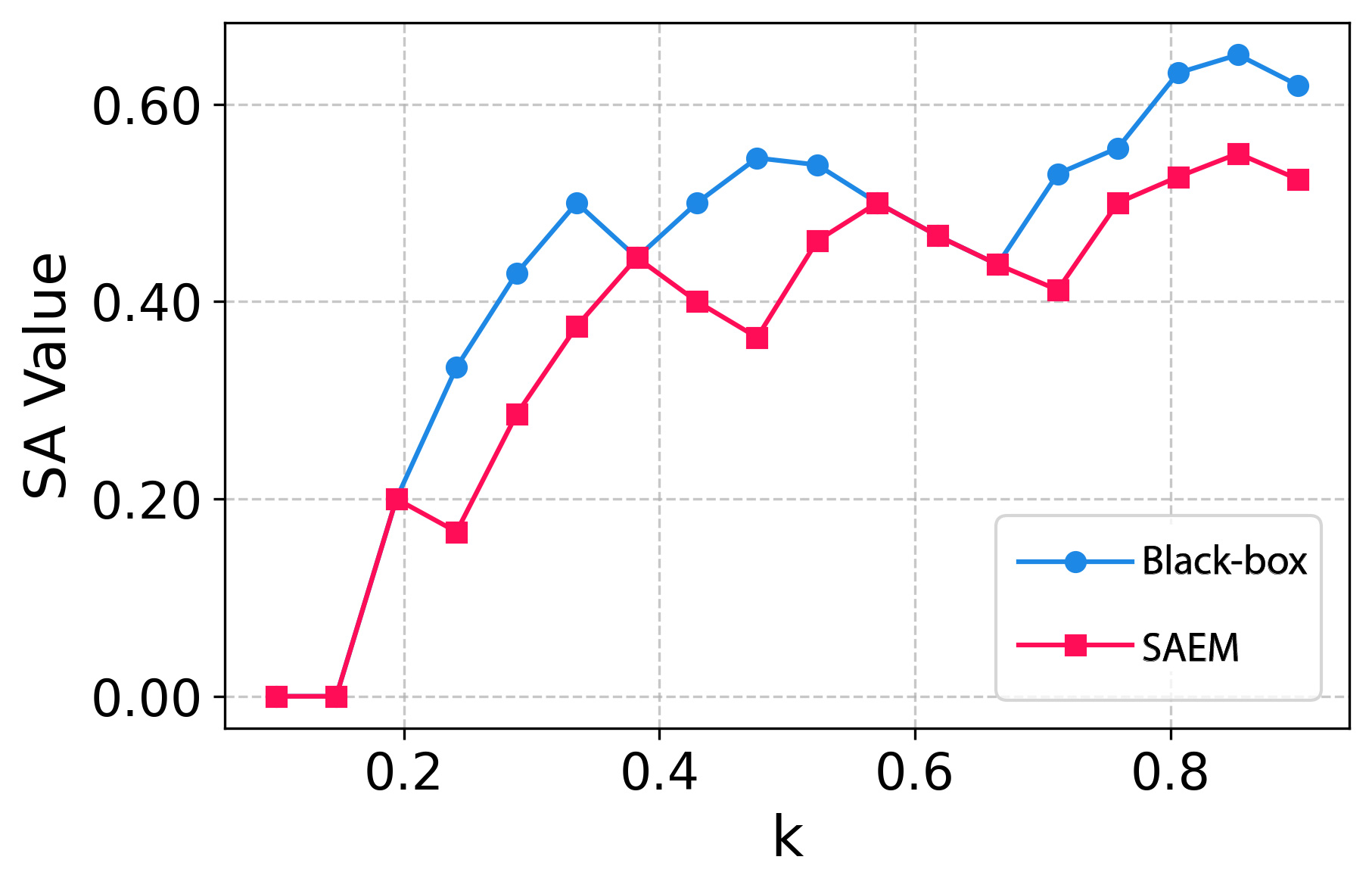}
            % \caption{SA improvement on Ground Truth}
        \end{subfigure}
        \hfill
        \begin{subfigure}[t]{0.24\textwidth}
            \centering
            \includegraphics[width=\textwidth]{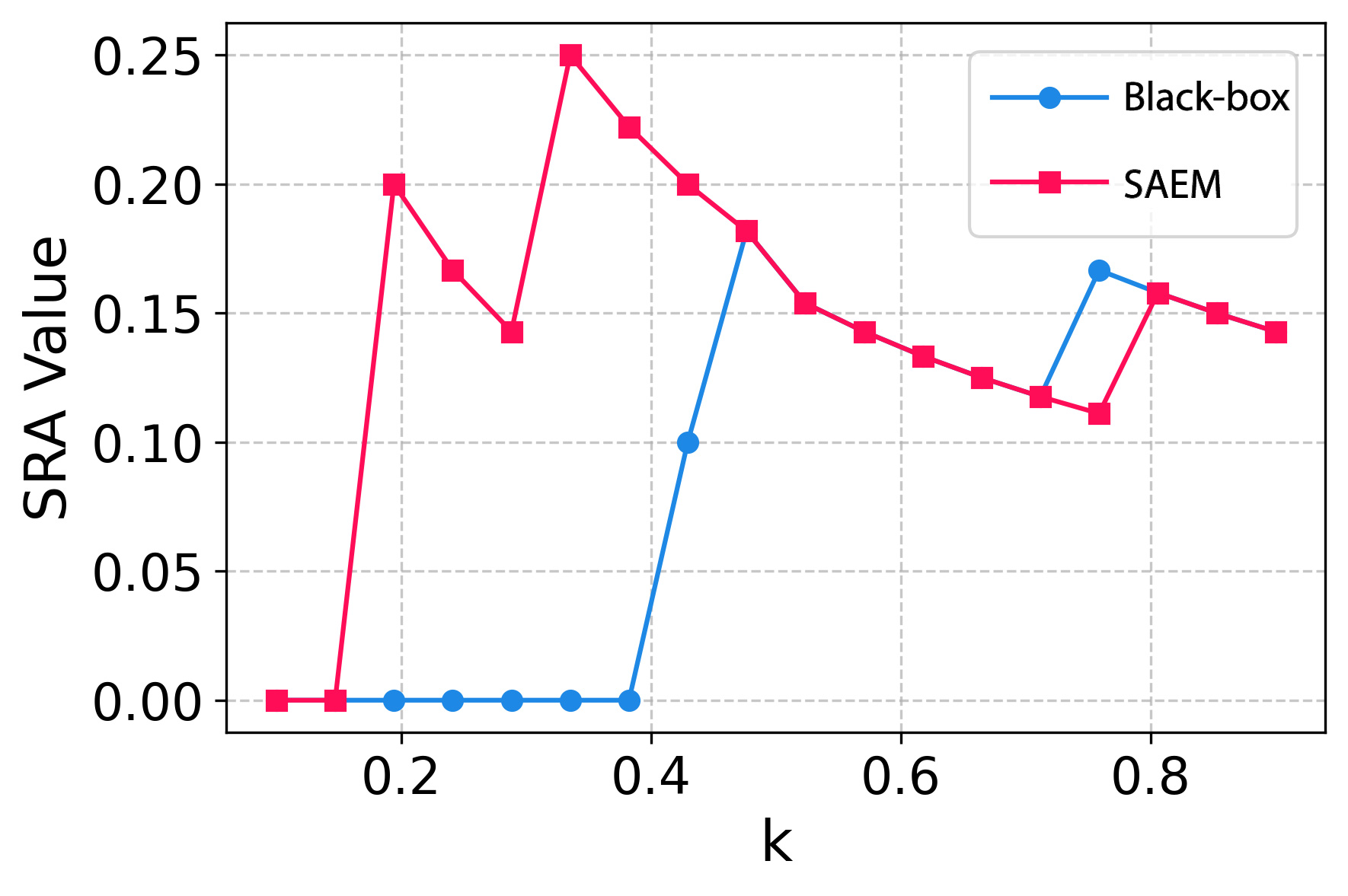}
            % \caption{SRA improvement on Ground Truth}
        \end{subfigure}
        \vfill
        \begin{subfigure}[t]{0.24\textwidth}
            \centering
            \includegraphics[width=\textwidth]{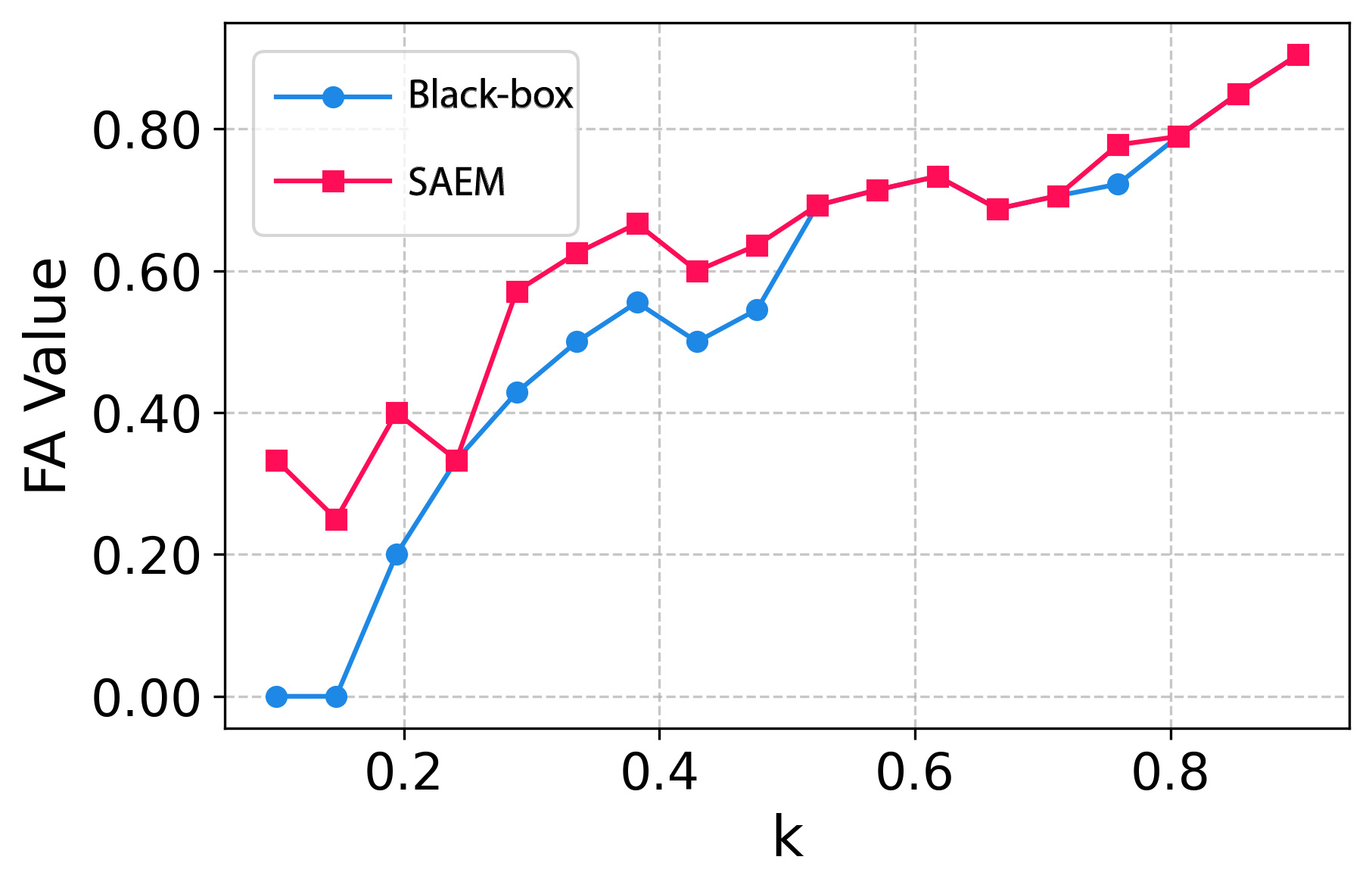}
            % \caption{Before}
        \end{subfigure}%
        \hfill
        \begin{subfigure}[t]{0.24\textwidth}
            \centering
            \includegraphics[width=\textwidth]{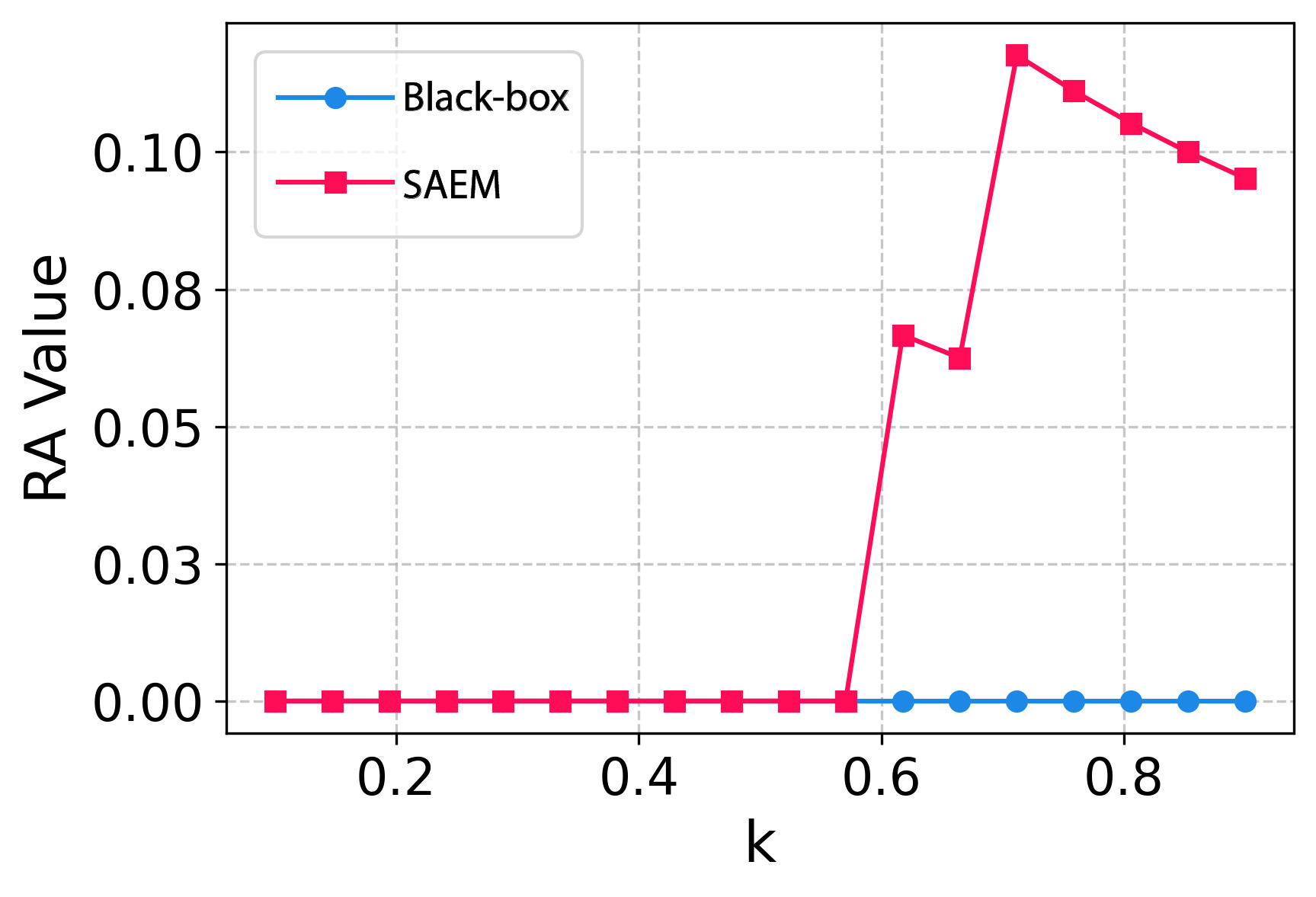}
            % \caption{After}
        \end{subfigure}
        \hfill
        \begin{subfigure}[t]{0.24\textwidth}
            \centering
            \includegraphics[width=\textwidth]{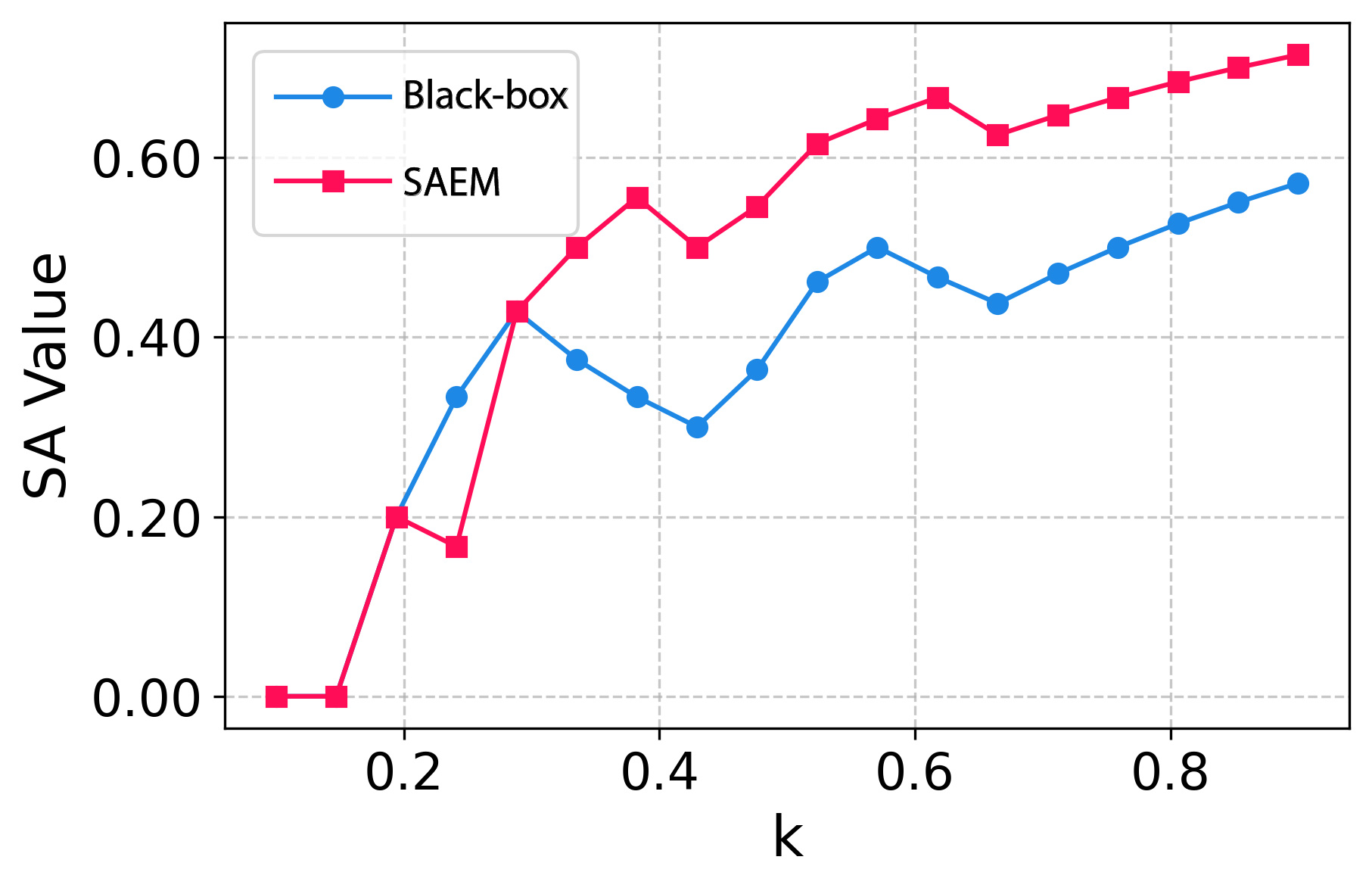}
            % \caption{After}
        \end{subfigure}
        \hfill
        \begin{subfigure}[t]{0.24\textwidth}
            \centering
            \includegraphics[width=\textwidth]{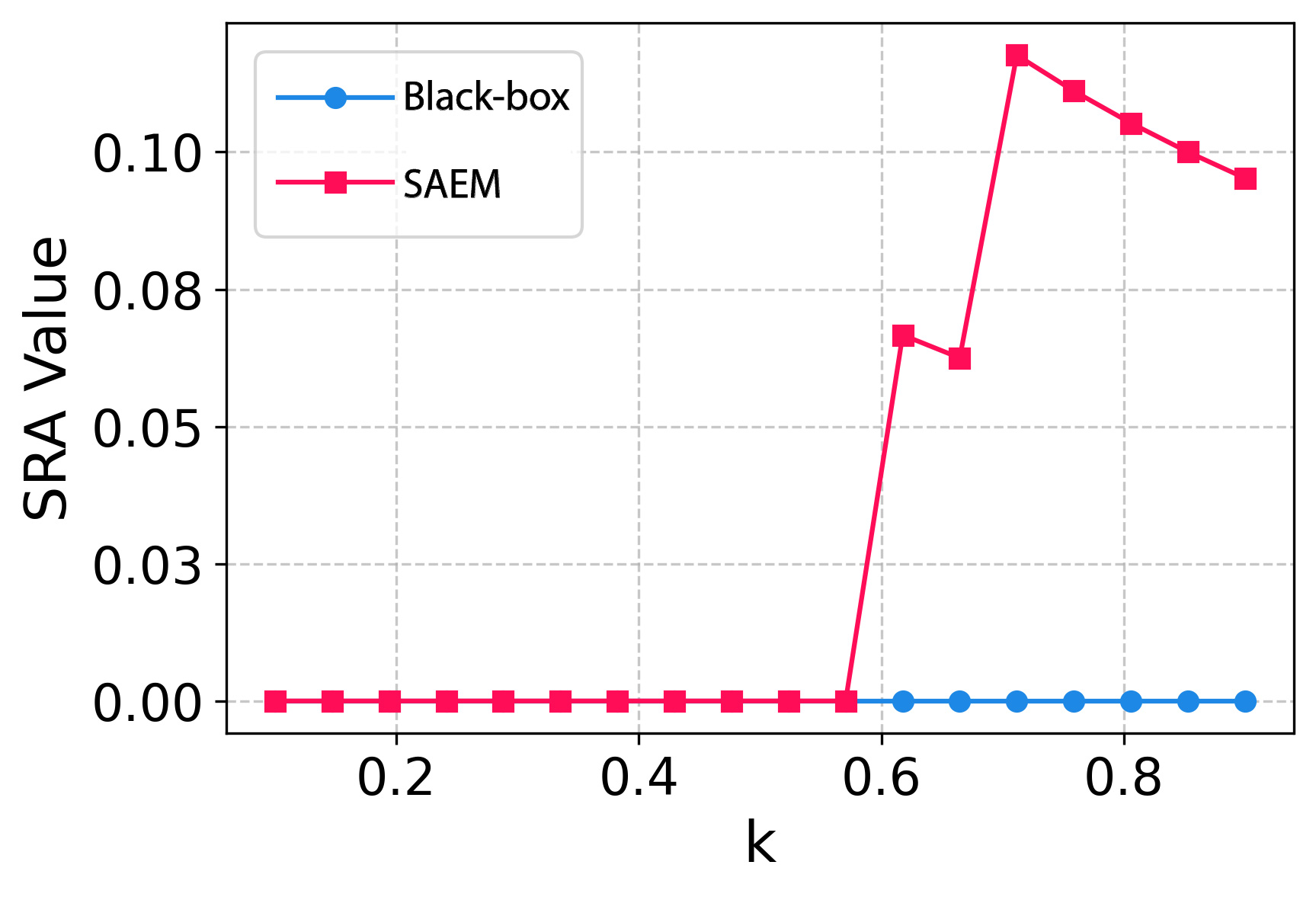}
            % \caption{After}
        \end{subfigure}
        \caption{HELOC: LR (top row) and ANN (bottom row) in blue; corresponding SAEMs in red.}
    \end{subfigure}
\caption{\label{fig:other_comparison_k} Comparison of metrics (FA, RA, SA, SRA) between delivered models (LR and ANN) and the identified SAEMs for varying $k$ values on different datasets.}
\end{figure*}

\section{Fairness Analysis in Subgroups}\label{sec:fairness-analysis}
Our fairness analysis, depicted in Fig. \ref{fig:fairness_comparison}, compares the faithfulness between the pre-trained LR and the identified SAEM on three datasets (Adult Income, Genman Credits, and COMPAS) that contain gender information (e.g., male and female). The provided model exhibits significant disparities in faithfulness metrics between majority and minority groups, indicating unfair explanations, particularly in the Adult Income dataset (Fig. \ref{fig:fairness_comparison} (a)).
The SAEMs identified by our framework reduce these inequities across all three datasets, showcasing an improvement in explanation fairness between subgroups.
\begin{figure*}[htp!]
\centering
    \begin{subfigure}[t]{0.3\textwidth}
        \centering
        \includegraphics[width=\textwidth]{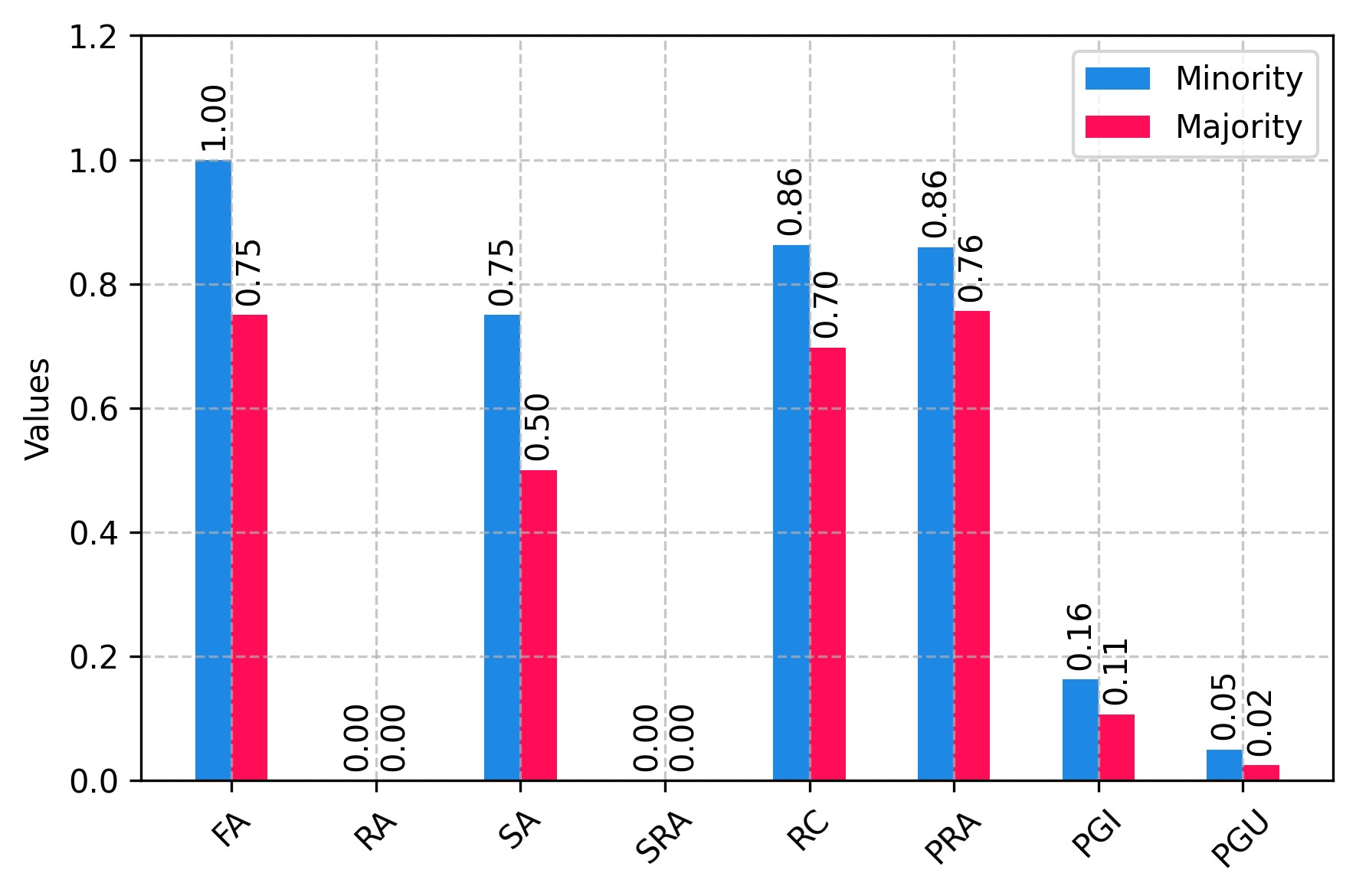}
    \end{subfigure}%
    \hfill
    \begin{subfigure}[t]{0.3\textwidth}
        \centering
        \includegraphics[width=\textwidth]{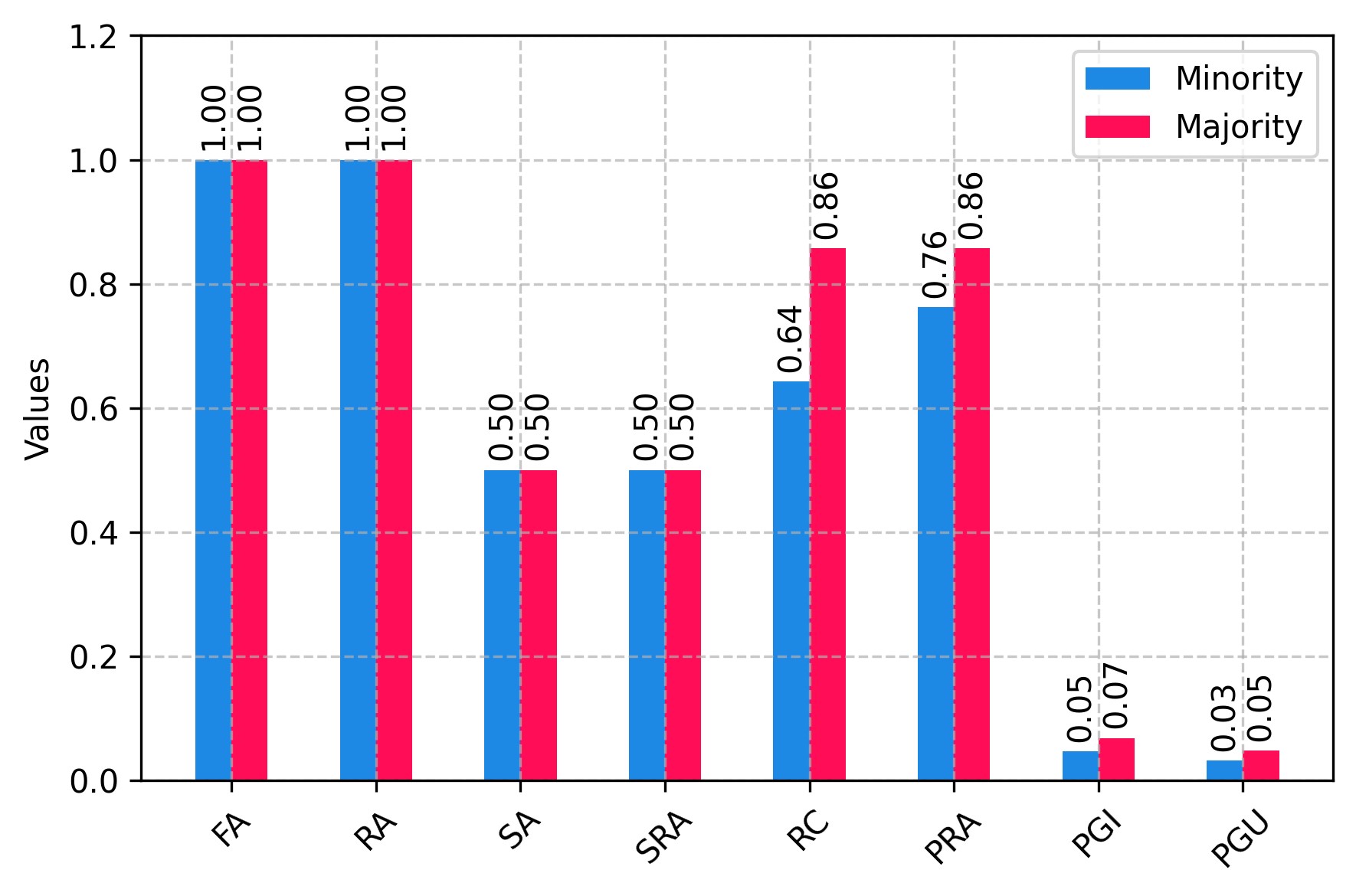}
    \end{subfigure}
    \hfill
    \begin{subfigure}[t]{0.3\textwidth}
        \centering
        \includegraphics[width=\textwidth]{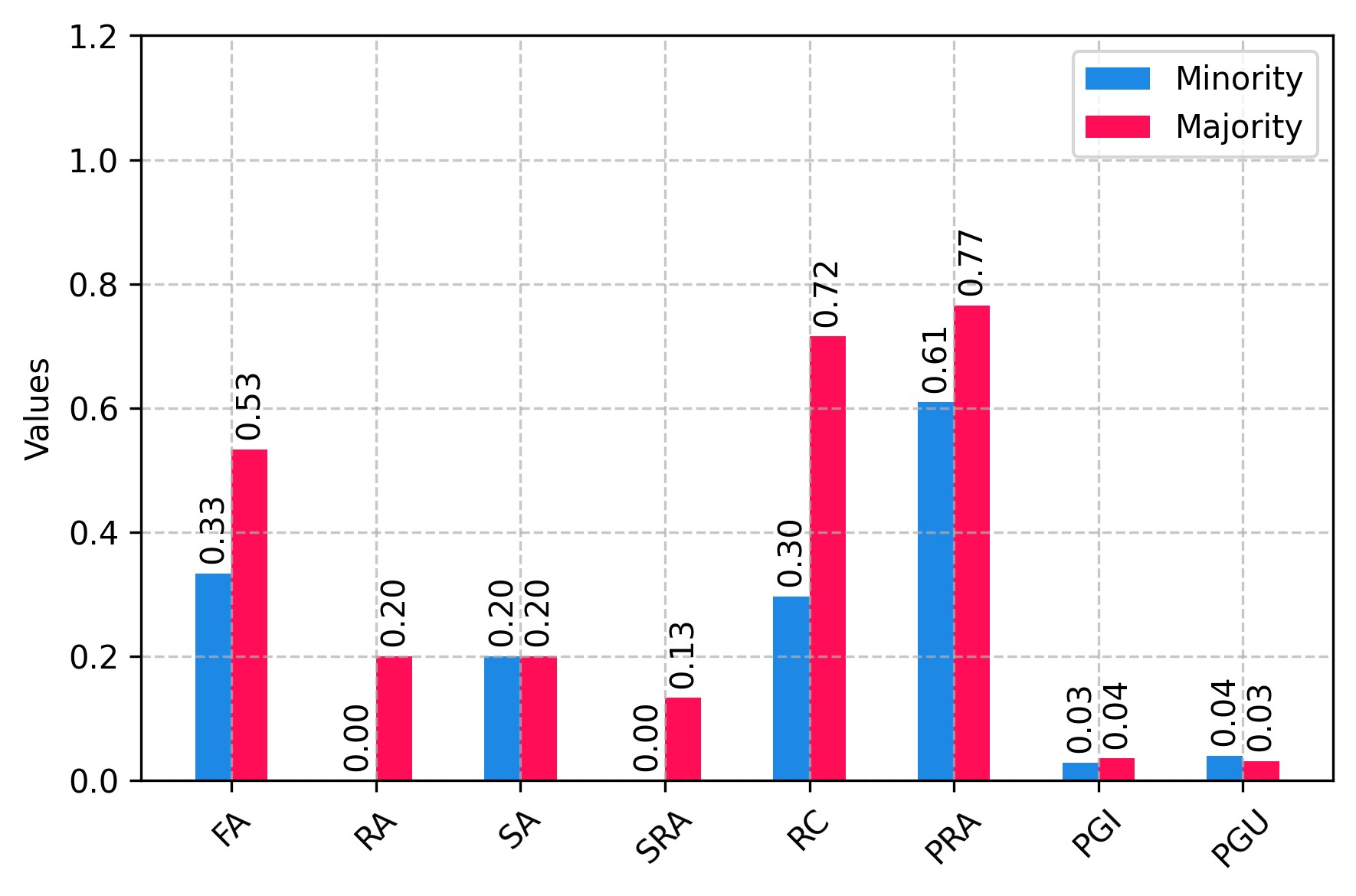}
    \end{subfigure}
    \vfill
    \begin{subfigure}[t]{0.3\textwidth}
        \centering
        \includegraphics[width=\textwidth]{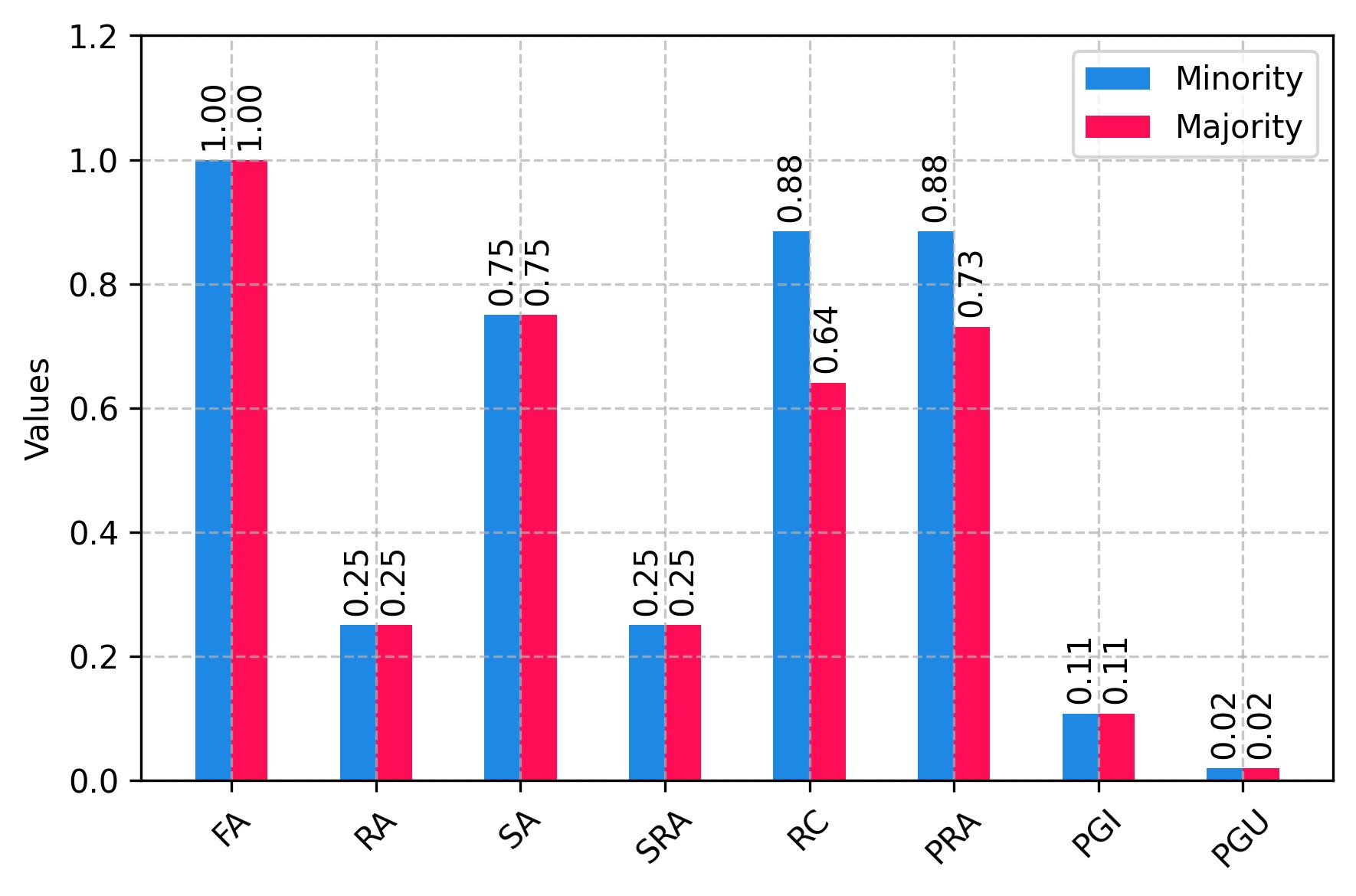}
        \caption{Adult Income}
    \end{subfigure}
    \hfill
    \begin{subfigure}[t]{0.3\textwidth}
        \centering
        \includegraphics[width=\textwidth]{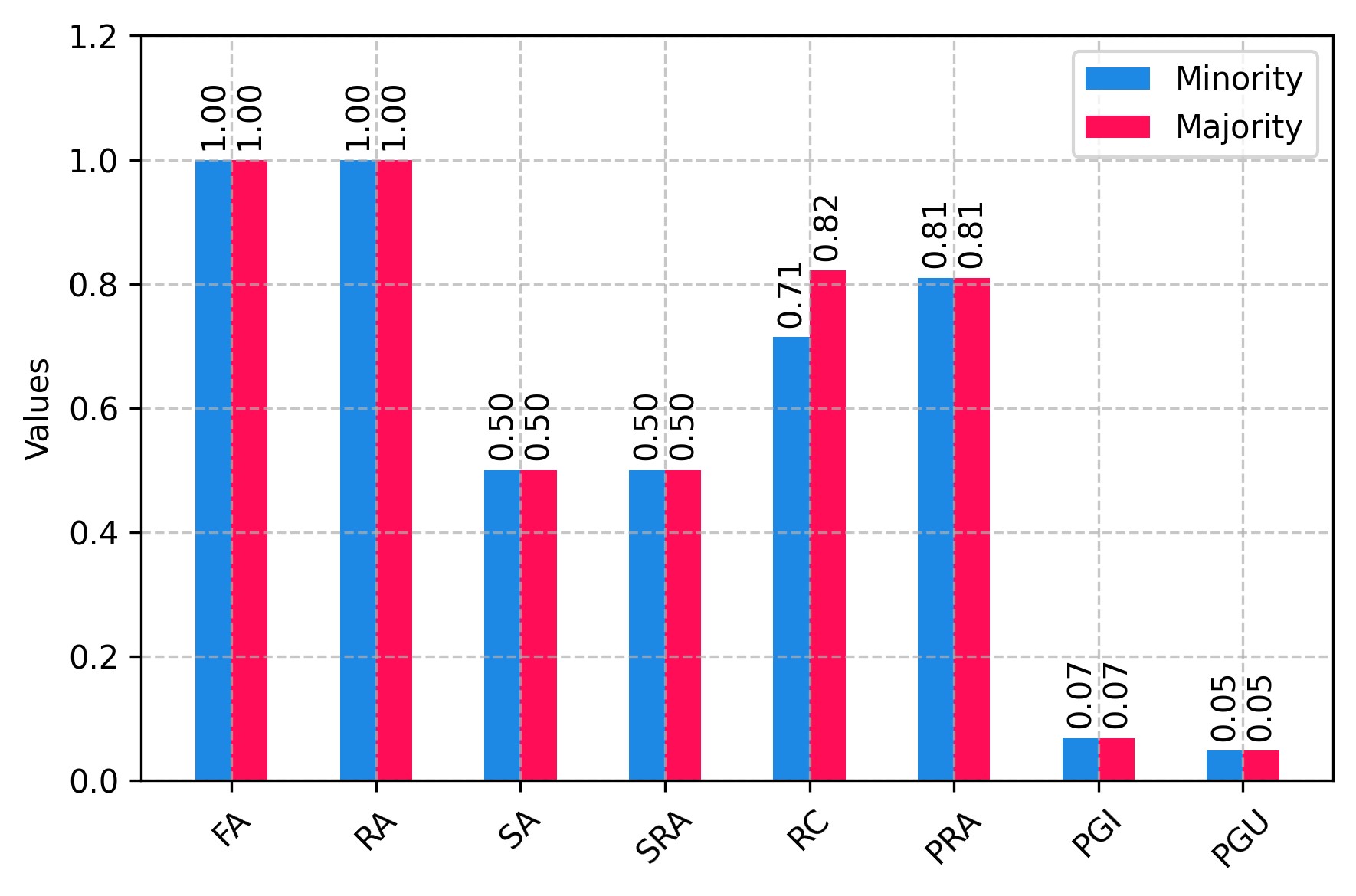}
        \caption{COMPAS}
    \end{subfigure}
    \hfill
    \begin{subfigure}[t]{0.3\textwidth}
        \centering
        \includegraphics[width=\textwidth]{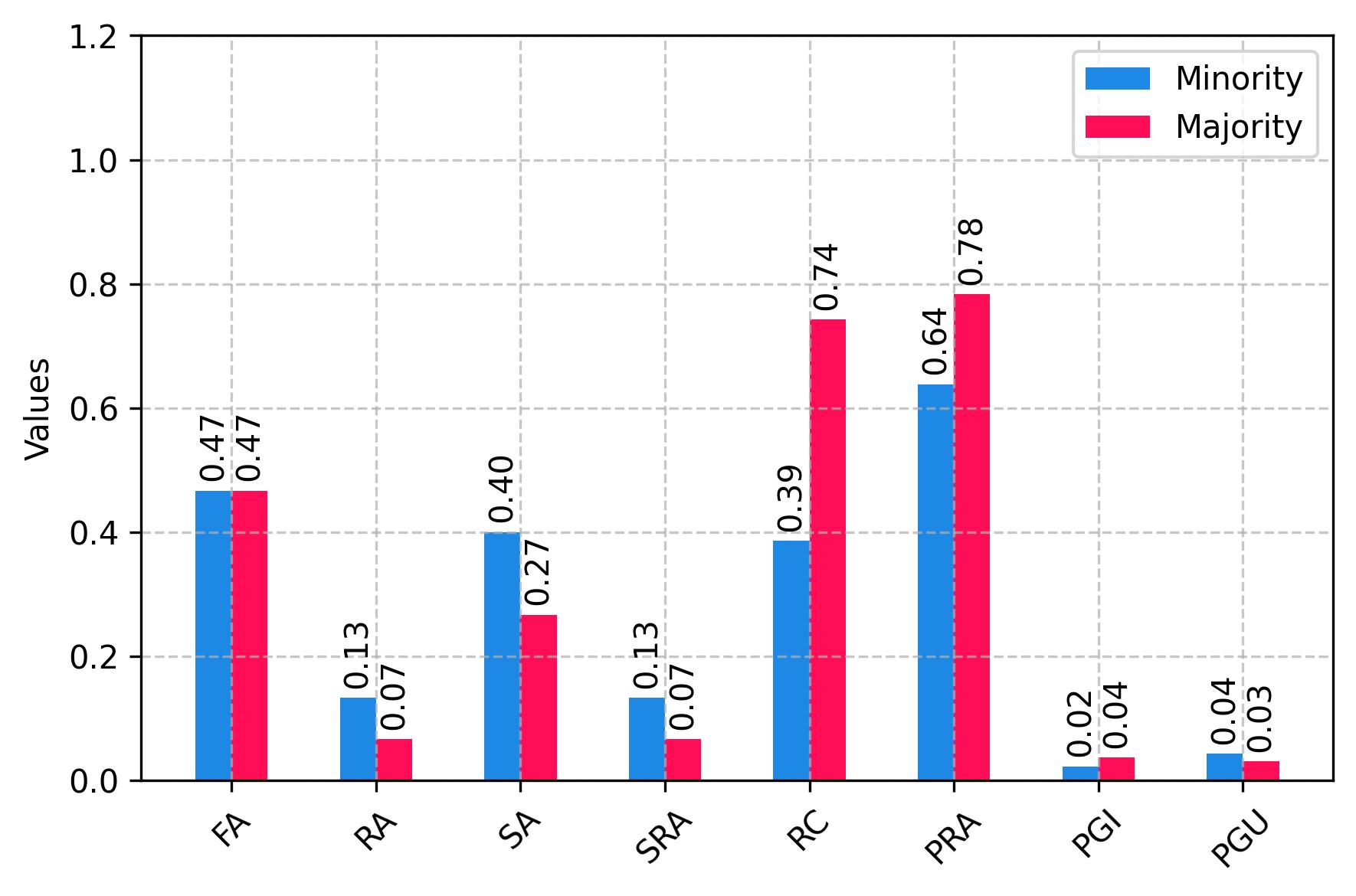}
        \caption{German Credit}
    \end{subfigure}
\caption{\label{fig:fairness_comparison} Comparison of fairness analysis between the LR model (top) and SAEM (bottom) for $k=0.25$ on the Adult Income, COMPAS, and German Credit datasets. Faithfulness metrics are shown for majority (male, red) and minority (female, blue) subgroups. Larger gaps between subgroup values indicate higher, undesirable disparities. }
\end{figure*}

\end{document}